\pdfoutput=1
\documentclass[10pt,twocolumn,letterpaper]{article}
\usepackage[pagenumbers]{paperstyle}      %
\usepackage{amsmath,amssymb,amsthm,booktabs,graphicx,multirow,xcolor,caption}
\usepackage{tikz}
\usepackage{adjustbox}           %

\usetikzlibrary{arrows.meta,positioning,calc}
\usepackage{microtype}
\definecolor{linkblue}{rgb}{0.21,0.49,0.74}
\usepackage[pagebackref,breaklinks,colorlinks,allcolors=linkblue]{hyperref}
\usepackage[capitalize]{cleveref}
\crefname{section}{Sec.}{Secs.}\Crefname{section}{Section}{Sections}
\crefname{table}{Tab.}{Tabs.}\Crefname{table}{Table}{Tables}

\newcommand{\AdaptAdvAnchRisk}{9.9}

\newcommand{\AdaptAdvLexRisk}{4.1}
\newcommand{\AdaptAdvLexSize}{8.92}

\newcommand{\AdaptFcAdvAnchRisk}{9.9}
\newcommand{\AdaptFcAdvAnchSize}{9.71}

\newcommand{\AdaptFcAdvFrozenRisk}{14.1}

\newcommand{\AdaptFcAdvLexFourRisk}{5.6}

\newcommand{\AdaptFcAdvLexOneRisk}{9.1}
\newcommand{\AdaptFcAdvLexOneSize}{8.71}
\newcommand{\AdaptFcAdvLexRisk}{5.6}
\newcommand{\AdaptFcAdvLexSize}{17.35}

\newcommand{\AdaptFcAdvRecalAbove}{67}
\newcommand{\AdaptFcAdvRecalRisk}{10.3}

\newcommand{\AdaptFcAdvWorstSize}{7.74}

\newcommand{\AdaptFcLexSize}{17,001}
\newcommand{\AdaptFcLexSmall}{1,690}

\newcommand{\AdaptFcPresentRecalRisk}{7.6}

\newcommand{\AdaptFcRandFiftyFrozenRisk}{10.3}

\newcommand{\AdaptFcRandFiftyRecalRisk}{7.3}

\newcommand{\AdaptFcRandFiveHundredRecalRisk}{9.4}

\newcommand{\AdaptFcTaggerRecalAbove}{62}
\newcommand{\AdaptFcTaggerRecalRisk}{10.2}

\newcommand{\AdaptLexFourNames}{17,001}

\newcommand{\AdaptLexSize}{17,001}

\newcommand{\AdaptPasAdvAnchRisk}{9.8}
\newcommand{\AdaptPasAdvAnchSize}{2.33}

\newcommand{\AdaptPasAdvFrozenRisk}{15.2}

\newcommand{\AdaptPasAdvRecalAbove}{51}
\newcommand{\AdaptPasAdvRecalRisk}{10.0}

\newcommand{\AdaptPasAdvWorstSize}{2.06}
\newcommand{\AdaptPasBAdvAnchRisk}{4.9}

\newcommand{\AdaptPasBAdvRecalRisk}{5.1}

\newcommand{\AdaptPasImages}{2,000}

\newcommand{\AdaptPasPresentFrozenRisk}{10.1}

\newcommand{\AdaptPasPresentNames}{1.1}

\newcommand{\AdaptPasPresentRecalRisk}{6.0}

\newcommand{\AdaptPasRandThreeHundredRecalRisk}{8.9}

\newcommand{\AdaptPasSedAdvAnchSize}{2.80}

\newcommand{\AdaptPasSedAdvFrozenRisk}{12.3}

\newcommand{\AdaptPasSedAdvRecalRisk}{21.1}

\newcommand{\AdaptPasSedAdvWorstSize}{2.42}
\newcommand{\AdaptPasSedBAdvAnchRisk}{4.8}

\newcommand{\AdaptPasSedBAdvRecalRisk}{12.7}

\newcommand{\AdaptPasSedPresentFrozenRisk}{9.9}

\newcommand{\AdaptPasSedRandFiftyFrozenRisk}{8.2}

\newcommand{\AdaptPasSedRandThreeHundredFrozenRisk}{5.0}

\newcommand{\AdaptPasSedRandThreeHundredRecalRisk}{10.6}

\newcommand{\AdaptPasSedTaggerFrozenRisk}{11.8}

\newcommand{\AdaptSedAdvAnchRisk}{9.9}
\newcommand{\AdaptSedAdvAnchSize}{5.87}

\newcommand{\AdaptSedAdvLexFourRisk}{25.4}

\newcommand{\AdaptSedAdvLexOneRisk}{21.8}

\newcommand{\AdaptSedAdvLexRisk}{25.4}
\newcommand{\AdaptSedAdvLexSize}{1.38}

\newcommand{\AdaptSedAdvRecalRisk}{18.8}

\newcommand{\AdaptSedAdvWorstSize}{5.00}

\newcommand{\AdaptSedImages}{2,000}

\newcommand{\AdaptSedLexSize}{17,001}
\newcommand{\AdaptSedLexSmall}{1,690}

\newcommand{\AdaptSedPresentFrozenRisk}{10.4}

\newcommand{\AdaptSedRandFiftyFrozenRisk}{9.0}

\newcommand{\AdaptSedRandFiveHundredFrozenRisk}{7.3}

\newcommand{\AdaptSedRandFiveHundredRecalRisk}{10.1}

\newcommand{\AdaptSedTaggerFrozenRisk}{12.7}

\newcommand{\AddedOracleMax}{45.7}
\newcommand{\AddedOracleMin}{26.7}

\newcommand{\AddedPixelFraction}{6.6}
\newcommand{\ApsAnchRiskHi}{11.4}
\newcommand{\ApsAnchRiskLo}{6.0}
\newcommand{\AuditBankPlusOneCostRatio}{130}
\newcommand{\AuditBankPlusOneDiff}{0.015}

\newcommand{\AuditImages}{400}
\newcommand{\AuditNativeFrozenRisk}{15.9}

\newcommand{\AuditUntruncatedFrozenRisk}{7.7}

\newcommand{\BankCandidateOnlyZeroSize}{82.34}

\newcommand{\BankCocoImages}{500}

\newcommand{\BankCocoOiSixEightThreeSize}{6.44}

\newcommand{\BankCocoOneSevenOneSize}{7.92}

\newcommand{\BankOverlapOiCocoAdded}{12}
\newcommand{\BankOverlapSource}{47}

\newcommand{\BankSeedsOverheadHi}{0.85}
\newcommand{\BankSeedsOverheadLo}{0.54}

\newcommand{\BankSeedsSharedHi}{20}
\newcommand{\BankSeedsSharedLo}{11}
\newcommand{\BankSeedsSizeSdSed}{0.11}
\newcommand{\BankSeedsSizeSdVitL}{0.11}
\newcommand{\BankSeedsSizeSpread}{0.28}
\newcommand{\BankSeedsSplitSdSed}{0.49}
\newcommand{\BankSeedsSplitSdVitL}{0.27}
\newcommand{\BankSeedsSplits}{200}
\newcommand{\BankSeedsTopOneMaxChange}{0.4}
\newcommand{\BankSweepImages}{400}

\newcommand{\CalSizeMaxN}{1,000}

\newcommand{\CheckCatsegReimpl}{1.4\times10^{-5}}

\newcommand{\CheckSedReimpl}{8.6\times10^{-6}}
\newcommand{\ClaudeCatAnchRisk}{9.8}
\newcommand{\ClaudeCatAnchSize}{4.56}
\newcommand{\ClaudeCatRecalSize}{3.69}
\newcommand{\ClaudeCatReusedRisk}{15.6}
\newcommand{\ClaudeCostImage}{0.034}
\newcommand{\ClaudeCrcSize}{845}
\newcommand{\ClaudeImages}{1,000}

\newcommand{\ClaudeProbChanged}{57}
\newcommand{\ClaudeSplits}{500}
\newcommand{\ClaudeTopChanged}{18}
\newcommand{\ClaudeTopOneFull}{62.8}
\newcommand{\ClaudeTopOneSrc}{67.5}
\newcommand{\ClaudeUnlistedFull}{17.1}
\newcommand{\ClaudeUnlistedSrc}{11.6}
\newcommand{\ClipsegFull}{46.2}
\newcommand{\ClipsegImages}{400}
\newcommand{\ClipsegRisk}{9.4}
\newcommand{\ClipsegSplits}{200}
\newcommand{\ClipsegSrc}{10.2}
\newcommand{\CocoAddedImages}{1,300}
\newcommand{\CocoAddedPixelFraction}{3.1}

\newcommand{\CocoDeclipBankAddedAnch}{25.2}

\newcommand{\CocoDeclipBankAddedMatched}{24.7}

\newcommand{\CocoFcclipBankAddedAnch}{23.1}

\newcommand{\CocoFcclipBankAddedMatched}{18.1}

\newcommand{\CocoFcclipBankNativeFrozenRisk}{17.1}

\newcommand{\CocoVitLBankNativeFrozenRisk}{12.8}

\newcommand{\ConfImages}{1,999}

\newcommand{\ConfPrimaryAnchRisk}{8.4}

\newcommand{\ConfPrimaryNativeFrozenRisk}{13.8}
\newcommand{\ConfPrimaryPrice}{0.85}

\newcommand{\ConfSecondaryAnchRisk}{9.7}

\newcommand{\ConfSecondaryNativeFrozenRisk}{15.0}

\newcommand{\DisjAnchRiskHi}{9.8}

\newcommand{\DisjBankImages}{400}
\newcommand{\DisjImages}{1,600}
\newcommand{\DisjSizeGapMax}{0.3}

\newcommand{\FcclipBankAddedAnch}{28.6}

\newcommand{\FcclipBankAddedMatched}{29.4}

\newcommand{\FcclipBankPrice}{2.99}

\newcommand{\FcclipNativeFrozenRisk}{13.8}

\newcommand{\FfAdded}{16.1}

\newcommand{\FfAnchRiskLo}{9.9}
\newcommand{\FfAnswered}{25.1}

\newcommand{\FfFcAnchSize}{8.00}

\newcommand{\FfFcLexRatio}{1.4}

\newcommand{\FfFcLexSize}{10.99}

\newcommand{\FfFcMatchedSize}{8.80}

\newcommand{\FfImages}{2,000}
\newcommand{\FfOutsideAde}{77}
\newcommand{\FfOutsideLex}{51}
\newcommand{\FfRecalRiskHi}{30.3}
\newcommand{\FfRecalRiskLo}{12.6}
\newcommand{\FfReusedRiskHi}{21.5}
\newcommand{\FfReusedRiskLo}{20.5}

\newcommand{\FfSedAnchSize}{5.69}

\newcommand{\FfSedLexRisk}{41.2}

\newcommand{\FfSedMatchedSize}{6.39}

\newcommand{\FfUnique}{5,797}

\newcommand{\FfVitLAnchSize}{4.59}

\newcommand{\FfVitLMatchedSize}{4.79}

\newcommand{\FtDistractAddedAnch}{44.9}

\newcommand{\FtDistractAddedMatched}{43.3}

\newcommand{\FtDistractAnchFull}{6.15}

\newcommand{\FtDistractHeldTopTen}{80.1}

\newcommand{\FtHeldLabels}{30}
\newcommand{\FtProbeImages}{400}

\newcommand{\FtTenKAnchFull}{7.01}

\newcommand{\HdFirstThreeMaxHard}{0.32}

\newcommand{\HdSedBQFiveRisk}{13.2}

\newcommand{\HdSedQFiveRisk}{25.6}

\newcommand{\HdTrials}{500}

\newcommand{\HdVitLBQFiveRisk}{12.5}

\newcommand{\HdVitLEpsTenRisk}{11.0}

\newcommand{\HdVitLPRisk}{9.4}

\newcommand{\HdVitLQAllRisk}{8.7}

\newcommand{\HdVitLQFiveBound}{36.1}

\newcommand{\HdVitLQFiveRisk}{24.9}

\newcommand{\LeanCheckImages}{20}
\newcommand{\LossAnchRiskHi}{10.06}
\newcommand{\LossAnchRiskLo}{9.84}

\newcommand{\LsAdeSedMHundredRisk}{9.3}
\newcommand{\LsAdeSedMHundredSize}{48.0}

\newcommand{\LsAdeSedMTenFloor}{86}

\newcommand{\LsAdeSedMTwentySize}{157.7}

\newcommand{\LsAdeSedRandSize}{45.1}

\newcommand{\LsAdeVitLMHundredHeld}{9.4}
\newcommand{\LsAdeVitLMHundredNatSize}{43.4}

\newcommand{\LsAdeVitLMHundredRisk}{9.3}
\newcommand{\LsAdeVitLMHundredSize}{29.5}

\newcommand{\LsAdeVitLMTenFloor}{86}

\newcommand{\LsAdeVitLMTwentySize}{102.1}

\newcommand{\LsAdeVitLRandSize}{27.6}

\newcommand{\LsCocSedMHundredNatSize}{32.7}

\newcommand{\LsCocSedMHundredSize}{29.4}

\newcommand{\LsCocVitLMHundredNatSize}{20.2}
\newcommand{\LsCocVitLMHundredPresent}{33}

\newcommand{\LsCocVitLMHundredSize}{13.7}

\newcommand{\LsCocVitLMTwentyFloor}{84}

\newcommand{\LsPasSedMHundredNatSize}{15.7}

\newcommand{\LsPasSedMHundredSize}{14.1}

\newcommand{\LsPasVitLMHundredNatSize}{11.5}
\newcommand{\LsPasVitLMHundredPresent}{42}

\newcommand{\LsPasVitLMHundredSize}{8.9}

\newcommand{\LsPasVitLMTwentyFloor}{54}

\newcommand{\LsPending}{}

\newcommand{\LxAdeFrozenRisk}{20.8}

\newcommand{\LxAdeImgRisk}{23.7}
\newcommand{\LxAdeImgSize}{6.8}

\newcommand{\LxAdeLabSize}{28.9}

\newcommand{\LxAdeRecalSize}{35.8}

\newcommand{\LxAdeSets}{10}

\newcommand{\LxCocSplitLabRisk}{17.6}

\newcommand{\LxCocSplitRecalRisk}{21.2}

\newcommand{\LxDAdeExchFrozenDelta}{11.6}

\newcommand{\LxDAdeSplitFrozenDelta}{24.4}

\newcommand{\LxDAdeSplitLabDelta}{12.8}
\newcommand{\LxDAdeSplitLabDeltaMax}{22.0}
\newcommand{\LxDAdeSplitLabDeltaRatio}{1.7}

\newcommand{\LxDAdeSplitRecalDelta}{13.3}

\newcommand{\LxDCocSplitLabDelta}{8.2}

\newcommand{\LxDExchDeltaMax}{1.3}

\newcommand{\LxDPasSplitLabDelta}{13.9}

\newcommand{\LxImgBMaxHeld}{4.5}

\newcommand{\LxImgBMinHeld}{2.4}
\newcommand{\LxImgMaxAbove}{41}
\newcommand{\LxImgMaxGap}{0.15}
\newcommand{\LxImgMaxHeld}{9.6}
\newcommand{\LxImgMinAbove}{23}
\newcommand{\LxImgMinHeld}{7.3}
\newcommand{\LxImgPending}{}

\newcommand{\LxModelsBMaxAbove}{36}

\newcommand{\LxModelsBMaxRisk}{4.6}
\newcommand{\LxModelsBMinAbove}{12}
\newcommand{\LxModelsBMinRisk}{2.5}
\newcommand{\LxModelsHeldMaxAbove}{42}
\newcommand{\LxModelsHeldMaxGap}{0.09}
\newcommand{\LxModelsHeldMinAbove}{22}
\newcommand{\LxModelsMaxAbove}{42}

\newcommand{\LxModelsMaxRisk}{9.6}
\newcommand{\LxModelsMinAbove}{23}
\newcommand{\LxModelsMinRisk}{7.5}
\newcommand{\LxModelsPending}{}

\newcommand{\LxPasSplitLabRisk}{22.3}

\newcommand{\LxPasSplitRecalRisk}{26.5}

\newcommand{\LxRsvAgree}{99.5}

\newcommand{\LxRsvBImgRisk}{13.9}

\newcommand{\LxRsvBLabRisk}{4.4}
\newcommand{\LxRsvBLabSize}{71.3}

\newcommand{\LxRsvBRecalRisk}{4.2}
\newcommand{\LxRsvBRecalSize}{200.3}

\newcommand{\LxRsvImgRisk}{23.1}

\newcommand{\LxRsvLabRisk}{9.3}
\newcommand{\LxRsvLabSize}{26.2}

\newcommand{\LxRsvRecalRisk}{9.4}
\newcommand{\LxRsvRecalSize}{32.8}

\newcommand{\LxRsvSplitLabDelta}{9.0}

\newcommand{\LxRsvSplitLabRisk}{18.4}

\newcommand{\LxTrials}{500}

\newcommand{\LxWAdePAllThreeRisk}{9.6}
\newcommand{\LxWAdePAllThreeSize}{27.4}

\newcommand{\MemBankSedMB}{160}
\newcommand{\MemBankVitLMB}{2.0}
\newcommand{\MemCalImages}{1,000}
\newcommand{\MemCalMB}{42}
\newcommand{\MemCard}{15.9}
\newcommand{\MemDeclipAnchMax}{8.44}

\newcommand{\MemDeclipNatFull}{9.10}

\newcommand{\MemFcMax}{2.7}

\newcommand{\MemImages}{3}
\newcommand{\MemSedAnchMax}{5.91}
\newcommand{\MemSedAnchMin}{4.65}
\newcommand{\MemSedNatFour}{27.4}
\newcommand{\MemSedNatFull}{8.01}

\newcommand{\MemVitLFullSaving}{0.7}

\newcommand{\MemVitLSrcExtra}{2.4}

\newcommand{\MiouGapFcclipAnchoredBankAdeEightFortySeven}{2.9}
\newcommand{\MiouGapFcclipAnchoredBankAdeOneFifty}{4.0}

\newcommand{\MiouGapHiPcFiftyNine}{1.7}

\newcommand{\MiouProtocolMaxChange}{0.02}

\newcommand{\NrSedAlphaPrime}{15.25}

\newcommand{\NrSedAnchSrcSize}{2.76}

\newcommand{\NrSedBOutside}{0.00}

\newcommand{\NrSedNatSrcSize}{2.22}
\newcommand{\NrSedOutside}{0.20}

\newcommand{\NrVitLAlphaPrime}{6.00}

\newcommand{\NrVitLAnchSrcSize}{2.25}

\newcommand{\NrVitLBOutside}{0.01}

\newcommand{\NrVitLNatSrcSize}{1.91}
\newcommand{\NrVitLOutside}{0.34}

\newcommand{\NrVitLRecalSize}{3.47}

\newcommand{\NrVitLTightSize}{3.48}
\newcommand{\NumAdeImages}{2,000}

\newcommand{\NumSplits}{500}
\newcommand{\OvdAccAddedBoxes}{12,952}

\newcommand{\OvdAccAllDetectors}{three}
\newcommand{\OvdAccAllRiseCount}{two}
\newcommand{\OvdAccAnchAAllHi}{54.7}
\newcommand{\OvdAccAnchAAllLo}{49.8}
\newcommand{\OvdAccAnchAHi}{16.5}
\newcommand{\OvdAccAnchALo}{14.7}
\newcommand{\OvdAccAnchALocHi}{22.3}
\newcommand{\OvdAccAnchALocLo}{20.3}
\newcommand{\OvdAccAnchASrcHi}{88.0}
\newcommand{\OvdAccAnchASrcLo}{79.7}
\newcommand{\OvdAccAnchBSrcHi}{51.3}
\newcommand{\OvdAccAnchBSrcLo}{43.0}
\newcommand{\OvdAccNativeAllHi}{59.4}
\newcommand{\OvdAccNativeAllLo}{48.3}
\newcommand{\OvdAccNativeHi}{34.2}
\newcommand{\OvdAccNativeLo}{23.8}
\newcommand{\OvdAccNativeLocHi}{38.7}
\newcommand{\OvdAccNativeLocLo}{28.6}
\newcommand{\OvdAccNativeSrcHi}{89.7}
\newcommand{\OvdAccNativeSrcLo}{65.3}

\newcommand{\OvdGdBAnchBSrcSize}{4.1}

\newcommand{\OvdGdBAnchUnlocSrc}{2.1}

\newcommand{\OvdGdBBAnchASize}{12.2}

\newcommand{\OvdGdBBRecalSize}{9.7}

\newcommand{\OvdGdBBReuseOneRisk}{9.9}

\newcommand{\OvdGdBCheckMax}{8.1\times10^{-6}}

\newcommand{\OvdGdBLxAnchBenchHeld}{66.1}

\newcommand{\OvdGdBLxNativeBenchHeld}{63.6}

\newcommand{\OvdGdBReuseOneRisk}{18.3}

\newcommand{\OvdGdBRtAnch}{0.52}
\newcommand{\OvdGdBRtNative}{3.31}

\newcommand{\OvdGdTAnchBSrcSize}{11.9}
\newcommand{\OvdGdTAnchTrivialImages}{57}

\newcommand{\OvdGdTAnchUnlocAdded}{23.6}
\newcommand{\OvdGdTAnchUnlocSrc}{2.5}

\newcommand{\OvdGdTBoxes}{14,855}
\newcommand{\OvdGdTCheckMax}{7.2\times10^{-6}}
\newcommand{\OvdGdTImages}{1,989}

\newcommand{\OvdGdTLxAnchBenchHeld}{38.1}

\newcommand{\OvdGdTLxAnchExchCal}{4.8}

\newcommand{\OvdGdTLxAnchExchHeld}{4.8}

\newcommand{\OvdGdTLxNativeBenchHeld}{19.0}

\newcommand{\OvdGdTLxNativeExchHeld}{7.3}
\newcommand{\OvdGdTNativeTrivialImages}{17}

\newcommand{\OvdGdTNativeUnlocAdded}{17.0}

\newcommand{\OvdGdTOneLabels}{100}

\newcommand{\OvdGdTRtAnch}{0.50}
\newcommand{\OvdGdTRtNative}{2.97}

\newcommand{\OvdMmTAnchTrivialImages}{74}

\newcommand{\OvdMmTAnchUnlocAdded}{29.9}

\newcommand{\OvdMmTBAnchTrivialImages}{100}

\newcommand{\OvdMmTBNativeTrivialImages}{77}

\newcommand{\OvdMmTCheckMax}{1.8\times10^{-4}}

\newcommand{\OvdMmTLxAnchExchCal}{3.2}

\newcommand{\OvdMmTLxAnchExchHeld}{3.2}

\newcommand{\OvdMmTLxNativeExchHeld}{7.9}
\newcommand{\OvdMmTNativeTrivialImages}{5}

\newcommand{\OvdMmTNativeUnlocAdded}{11.6}

\newcommand{\OvdOwlAccOwlAddTop}{55.4}

\newcommand{\OvdOwlAddedHeld}{7.9}

\newcommand{\OvdOwlRisk}{9.9}

\newcommand{\OvdOwlSize}{8.3}
\newcommand{\OvdOwlTrivialImages}{3}
\newcommand{\OvdOwlUnlocAdded}{10.7}

\newcommand{\OvdPending}{}
\newcommand{\OvdRtSpeedupHi}{6.3}
\newcommand{\OvdRtSpeedupLo}{6.0}
\newcommand{\OverviewAnchFracMax}{43}

\newcommand{\OverviewAnchFracMin}{38}

\newcommand{\OverviewCostAggPriceMax}{0.99}
\newcommand{\OverviewCostAggPriceMaxB}{1.16}
\newcommand{\OverviewReplayFracMax}{46}

\newcommand{\OverviewReplayFracMin}{38}

\newcommand{\PascalAddedPixelFraction}{5.5}

\newcommand{\PascalFcclipBankAddedAnch}{36.5}

\newcommand{\PascalFcclipBankAddedMatched}{34.9}

\newcommand{\PnCalAdeVitL}{160}

\newcommand{\PnHybRatioHi}{3.6}

\newcommand{\PnHybRatioLo}{1.8}

\newcommand{\PnHybRiskHi}{4.9}

\newcommand{\PnImgOverAdeVitL}{154}

\newcommand{\PnNatEstAdeFc}{131}

\newcommand{\PnNatEstAdeVitL}{131}

\newcommand{\PnNatOverAdeFc}{65}

\newcommand{\PnNatOverAdeVitL}{89}

\newcommand{\PnPresentAdeVitL}{836}

\newcommand{\PnSizeRatioHi}{3.2}

\newcommand{\PnSizeRatioLo}{1.5}

\newcommand{\QualAnchMiss}{2.7}

\newcommand{\QualPanelAnchMiss}{22.1}
\newcommand{\QualPanelAnchSize}{4.8}
\newcommand{\QualPanelImages}{6}
\newcommand{\QualPanelPool}{305}
\newcommand{\QualPanelRecalMiss}{20.1}
\newcommand{\QualPanelRecalSize}{4.6}
\newcommand{\QualPanelReusedMiss}{26.0}
\newcommand{\QualPanelReusedSize}{2.9}
\newcommand{\QualPoolAnchMiss}{9.7}
\newcommand{\QualPoolImages}{305}
\newcommand{\QualPoolRecalMiss}{9.8}
\newcommand{\QualPoolReusedMiss}{15.6}
\newcommand{\QualPoolSourceMiss}{9.8}

\newcommand{\QualRecalMiss}{5.5}

\newcommand{\QualReusedMiss}{29.2}

\newcommand{\QualSourceMiss}{2.5}

\newcommand{\RapsAnchRiskHi}{12.2}

\newcommand{\RrDelta}{10}

\newcommand{\RrFcAdeCross}{1,000}

\newcommand{\RrSedAdeCross}{5}

\newcommand{\RrSedCocCross}{1}

\newcommand{\RrVitLAdeAnchSize}{5.11}

\newcommand{\RrVitLAdeCross}{50}
\newcommand{\RrVitLAdeNatHundred}{5.37}

\newcommand{\RtCalImages}{1,000}

\newcommand{\RtFcclipBankBreakEvenMax}{3,000}

\newcommand{\RtFcclipBankRatioMin}{1.3}
\newcommand{\RtFcclipBankRecalMinMax}{5.1}

\newcommand{\RtLeanDeclipBankRatio}{3.6}

\newcommand{\RtLeanSedBankRatio}{1.4}
\newcommand{\RtLeanSedBankSaving}{16}

\newcommand{\RtLeanVitLBankRatio}{3.7}
\newcommand{\RtLeanVitLBankSaving}{15}

\newcommand{\RtLocalDeclipBankRatio}{4.2}

\newcommand{\RtLocalSedBankRatio}{1.6}

\newcommand{\RtLocalVitLBankRatio}{4.3}
\newcommand{\RtLocalVitLBankRecalMin}{16.4}

\newcommand{\RtSedBankBreakEven}{1,400}
\newcommand{\RtSedBankBreakEvenCached}{1,300}

\newcommand{\RtSedBankRatio}{1.7}
\newcommand{\RtSedBankRecalMin}{6.1}

\newcommand{\RtVitBBankBreakEven}{300}
\newcommand{\RtVitBBankBreakEvenCached}{270}

\newcommand{\RtVitLBankBreakEven}{340}
\newcommand{\RtVitLBankBreakEvenCached}{270}

\newcommand{\RtVitLBankRatio}{4.0}
\newcommand{\RtVitLBankRecalMin}{5.7}

\newcommand{\SedBankCalSettleN}{100}

\newcommand{\SedBankNativeFrozenSize}{8.80}

\newcommand{\SedBankReplaySize}{5.28}

\newcommand{\SedFastNativeFrozenRisk}{12.8}

\newcommand{\SedNativeFrozenRisk}{7.1}

\newcommand{\SegAddClassesAdeEightFortySeven}{697}
\newcommand{\SegAddClassesPcFourFiftyNine}{399}

\newcommand{\SegAddFcclipAnchoredBankAdeEightFortySevenAdd}{23.2}

\newcommand{\SegAddFcclipAnchoredBankPcFourFiftyNineAdd}{31.4}

\newcommand{\SegAddFcclipNativeAdeEightFortySevenAdd}{37.5}

\newcommand{\SegAddFcclipNativePcFourFiftyNineAdd}{35.5}

\newcommand{\SegAddPixCocoGainHi}{3.5}
\newcommand{\SegAddPixCocoGainLo}{1.3}
\newcommand{\SegAddPixMaxAbsChange}{1.4}

\newcommand{\SegAddPresentPcFourFiftyNine}{285}

\newcommand{\SegAddShareAdeEightFortySeven}{6.9}
\newcommand{\SegAddSharePcFourFiftyNine}{5.0}

\newcommand{\SilentGrowthAnchHi}{4.58}
\newcommand{\SilentGrowthAnchLo}{1.54}
\newcommand{\SilentGrowthReusedHi}{1.62}
\newcommand{\SilentGrowthReusedLo}{1.09}
\newcommand{\SilentN}{15}
\newcommand{\SilentNUnder}{12}
\newcommand{\SilentShareHi}{40}
\newcommand{\SilentShareLo}{14}
\newcommand{\SoftmaxAnchRiskHi}{17.4}
\newcommand{\SoftmaxAnchRiskLo}{12.2}

\newcommand{\SsdConfigs}{5}

\newcommand{\SsdLConfigs}{5}

\newcommand{\SsdLPascalVitLBankImgMedian}{2}

\newcommand{\SsdLPascalVitLBankLabMedian}{6}

\newcommand{\SsdLSplitVitLBankLabGtTen}{32.9}

\newcommand{\SsdLSplitVitLBankLabMedian}{7}

\newcommand{\SsdLVitLBankImgGtTen}{20.3}

\newcommand{\SsdLVitLBankImgMedian}{5}

\newcommand{\SsdLVitLBankLabGtTen}{75.3}

\newcommand{\SsdLVitLBankLabMedian}{20}

\newcommand{\SsdLVitLBankNatLabGtTen}{87.4}

\newcommand{\SsdLVitLBankNatLabMedian}{27}

\newcommand{\SsdPascalVitLBankAnchOne}{44.6}

\newcommand{\SsdPascalVitLBankRecalOne}{47.4}

\newcommand{\SsdSedBankAnchGtTen}{17.1}

\newcommand{\SsdSedBankAnchOne}{19.4}

\newcommand{\SsdSedBankRecalGtTen}{4.8}

\newcommand{\SsdSedBankRecalOne}{10.5}

\newcommand{\SsdSedBankReusedGtTen}{33.0}

\newcommand{\SsdVitLBankAnchGtTen}{5.5}

\newcommand{\SsdVitLBankAnchOne}{23.5}

\newcommand{\SsdVitLBankRecalGtTen}{0.6}

\newcommand{\SsdVitLBankRecalOne}{22.1}

\newcommand{\VitBAnchFull}{7.32}

\newcommand{\VitBBankAnchFull}{5.98}

\newcommand{\VitBBankAnchRiskOne}{9.9}

\newcommand{\VitBBankAnchSrc}{2.76}

\newcommand{\VitBCalSettleN}{200}

\newcommand{\VitBHeldTopTen}{82.7}

\newcommand{\VitBNativeFrozenRisk}{15.4}

\newcommand{\VitBSclipFull}{99.3}
\newcommand{\VitBSclipRisk}{9.8}
\newcommand{\VitBSclipSrc}{25.0}

\newcommand{\VitBWorstSceneAnch}{20.7}
\newcommand{\VitBWorstSceneName}{bathroom}
\newcommand{\VitBWorstSceneReplay}{11.1}

\newcommand{\VitLBankAddedAnch}{32.4}

\newcommand{\VitLBankAddedMatched}{36.1}

\newcommand{\VitLBankCalSettleN}{50}

\newcommand{\VitLBankMacroAnch}{16.6}

\newcommand{\VitLBankMacroReplay}{20.6}
\newcommand{\VitLBankMatchedSourceRisk}{8.1}

\newcommand{\VitLBankNativeFrozenRisk}{15.2}

\newcommand{\VitLBankNativeFrozenSize}{2.16}

\newcommand{\VitLBankPrice}{0.79}

\newcommand{\VitLBankReplaySize}{3.47}

\newcommand{\VitLBankSrcSize}{1.91}

\newcommand{\VitLPrice}{1.61}

\newtheorem{proposition}{Proposition}
\usepackage{needspace,etoolbox}\BeforeBeginEnvironment{proposition}{\needspace{3\baselineskip}}%

\newcommand{\ind}{\mathbf{1}}
\newcommand{\Vsrc}{V_0}
\newcommand{\Vdep}{V}
\newcommand{\bank}{B}
\newcommand{\score}{s}

\title{Calibrate Once, Grow the Vocabulary:\\Candidate-Independent Scores for Conformal Risk Control\\in Open-Vocabulary Segmentation}

\author{Md Tanveer Hossain Munim$^{1}$ \qquad Bijoy Ahmed Saiem$^{1}$ \qquad Muhammad Abdullah Adnan$^{1}$\\
$^{1}$Bangladesh University of Engineering and Technology\\
{\tt\small buet.ac.bd}}

\hypersetup{pdftitle={Calibrate Once, Grow the Vocabulary: Candidate-Independent Scores for Conformal Risk Control in Open-Vocabulary Segmentation}, pdfauthor={Md Tanveer Hossain Munim, Bijoy Ahmed Saiem, Muhammad Abdullah Adnan}}
\begin{document}
\maketitle
\begin{abstract}
Open-vocabulary segmenters accept class names at inference. Conformal risk control can bound how often their prediction sets miss the true label, but we show that this bound breaks when names are added after calibration. On ADE20K, a threshold calibrated for $10\%$ risk at $150$ names reaches up to $\VitBNativeFrozenRisk\%$ risk at $845$ names in released CAT-Seg models. The risk exceeds the target on every calibration split, yet the prediction sets grow far less than the target requires. In SED they instead grow larger than needed. The cause is that pruning, cross-label attention and vocabulary normalisation make each label's score depend on the other requested labels. Scores can move either way, so a threshold recalibrated at a larger vocabulary need not hold at smaller ones. We remove this dependence without training by scoring each label against a fixed reference bank of names. A label's score then stays the same when names are added, so one calibration keeps its guarantee as the vocabulary grows. The guarantee holds whatever names are added, even ones chosen per image after seeing it. On three benchmarks and five segmenters, the converted models meet the target at every vocabulary size we evaluate. For CAT-Seg and SED at $\alpha=10\%$, they use at most one more label per pixel than the released models recalibrated at the grown vocabulary. Treating names instead of images as calibration samples also bounds the miss rate of added names. If the annotated names are a random sample of the names in the calibration images, an added name's expected miss rate on those images is at most the target. For harder added names, the excess is at most their gap in mean miss rate. Applied to Grounding DINO, which splits large vocabularies into chunks, the conversion also makes inference about six times faster.
\end{abstract}

\section{Introduction}
\label{sec:intro}

Open-vocabulary segmenters assign pixels to classes named in text \cite{li2022lseg,ghiasi2022openseg,cho2024catseg,xie2024sed,yu2023fcclip}. The list of requested names, the \emph{vocabulary}, often changes after deployment. Robots are asked for objects outside the list their map was built with \cite{huang2023vlmaps,liu2024okrobot,yenamandra2023homerobot}, species catalogues grow between releases \cite{vanhorn2018inat,vanhorn2021inat,stevens2024bioclip}, and taggers, captioners or language models choose names per image \cite{conti2023vocabfree,vocabfree2025,zhang2024ram,ren2024groundedsam,gu2024conceptgraphs,gupta2023visprog,pi2023detgpt,liu2024llavaplus} (\cref{sec:deploy}). (Lettered sections and S-numbered tables and figures are in the supplementary material.)

A prediction set offers several plausible labels for a pixel rather than a single guess. In an annotation tool, for example, it shows the annotator the plausible labels and bounds how often the correct one is missing. Conformal prediction \cite{vovk2005alrw,angelopoulos2023gentle} and conformal risk control (CRC) \cite{angelopoulos2024crc}, both applied to segmentation \cite{mossina2024conformal,brunekreef2024kandinsky}, calibrate a score threshold on labelled images so that the expected fraction of missed pixels, the \emph{risk}, stays below a target $\alpha$. The guarantee requires calibration and test images to be exchangeable.

\begin{figure*}[t]
  \centering
  \includegraphics[width=\linewidth]{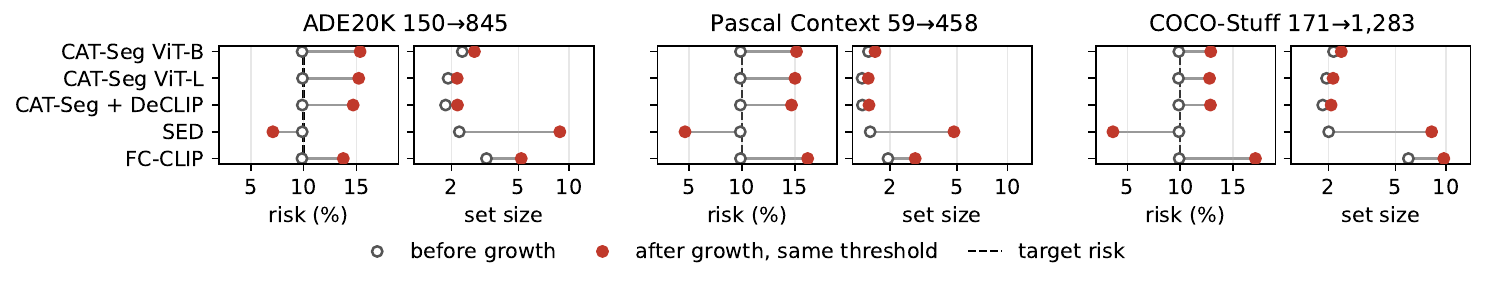}
  \caption{\textbf{Calibration breaks when names are added.} Each released segmenter is calibrated by CRC for a risk of $\alpha=10\%$ (dashed) on the source vocabulary, and the same threshold is used after the vocabulary grows (means over \NumSplits\ random calibration splits, set size in labels per pixel). After growth, the risk of CAT-Seg and FC-CLIP exceeds the target, and the prediction sets of SED become several times larger.}
  \label{fig:teaser}
\end{figure*}

Calibration uses a \emph{source vocabulary}, and the guarantee counts pixels whose true label belongs to it. Adding names produces a \emph{grown vocabulary} and raises two questions: does the guarantee for the source labels survive, and what can be guaranteed for the unannotated \emph{added labels}? Recalibrating on the labelled images restores the source-label guarantee when the new vocabulary, or a fixed rule for choosing it, is known in advance \cite{jin2025focal}. This needs little compute but is costly to operate. Every change of vocabulary needs another recalibration, images processed before it finishes are not covered, and the outputs barely show when the guarantee has broken. We study how to keep the guarantee without recalibrating (\cref{sec:recaldeploy}).

In released segmenters, adding names can push the risk above the target or make the sets \emph{over-conservative}, larger than the target requires. The cause is that the score of a requested label, or \emph{candidate}, depends on the other candidates. We call this \emph{candidate dependence}. CAT-Seg's code \cite{cho2024catseg} keeps only the $256$ labels with the highest initial scores in each image crop, SED \cite{xie2024sed} lets the labels attend to one another, and FC-CLIP \cite{yu2023fcclip} normalises its scores with a softmax over the vocabulary. Adding labels therefore changes the scores of the labels already present, and with them the risk of a threshold calibrated at the source vocabulary (\cref{fig:teaser}). When ADE20K \cite{zhou2017ade20k} grows from its $150$ benchmark classes to $845$ classes, a threshold calibrated for $10\%$ risk reaches $\FcclipNativeFrozenRisk$--$\VitBNativeFrozenRisk\%$ risk with CAT-Seg and FC-CLIP. It exceeds the target on every one of \NumSplits\ random calibration splits.

\begin{figure}[t]
  \centering
  \resizebox{\linewidth}{!}{\begin{tikzpicture}[
  font=\scriptsize,
  tok/.style={draw, rounded corners=1pt, minimum width=6.2mm, minimum height=4mm, inner sep=1pt},
  cand/.style={tok, fill=blue!12, draw=blue!60!black},
  banktok/.style={tok, fill=gray!18, draw=gray!60!black},
  pad/.style={tok, fill=gray!4, draw=gray!60!black, densely dashed},
  sum/.style={draw, rounded corners=2pt, fill=orange!15, draw=orange!70!black, inner sep=2pt, minimum height=5mm},
  arr/.style={-{Latex[length=1.3mm]}, gray!60!black},
  att/.style={-{Latex[length=1.3mm]}, thick, blue!60!black},
  ttl/.style={font=\scriptsize\bfseries, anchor=west},
]
\node[ttl] at (-0.35,1.5) {(a) native};
\foreach \i/\n in {0/{c_1},1/{c_2},2/{\cdots},3/{c_{|\Vdep|}}} {\node[cand] (n\i) at (0.8*\i,0.85) {$\n$};}
\node[pad] (np) at (3.2,0.85) {$\Pi$};
\node[font=\tiny, red!60!black, anchor=east] at (3.55,1.5) {top-256 per crop (CAT-Seg)};
\node[sum, minimum width=3.85cm] (ns) at (1.6,0) {$\sum_{j\in\Vdep\cup\Pi}$};
\foreach \i in {0,1,3} {\draw[arr] (n\i.south) -- (n\i.south |- ns.north);}
\draw[arr] (np.south) -- (np.south |- ns.north);
\foreach \i/\n in {0/{c_1},1/{c_2},3/{c_{|\Vdep|}}} {\node[cand] (o\i) at (0.8*\i,-0.8) {$\n$};}
\node at (1.6,-0.8) {$\cdots$};
\foreach \i in {0,1,3} {\draw[att] (o\i.north |- ns.south) -- (o\i.north);}
\node[anchor=north, align=center] at (1.6,-1.1) {$\score_{\Vdep}(x,p,c)$ depends on all of $\Vdep$};
\begin{scope}[xshift=4.6cm]
\node[ttl] at (-0.35,1.5) {(b) anchored};
\foreach \i/\n in {0/{b_1},1/{b_2},2/{\cdots},3/{b_{|\bank|}}} {\node[banktok] (b\i) at (0.8*\i,0.85) {$\n$};}
\node[pad] (bp) at (3.2,0.85) {$\Pi$};
\node[font=\tiny, gray!40!black, anchor=east] at (3.55,1.5) {fixed bank, run once per image};
\node[sum, minimum width=3.85cm] (bs) at (1.6,0) {$S_{kv},\,S_k$ (cached)};
\foreach \i in {0,1,3} {\draw[arr] (b\i.south) -- (b\i.south |- bs.north);}
\draw[arr] (bp.south) -- (bp.south |- bs.north);
\node[cand] (c1) at (0.4,-0.8) {$c$};
\node[cand] (c2) at (2.8,-0.8) {$c'$};
\draw[att] (c1.north |- bs.south) -- (c1.north);
\draw[att] (c2.north |- bs.south) -- (c2.north);
\draw[att] (c1.south west) to[out=-120, in=180, looseness=5] (c1.west);
\draw[att] (c2.south east) to[out=-60, in=0, looseness=5] (c2.east);
\draw[red!60!black, dashed, {Latex[length=1.1mm]}-{Latex[length=1.1mm]}] (c1.east) -- node[fill=white, inner sep=0.5pt, font=\scriptsize] {$\times$} (c2.west);
\node[anchor=north, align=center] at (1.6,-1.1) {$\score(x,p,c)$ depends on $x$, $c$, $\bank$ only};
\end{scope}
\end{tikzpicture}}
  \caption{\textbf{Class attention at one pixel.} (a) The sums \eqref{eq:linear} run over the requested labels, so every score depends on $\Vdep$. (b) They run over a fixed bank, computed once per image. Each candidate attends to the bank and to itself but not to other candidates.}
  \label{fig:arch}
\end{figure}
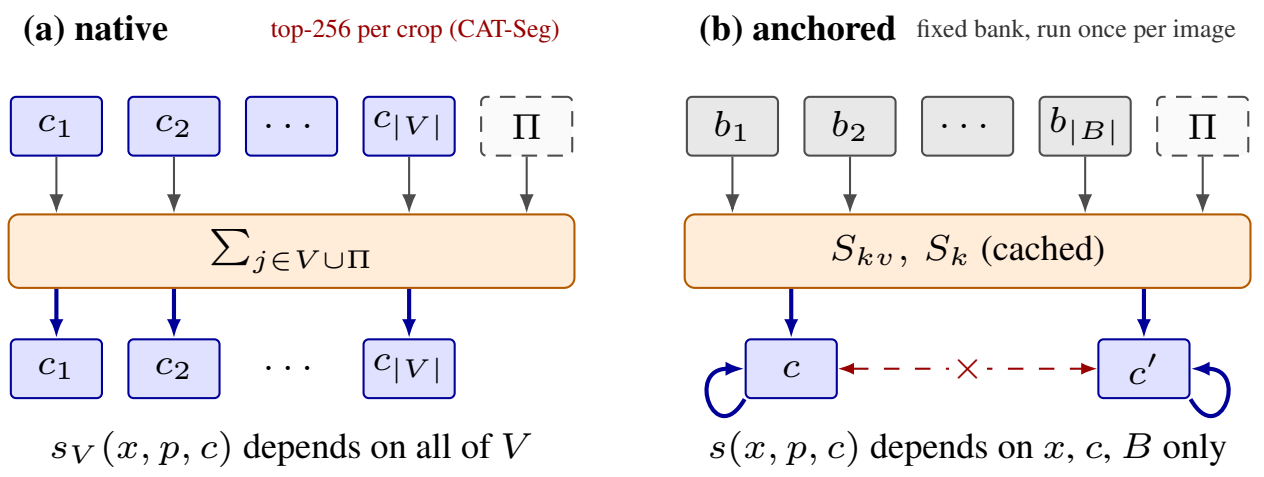

We make the scores \emph{candidate-independent} by scoring each label against a fixed \emph{reference bank} of label names instead of the requested vocabulary. The released network processes the bank without retraining (\cref{fig:arch}). Each candidate attends to the bank and to itself but not to other candidates, so its score depends only on the image, the label and the bank. We call the converted model \emph{anchored} and the released model \emph{native}. The score of a source label then stays the same when names are added. A threshold calibrated once therefore keeps its guarantee for every larger vocabulary, including vocabularies chosen per image. The anchored sets remain far smaller than those of models that are candidate-independent by design, such as CLIPSeg \cite{luddecke2022clipseg} (\cref{sec:main}).

Candidate independence also permits a separate guarantee for added labels. On fixed images, a label's \emph{miss rate}, the fraction of its pixels left out at a given threshold, then does not depend on the vocabulary. CRC can therefore treat labels instead of images as its exchangeable samples. If the annotated source labels are a random sample of the names that may be requested, CRC bounds the expected miss rate of a randomly added label on the calibration images. Annotating a random subset of the names gives such a sample. When added labels are harder, such as rare benchmark classes, we bound the excess risk by the difference in mean miss rate between added and source labels. We also characterise the worst case, when nothing is assumed about the added labels.

We evaluate five released models (CAT-Seg with ViT-B, ViT-L and DeCLIP \cite{wang2025declip} backbones, SED and FC-CLIP) on ADE20K $150\!\to\!845$, Pascal Context \cite{mottaghi2014pascalcontext} $59\!\to\!458$ and COCO-Stuff \cite{caesar2018cocostuff} $171\!\to\!1{,}283$, grown with LVIS \cite{gupta2019lvis} categories. The anchored models meet the target at every evaluated vocabulary size on every benchmark. At $\alpha=10\%$, their prediction sets have at most one more label per pixel than native recalibration for CAT-Seg and SED, and about three more for FC-CLIP on ADE20K. For CAT-Seg and SED, their mIoU is at most $\MiouGapHiPcFiftyNine$ points below the native models. Our contributions are:
\begin{itemize}\setlength\itemsep{0pt}
\item We show that CRC thresholds calibrated at a source vocabulary fail after growth in five released segmenters. For CAT-Seg and FC-CLIP the risk exceeds the target on every split while the sets grow far less than needed. SED's sets become over-conservative. We trace both failures to pruning, attention across labels and softmax normalisation, and show that recalibrating at a larger vocabulary does not cover per-image vocabularies (\cref{sec:problem}).
\item We convert released CAT-Seg, SED and FC-CLIP segmenters and Grounding DINO detectors to candidate-independent scores without additional training, so that one threshold calibrated at the source vocabulary remains valid for every grown vocabulary (\cref{sec:conversion,sec:main,sec:adaptive}).
\item We give a guarantee for added labels that are exchangeable with the source labels, bound the excess risk when added labels are harder, and characterise the worst case with no assumption on the added labels (\cref{sec:addedtheory,sec:labelguarantee}).
\end{itemize}

\section{Related work}
\label{sec:related}

\paragraph{Conformal prediction in dense and zero-shot vision.}
Conformal prediction and risk control \cite{angelopoulos2024crc} have been applied to semantic segmentation \cite{mossina2024conformal} through region-pooled calibration \cite{brunekreef2024kandinsky}, image-level guarantees \cite{consign2026} and improved image-conditional coverage \cite{coat2026}, and to object detection \cite{andeol2023confident,timans2024adaptive}. Extensions support input-adaptive and non-monotone losses \cite{blot2025aacrc,angelopoulos2025ltt}. These methods calibrate for a fixed label set. For zero-shot CLIP classifiers, Conf-OT \cite{silva2025confot} transports calibration and test scores jointly. A recent study \cite{uncv2026} adds up to $750$ distractor names to CLIP's candidate list, measures the loss of coverage and restores it by recalibrating. Work on temperature scaling across label sets \cite{tu2024calibrating} and calibration for novel names after fine-tuning \cite{wang2024ovcalib} addresses calibration error rather than a risk bound. We show that calibrating at the largest vocabulary does not cover its subsets (\cref{sec:adaptive}) and construct scores that do not change when names are added.

\paragraph{Changing distributions and label sets.}
Conformal methods for non-exchangeable or drifting data \cite{farinhas2024nonexchangeable,promptshift2026,anytime2026} and for label shift \cite{podkopaev2021labelshift,si2024pac} reweight calibration data to account for changes in images or class proportions. Vocabulary growth instead changes the score function itself. A selection rule keeps the conformal guarantee if it is applied in the same way to calibration and test points \cite{jin2025focal,bao2024selective}. Calibrating for the worst member of a fixed family covers every member \cite{bates2021rcps,angelopoulos2024crc}, but choosing per point among separately calibrated sets can break coverage \cite{hegazy2025valid}. With candidate-independent scores, the worst case over vocabularies is the loss at the source vocabulary. Concurrent work \cite{wang2026cosmo} studies a growing hazard vocabulary with vision--language embedding distances, which are candidate-independent. We convert released segmenters whose scores depend on the vocabulary. Xie \etal \cite{xie2025newlabels,xie2025openset} consider classes unseen at calibration that appear in test data. Here the deployer adds the labels. Our added-label guarantee treats labels as exchangeable units, like the tasks in conformal few-shot learning \cite{fisch2021fewshot} and class pooling in clustered conformal prediction \cite{ding2023classconditional}. We also test weighted conformal prediction \cite{tibshirani2019weighted} when added labels differ.

\paragraph{Candidate dependence in open-vocabulary models.}
In vision--language models, a class score depends on the other classes queried. Accuracy falls as the vocabulary grows \cite{ren2023openness,miller2024openset}, and tuning a model for openness reduces the drop \cite{chen2026svft}. In segmentation, results change with the evaluation vocabulary and the class names \cite{huang2025revisit,huang2024renovate}. Several methods prune large vocabularies to improve accuracy \cite{freecp2025,s2c2seg2026}, as CAT-Seg does within each crop, and \v{S}ari\'c \etal \cite{whatholdsback2025} analyse FC-CLIP's classification, including its learned void class.

\paragraph{Scoring against a fixed reference.}
Normalising a score against a fixed reference set is standard in speaker verification \cite{reynolds2000ubm,auckenthaler2000tnorm}. It has also been applied to cross-modal retrieval \cite{bogolin2022qbnorm} and used with CLIP for out-of-distribution detection \cite{jiang2024neglabel}, score calibration \cite{bsap2024} and 3D segmentation \cite{engel2026querytime}. Queries that attend to a shared context but not to other classes' queries appear in SegPrompt \cite{zhu2023segprompt} and Perceiver IO \cite{jaegle2022perceiverio}, and CLIPSeg \cite{luddecke2022clipseg} segments each query independently. In CAT-Seg and SED, attention across labels is linear, so a fixed bank reduces to one cached sum per layer. We use the resulting scores for calibration, not to improve accuracy.

\section{Risk control under a growing vocabulary}
\label{sec:setting}

\subsection{Prediction sets and conformal risk control}
\label{sec:crc}

A segmenter deployed with vocabulary $\Vdep$ assigns to each pixel $p$ of an image $x$ and each label $c\in\Vdep$ a score $\score_{\Vdep}(x,p,c)$. The score is \emph{candidate-independent} if $\score_{\Vdep}(x,p,c) = \score(x,p,c)$ for every vocabulary $\Vdep$ that contains $c$, that is, if it does not depend on the other requested labels. Given a threshold $t$, the prediction set at a pixel contains the labels that score at least $t$, $C_t(x,p) = \{c\in\Vdep : \score_{\Vdep}(x,p,c) \ge t\}$. We write $C^{\Vdep}_t(x,p)$ when the vocabulary matters. All labels share the threshold $t$ \cite{sadinle2019least}. Set constructions that normalise over the vocabulary \cite{romano2020aps,angelopoulos2021raps} make sets depend on it even when the scores do not (\cref{sec:constructions}).

Let $y_p$ be the annotated label of pixel $p$, $\Vsrc$ the \emph{source vocabulary} at which the threshold is calibrated, and $P_x$ the annotated pixels of $x$ whose label lies in $\Vsrc$ (images with none are excluded). The loss of $x$ at threshold $t$ is the fraction of these pixels whose true label is missing from the set,
\begin{equation}
  L_t(x) = \frac{1}{|P_x|}\sum_{p\in P_x} \ind\big[y_p \notin C_t(x,p)\big].
  \label{eq:loss}
\end{equation}
It is non-decreasing in $t$, since a higher threshold yields smaller sets. Given $n$ calibration images, CRC \cite{angelopoulos2024crc} selects the largest threshold $\hat t$ on a fixed grid with $\big(\sum_{i=1}^n L_{t}(x_i) + 1\big)/(n+1) \le \alpha$. If no grid point qualifies, it selects the lowest one, at which every label is included (the \emph{all-label endpoint}). If the test image $x_{n+1}$ is exchangeable with the calibration images (equally likely in any order), then $\mathbb E[L_{\hat t}(x_{n+1})] \le \alpha$. We call $\mathbb E[L_{\hat t}]$ the \emph{risk}. The guarantee holds on average over calibration and test images, and counts only pixels of labels in $\Vsrc$. We evaluate the loss on a random sample of pixels per image, drawn independently of the model. CRC on these samples still bounds the full-image risk over images with a sampled source pixel (\cref{sec:sampling}).

\subsection{Reusing a threshold after vocabulary growth}
\label{sec:transfer}

The threshold $\hat t$ is calibrated with the scores $\score_{\Vsrc}$. At deployment the vocabulary grows to $\Vdep\supseteq\Vsrc$, the \emph{grown vocabulary}, and the scores become $\score_{\Vdep}$. The labels of $\Vdep$ outside $\Vsrc$ are the \emph{added labels}. Growth affects the loss \eqref{eq:loss} in two ways. First, added labels enter the prediction sets. This only enlarges the sets, so it cannot increase the loss on source labels. Second, the model now sees more candidates, so the score of a source label can change and its true label can cross the threshold in either direction. All models share the first effect. The second is the failure we study, and candidate-independent scores remove it:
\begin{proposition}[Transfer of the guarantee]
\label{prop:transfer}
Let $\score$ be candidate-independent and let $\hat t$ be calibrated by CRC at $\Vsrc$ on images exchangeable with the test image. For any $\Vdep\supseteq\Vsrc$, including one chosen from the test image, $C^{\Vdep}_{\hat t}(x,p)\cap\Vsrc = C^{\Vsrc}_{\hat t}(x,p)$ at every pixel. Hence the loss \eqref{eq:loss} is identical under $\Vsrc$ and $\Vdep$, and $\mathbb E[L_{\hat t}(x_{n+1})]\le\alpha$ holds under $\Vdep$.
\end{proposition}
Whether a source label enters the set depends only on $\score(x,p,c)\ge\hat t$, which does not involve $\Vdep$, so the guarantee follows directly from CRC. Candidate-dependent scores can also support per-image vocabularies when the selection rules are fixed before calibration:
\begin{proposition}[Per-image vocabularies]
\label{prop:adaptive}
Let $r_1,\dots,r_K$ be rules fixed before calibration. Rule $r_k$ maps an image $x$ and randomness $U$, drawn independently for each image, to a vocabulary $r_k(x,U)\supseteq\Vsrc$. Let $L^V_t(x)$ be the loss \eqref{eq:loss} under $\score_V$. CRC applied to $M_t(x)=\max_k L^{r_k(x,U)}_t(x)$ yields a threshold $\hat t_{\max}$ satisfying $\mathbb E[L^{r_{k(x)}(x,U)}_{\hat t_{\max}}(x)]\le\alpha$ for every choice $k(x)$ of a rule per image, provided calibration and test examples are exchangeable.
\end{proposition}
The loss $M_t$ is bounded, non-decreasing and at least as large as every selected loss, so CRC applies \cite[Sec.~4.4]{angelopoulos2024crc} (see also \cite[Sec.~6.3]{bates2021rcps}). This requires running every rule on the calibration images. Alternatively, a single fixed rule can be rerun on the calibration images \cite{jin2025focal}, but changing the rule later requires recalibration. With candidate-independent scores, the maximum over supersets equals $L^{\Vsrc}_t$, so calibration at the source vocabulary covers rules chosen later.

If adding names can only lower source-label scores, calibration at a \emph{lexicon} $V_{\max}$ (a large fixed name list) also covers every vocabulary between $\Vsrc$ and $V_{\max}$. A single softmax over fixed CLIP logits has this property \cite{uncv2026}. The segmenters studied here do not, since cross-label attention and FC-CLIP's void term can also raise scores (\cref{sec:mechanism}).

\section{Why calibration fails}
\label{sec:problem}

\subsection{Measured failure}
\label{sec:failure}
\begin{figure}[t]
  \centering
  \includegraphics[width=\linewidth]{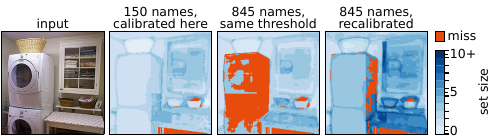}
  \caption{\textbf{Same threshold before and after growth} (CAT-Seg ViT-L, $\alpha=10\%$, median split, selected ADE20K image). Blue: labels per pixel. Orange: missed source-labelled pixels. From left to right, the maps miss $\QualSourceMiss$, $\QualReusedMiss$ and $\QualRecalMiss\%$ of them in this image and $\QualPoolSourceMiss$, $\QualPoolReusedMiss$ and $\QualPoolRecalMiss\%$ over \QualPoolImages\ of the split's test images.}
  \label{fig:problem}
\end{figure}

On the five released models and three benchmarks of \cref{sec:results}, we calibrate a CRC threshold at the source vocabulary for $\alpha=10\%$ and \emph{reuse} it at the grown one instead of \emph{recalibrating}. For the three CAT-Seg models and FC-CLIP, the risk exceeds the target on each of \NumSplits\ splits on all benchmarks ($\CocoVitLBankNativeFrozenRisk$--$\CocoFcclipBankNativeFrozenRisk\%$; \cref{fig:teaser}). SED fails in the other direction. Its reused threshold is over-conservative, with $\SedBankNativeFrozenSize$ labels per pixel on ADE20K where recalibration needs $\SedBankReplaySize$.

\paragraph{The outputs give little sign of the failure.}
Without labels at deployment, the prediction sets are the only signal. For CAT-Seg ViT-L on ADE20K they grow from $\VitLBankSrcSize$ to $\VitLBankNativeFrozenSize$ labels per pixel while the risk rises to $\VitLBankNativeFrozenRisk\%$. Meeting the target would need $\VitLBankReplaySize$ (\cref{fig:problem}). In the \SilentNUnder\ combinations that exceed the target, the sets gain only $\SilentShareLo$--$\SilentShareHi\%$ of the labels that meeting the target would need. SED's oversized sets give no sign that they are too large. Detecting either failure requires labelled images at the grown vocabulary, the same data that recalibration needs.

\subsection{Sources of candidate dependence}
\label{sec:mechanism}

\paragraph{Cost aggregation: CAT-Seg and SED.}
CAT-Seg \cite{cho2024catseg} computes a cost volume between CLIP image patches and the text embedding of every label. It refines this volume by alternating two steps: \emph{spatial} aggregation, a Swin block \cite{liu2021swin} applied to each label separately, and \emph{class} aggregation, a transformer layer over the labels at each position. Class aggregation uses linear attention \cite{katharopoulos2020linear}: with the feature map $\phi(q)=\mathrm{elu}(q)+1$, the output for label $\ell$ at a position is
\begin{equation}
  \mathrm{Att}(\ell) = \frac{\phi(q_\ell)^\top \sum_{j\in\Vdep\cup\Pi}\phi(k_j)\,v_j^\top}{\phi(q_\ell)^\top \sum_{j\in\Vdep\cup\Pi}\phi(k_j)},
  \label{eq:linear}
\end{equation}
where $q_\ell$, $k_j$ and $v_j$ are the query, key and value vectors of label tokens $\ell$ and $j$ (spatial and layer indices omitted), and $\Pi$ is a set of learned padding tokens that fill the vocabulary up to $256$ labels. The sums run over the requested vocabulary $\Vdep$, so the output for each label depends on every other label. The decoder then processes each label's map separately and outputs a per-label logit, trained with a sigmoid loss. CAT-Seg's code adds a second source of dependence. When $|\Vdep|>256$, each image crop keeps only the $256$ labels with the highest initial score and assigns the others a logit of $-100$. SED \cite{xie2024sed} follows the same design, with a ConvNeXt backbone \cite{liu2022convnext} and linear class attention \eqref{eq:linear} in each of three decoder stages, but without padding or pruning. Its optional fast mode adds \emph{category early rejection}: each stage keeps only the labels in the top $K$ at some pixel ($K=8$).

\paragraph{Mask classification: FC-CLIP.}
FC-CLIP \cite{yu2023fcclip} predicts $N$ mask proposals $m_q$, each with two logit vectors over the labels: one from its trained classifier and one from mask-pooled CLIP features. The score of label $c$ at pixel $p$ is $\sum_q P(c\mid q)\,m_q(p)$, where $P(c\mid q)$ depends on the vocabulary in three places. Both logit vectors are normalised by a softmax over $\Vdep$, their weighted geometric mean is normalised again, and the result is multiplied by one minus a void probability, from a softmax over $\Vdep\cup\{\text{void}\}$ \cite{whatholdsback2025}. The masks and per-query logits do not depend on other labels.

\paragraph{Direction of the effect.}
Attention over more labels can move source-label scores in either direction. Discarding a label lowers its score, and a single softmax over fixed logits lowers existing probabilities when names are added. FC-CLIP's void probability also falls, which raises every score, and the combined effect need not be monotone. The mechanisms therefore predict that pruning and a single softmax tend to raise the risk, while cross-label attention can also make a reused threshold over-conservative. Two controls on ADE20K support this. Removing the pruning from CAT-Seg ViT-B lowers the risk of the reused threshold from $\AuditNativeFrozenRisk\%$ to $\AuditUntruncatedFrozenRisk\%$ (\AuditImages\ images). SED's fast mode, whose early rejection discards labels, raises it from $\SedNativeFrozenRisk\%$ to $\SedFastNativeFrozenRisk\%$. Because scores can also rise, recalibrating at a larger vocabulary need not cover the vocabularies within it (\cref{sec:transfer}). With per-image vocabularies, SED's threshold recalibrated at $845$ names reaches $\AdaptSedAdvRecalRisk\%$ risk, and $\AdaptSedAdvLexRisk\%$ when recalibrated at a $\AdaptLexFourNames$-name lexicon (\cref{sec:adaptive}).

\section{Candidate-independent scores from released models}
\label{sec:method}

\subsection{Reference-anchored conversion}
\label{sec:conversion}

The conversion keeps the released weights and changes only the set over which the sums in \eqref{eq:linear} run. Fix a \emph{reference bank} $\bank$ of label names. For each image, we first run the bank through the network with pruning disabled. At every class-attention layer we record the key--value sum $S_{kv}=\sum_{j\in\bank\cup\Pi}\phi(k_j)v_j^\top$ and the key sum $S_k=\sum_{j\in\bank\cup\Pi}\phi(k_j)$ over the bank and padding tokens ($\Pi$ is empty for SED and for banks of $256$ or more names). Each candidate $c$ is then run through the same layers with the attention output
\begin{equation}
  \mathrm{Att}(c) = \frac{\phi(q_c)^\top\big(S_{kv} + \phi(k_c)v_c^\top\big)}{\phi(q_c)^\top\big(S_k+\phi(k_c)\big)}.
  \label{eq:anchored}
\end{equation}
The candidate attends to the bank, the padding and itself. The bank does not attend to the candidate, and candidates do not attend to one another. (A candidate that is also a bank name attends to both its bank token and itself.) The other components (spatial aggregation, decoder and per-label logit) already treat each label separately. We do not apply the pruning to $256$ labels. The anchored score $\score(x,p,c)$ therefore depends only on the image, the candidate and $\bank$. It is candidate-independent, so \cref{prop:transfer} applies. When a tagger or language model produces the names, adding the source names to its output ensures $\Vdep\supseteq\Vsrc$.

\paragraph{FC-CLIP.}
The same idea applies to the softmax normalisations. For a label $c$, every softmax over $\Vdep$ becomes a softmax over $\{c\}\cup\bank$ (with $c$ removed from the bank if it is a bank entry), and the void probability becomes a softmax over $\bank\cup\{\text{void}\}$. Masks and per-query logits are unchanged. For a label that is a bank entry with a single name, this reproduces the released pipeline run on the vocabulary $\bank$.

\paragraph{Implementation and cost.}
Our reimplementation of the native forward pass matches the released code (maximum absolute logit difference $\CheckCatsegReimpl$ for CAT-Seg and $\CheckSedReimpl$ for SED). Anchored scores change by no more than this when the other candidates or their order change. Letting the bank also attend to the candidate needs one native pass on the bank plus each candidate. It costs $\AuditBankPlusOneCostRatio$ times as much as anchored scoring and changes the logits by only $\AuditBankPlusOneDiff$ on average (ViT-B, COCO-171 bank; banks in \cref{sec:bankchoice}). On a GH200 with the default banks, the conversion takes $\RtVitLBankRatio\times$ the time of one native pass at $845$ labels for CAT-Seg ViT-L, $\RtSedBankRatio\times$ for SED and at least $\RtFcclipBankRatioMin\times$ for FC-CLIP. Stopping the bank pass once the key and value sums are recorded gives the same scores and, on an RTX 5070 Ti, lowers the ViT-L ratio from $\RtLocalVitLBankRatio$ to $\RtLeanVitLBankRatio\times$ (\cref{sec:runtime}).

\subsection{The reference bank}
\label{sec:bankchoice}
The bank sets the context in which each candidate is scored. In our experiments, banks work best when they contain the training vocabulary plus enough other names that a candidate does not dominate its own attention (\cref{sec:bank}). For CAT-Seg and SED we start from the $171$ COCO-Stuff names on which both models were trained (the \emph{COCO-171 bank}). From the Open Images class list \cite{kuznetsova2020openimages} we remove every name that matches an ADE20K or Pascal Context label and add $512$ names drawn at random from the rest (the \emph{$683$-name bank}). For FC-CLIP the default bank is its training vocabulary (the $133$ COCO panoptic classes) merged with the $683$-name bank, giving $701$ names. Some bank names coincide with requested labels: $\BankOverlapSource$ of the $150$ ADE20K source names are in the COCO-171 part, and $\BankOverlapOiCocoAdded$ of the Open Images names are COCO-Stuff added labels. The guarantees hold for any bank chosen without the calibration and test images. We chose ours on $\DisjBankImages$ ADE20K validation images. The other \DisjImages, Pascal Context, COCO-Stuff and the reserved images (\cref{sec:reserve}) test it independently.

\subsection{Guarantees for the added labels}
\label{sec:addedtheory}

\Cref{prop:transfer,prop:adaptive} concern the source labels. To obtain a guarantee for unannotated added labels, we need an assumption linking their losses to those of the source labels. Fix the calibration images and define the \emph{label loss} $\ell_c(t)$ as the fraction of the pixels with true label $c$ whose score for $c$ is below $t$. For a candidate-independent score, $\ell_c$ depends on the label and the images but not on the vocabulary in which $c$ is requested. Labels can then take the place of images as the exchangeable units of calibration:
\begin{proposition}[Added labels]
\label{prop:labels}
Let $\score$ be candidate-independent, and let the source labels $c_1,\dots,c_m$ and an added label $c_{m+1}$, all present in the fixed calibration images, be exchangeable, for instance because $c_1,\dots,c_{m+1}$ are drawn uniformly without replacement from the requested names present in these images. The label-level threshold $\hat t_\ell$, the largest grid threshold with $\big(\sum_{i\le m}\ell_{c_i}(t)+1\big)/(m+1)\le\alpha$ (the all-label endpoint if none), satisfies $\mathbb E[\ell_{c_{m+1}}(\hat t_\ell)]\le\alpha$.
\end{proposition}
This is CRC applied to the label losses (\cref{sec:labelsproof}). The expectation is over the label draw with the images fixed, so the bound is for the calibration images, not new ones. A label absent from the calibration images is not covered. If each label comes with its own sample of images, drawn the same way for every label, the same construction also bounds an added label's expected miss rate on a new image of it (\cref{prop:newimages,sec:newimagesproof}). The label-level threshold needs annotations of the source labels only. Deploying $\min(\hat t,\hat t_\ell)$, with $\hat t$ the \emph{image-level} threshold of \cref{sec:crc}, keeps both guarantees. For candidate-dependent scores, the source-label losses are computed at $\Vsrc$ and the added label's at $\Vdep$, so exchangeable names do not give exchangeable losses. When source and added labels come from different distributions, their gap in difficulty bounds the excess risk:
\begin{proposition}[Harder added labels]
\label{prop:shift}
Let $\score$ be candidate-independent, let the source labels be drawn i.i.d.\ from a distribution $P$ and the added label independently from $Q$, both over names present in the calibration images, and let $\Delta(t)=\bar\ell_Q(t)-\bar\ell_P(t)$, where $\bar\ell_P(t)=\mathbb E_{c\sim P}[\ell_c(t)]$ and $\bar\ell_Q(t)=\mathbb E_{c\sim Q}[\ell_c(t)]$. Then $\mathbb E[\ell_{c_{m+1}}(\hat t_\ell)]\le\alpha+\mathbb E[\Delta(\hat t_\ell)]\le\alpha+\max_t\Delta(t)$.
\end{proposition}
Since the added label is independent of $\hat t_\ell$, its risk is $\mathbb E[\bar\ell_Q(\hat t_\ell)]=\mathbb E[\bar\ell_P(\hat t_\ell)]+\mathbb E[\Delta(\hat t_\ell)]$, and the first term is the risk of a further draw from $P$, which is at most $\alpha$ by \cref{prop:labels}. If a fraction $\varepsilon$ of the added labels comes from an arbitrary distribution and the rest from $P$, the excess is at most $\varepsilon$. If unannotated added labels can be adversarial, any valid method, recalibration included, selects the all-label endpoint with probability at least $1-\alpha$ (\cref{prop:impossible}).

\paragraph{Obtaining exchangeable labels.}
Dense annotation takes more than $1.5$ hours per image for the $30$ classes of Cityscapes \cite{cordts2016cityscapes}, so large vocabularies are annotated only in part. When the names that may be requested are known (a taxonomy, a product catalogue), annotating a uniformly random subset makes the remaining names exchangeable with the annotated ones. The guarantee is then an average over the remaining names. LVIS and Open Images let annotators or a classifier choose which names to annotate \cite{gupta2019lvis,kuznetsova2020openimages}. For such labels, the gap bound of \cref{prop:shift} applies (\cref{sec:hardness}).

\section{Experiments}
\label{sec:results}

\paragraph{Benchmarks.}
On ADE20K \cite{zhou2017ade20k} the source vocabulary is the $150$-class benchmark, and the grown one adds the $695$ classes of ADE20K-847 not among them, in a fixed nested order. On Pascal Context \cite{mottaghi2014pascalcontext} the vocabulary grows from the $59$-class benchmark to the $458$ named classes of its $459$-class version. On COCO-Stuff \cite{caesar2018cocostuff} the source vocabulary is the $171$ classes on which CAT-Seg and SED were trained, and the grown one adds the $1{,}112$ LVIS \cite{gupta2019lvis} categories that are not COCO-Stuff concepts (\cref{sec:cocoall}). We use $\NumAdeImages$ validation images per benchmark and the official class names.

\paragraph{Models.}
We evaluate five released models and their anchored conversions: CAT-Seg with ViT-B and ViT-L \cite{dosovitskiy2021vit} CLIP backbones, CAT-Seg with the DeCLIP \cite{wang2025declip} EVA02-L \cite{fang2024eva02} backbone, SED ConvNeXt-L, and FC-CLIP ConvNeXt-L. Inference follows the released pipelines (\cref{sec:miouproto}). Unless stated otherwise, the anchored models use the $683$-name bank ($701$ names for FC-CLIP), chosen on ADE20K and kept fixed for Pascal Context and COCO-Stuff.

\paragraph{Protocol and metrics.}
We store scores at $4{,}096$ uniformly sampled pixels per image. Each of \NumSplits\ random splits divides the images into equal calibration and test halves. CRC chooses the threshold on the calibration half at $\alpha=10\%$, and we evaluate it on the test half. At the grown vocabulary we report the \emph{risk} (the loss \eqref{eq:loss} averaged over splits) and the \emph{set size} (the mean number of labels per pixel). We compare three thresholds: the reused native one, the recalibrated native one, which is the reference for set size (\cref{sec:failure}), and our conversion calibrated once at the source vocabulary (\emph{anchored}). The \emph{overhead} of the conversion is the difference in set size between the anchored and recalibrated native models. Thresholds are image-level (\cref{sec:crc}) except in \cref{sec:labelguarantee}. We developed the method on the validation images. \Cref{sec:reserve} confirms the results on images not used in development. Code, manifests and results are available at \url{https://github.com/munim110/open-vocab-crc} (\cref{sec:repro}).

\subsection{Anchored models after growth}
\label{sec:main}

\begin{figure*}[t]
  \centering
  \includegraphics[width=\linewidth]{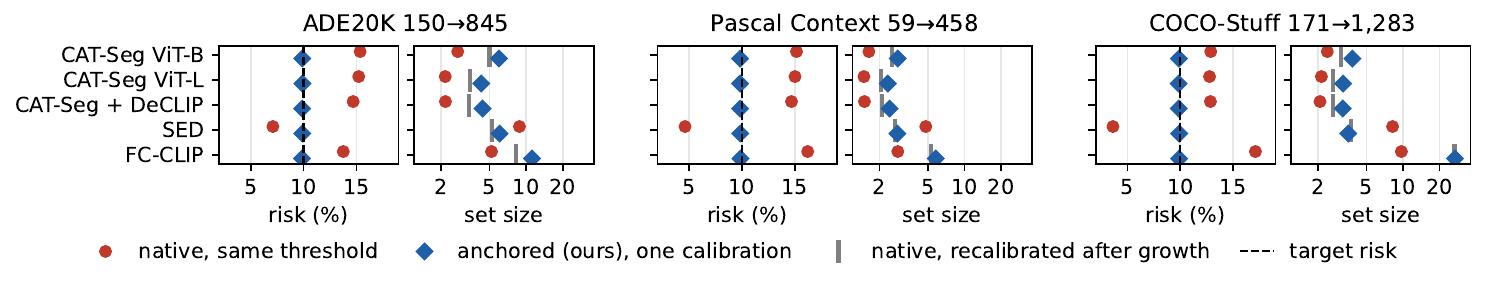}
  \caption{\textbf{The anchored models meet the target in every case.} Risk and set size after growth, as in \cref{fig:teaser}. Blue: our anchored models, with the threshold from the source vocabulary. Grey: native recalibration at the grown vocabulary. The values, in labels per pixel, are in \cref{tab:main}.}
  \label{fig:anchored}
\end{figure*}

With the source-vocabulary threshold, the anchored models meet the target for every model and benchmark (\cref{fig:anchored}; images in \cref{fig:qual,fig:qualpanel}). In each of the \SilentNUnder\ combinations where the native threshold fails, the anchored sets grow by $\SilentGrowthAnchLo$--$\SilentGrowthAnchHi\times$ with the vocabulary and the native sets by $\SilentGrowthReusedLo$--$\SilentGrowthReusedHi\times$, so growth of the sets does not show that the target is met. The overhead is at most $\OverviewCostAggPriceMax$ labels per pixel for CAT-Seg and SED, and on COCO-Stuff SED's anchored sets are smaller than the recalibrated ones. FC-CLIP's overhead is $\FcclipBankPrice$ labels on ADE20K (\cref{tab:fcclip}), and on COCO-Stuff its sets are large with or without the conversion. Models that are candidate-independent by design need $\ClipsegFull$ (CLIPSeg) and $\VitBSclipFull$ (SCLIP \cite{wang2024sclip}) labels per pixel at $845$ names (\cref{tab:indep}). Claude Opus 5.5, asked to name marked pixels, omits the true label so often that CRC includes all $845$ names (\cref{sec:mllm}).

\subsection{Vocabularies chosen per image}
\label{sec:adaptive}

\begin{table}[t]
  \centering\footnotesize
  \caption{\textbf{Per-image vocabularies} ($\alpha=10\%$): risk (\%) / labels per pixel under each image's worst (highest-loss) rule (last block: a language model's names). \emph{Calibration needs}: names known in advance (\emph{all}: every grown name; \emph{17k}: a $\AdaptLexFourNames$-name lexicon). \emph{random}: calibrated on source names plus $500$ (ADE20K) or $300$ (Pascal) random grown names. \emph{rules}: calibrated under the worst rule or on those names. --: not run. All rules: \cref{tab:adaptive,tab:adaptivesed,tab:adaptivefcclip,tab:adaptivepascal}.}
  \label{tab:adaptivemain}
  \resizebox{\columnwidth}{!}{\setlength{\tabcolsep}{3pt}\begin{tabular}{@{}lrrrrrr@{}}
\toprule
& \multicolumn{1}{c}{anchored} & \multicolumn{5}{c}{native} \\ \cmidrule(lr){2-2}\cmidrule(l){3-7}
& \multicolumn{1}{c}{once} & \multicolumn{1}{c}{reused} & \multicolumn{1}{c}{recal.} & \multicolumn{1}{c}{random} & \multicolumn{1}{c}{recal.} & \multicolumn{1}{c}{rules} \\
\emph{calibration needs} & \multicolumn{1}{c}{$\Vsrc$} & \multicolumn{1}{c}{$\Vsrc$} & \multicolumn{1}{c}{all} & \multicolumn{1}{c}{all} & \multicolumn{1}{c}{17k} & \multicolumn{1}{c}{all rules} \\
\midrule
\multicolumn{7}{@{}l}{\emph{ADE20K, $150\to845$}} \\
CAT-Seg ViT-L & 9.9/4.2 & 14.9/2.2 & 9.5/3.6 & 10.4/3.3 & 4.1/8.9 & 9.9/3.4 \\
SED & 9.9/5.9 & 12.8/3.4 & 18.8/2.0 & 18.4/2.1 & 25.4/1.4 & 9.9/5.0 \\
FC-CLIP & 9.9/9.7 & 14.1/4.8 & 10.3/7.4 & 10.8/6.9 & 5.6/17.4 & 9.9/7.7 \\
\midrule
\multicolumn{7}{@{}l}{\emph{Pascal Context, $59\to458$}} \\
CAT-Seg ViT-L & 9.8/2.3 & 15.2/1.5 & 10.0/2.0 & 10.9/1.9 & -- & 9.8/2.1 \\
SED & 9.8/2.8 & 12.3/1.9 & 21.1/1.1 & 19.8/1.2 & -- & 9.8/2.4 \\
\midrule
\multicolumn{7}{@{}l}{\emph{ADE20K, names from a language model}} \\
CAT-Seg ViT-L & 9.9/4.6 & 20.6/2.7 & 12.6/4.1 & -- & -- & 9.9/4.8 \\
SED & 9.9/5.7 & 20.5/3.4 & 30.3/2.3 & -- & 41.2/1.5 & 9.9/6.4 \\
FC-CLIP & 9.9/8.0 & 21.5/4.4 & 15.6/5.9 & -- & 7.8/11.0 & 9.9/8.8 \\
\bottomrule
\end{tabular}}
\end{table}

\Cref{prop:transfer} covers names added to $\Vsrc$ for each test image by any process, even one that cannot be rerun on the calibration images. On ADE20K, we add to the $150$ source names of each image one of: the names annotated in that image (a perfect tagger), $50$, $200$ or $500$ random names, the $50$ names the native model scores highest (a tagger sharing the model's confusions), or the worst of these for that image. A language model listing what it sees, mostly names outside the $845$, is a further rule (last block of \cref{tab:adaptivemain}; \cref{sec:freeform}). For every rule and every evaluated model, the anchored threshold has $\AdaptAdvAnchRisk\%$ risk on ADE20K and $\AdaptPasAdvAnchRisk\%$ on Pascal Context (\cref{tab:adaptivemain}; Pascal rules in \cref{sec:adaptivepascal}). Under the worst rule the reused native threshold exceeds the target for every model, including SED (over-conservative in \cref{sec:failure}). Native calibration also meets the target when it uses the deployed rule, or the worst case over the rules (\cref{prop:adaptive}). Without knowing the rule, every native calibration we tested exceeds the target for at least one model (\cref{sec:mechanism,sec:adaptivesed}; by deployment in \cref{tab:regimes}). With the language model's names, recalibration at all $845$ names exceeds it for every model ($\FfRecalRiskLo$--$\FfRecalRiskHi\%$).

\subsection{A guarantee for the added labels}
\label{sec:labelguarantee}

\begin{table*}[t]
  \centering\small
  \caption{\textbf{Added-label risk} on the calibration images (\%) / set size at the grown vocabulary (CAT-Seg ViT-L, \LxTrials\ trials). A dash means not run. Each trial draws a calibration half and, for \emph{random draw}, a source vocabulary (on ADE20K one of \LxAdeSets\ fixed draws; \cref{tab:labelmodels} redraws it).}
  \label{tab:labelexch}
  \begin{tabular}{@{}lcrrrr@{}}
\toprule
& & \multicolumn{2}{c}{anchored} & \multicolumn{2}{c}{native, label-level} \\
\cmidrule(lr){3-4}\cmidrule(l){5-6}
& & \multicolumn{1}{c}{image-level} & \multicolumn{1}{c}{label-level (ours)} & \multicolumn{1}{c}{recal.\ at grown} & \multicolumn{1}{c}{reused from source} \\
Source vocabulary & $\alpha$ & \multicolumn{4}{c}{added-label risk (\%) / set size} \\
\midrule
\multicolumn{6}{@{}l}{\emph{random draw}} \\
ADE20K $150\!\to\!845$ & .10 & 23.7 / 6.8 & 9.1 / 28.9 & 9.2 / 35.8 & 20.8 / 9.7 \\
ADE20K $150\!\to\!845$ & .05 & 14.3 / 15.0 & 4.5 / 74.5 & 4.3 / 228.1 & 11.8 / 24.3 \\
Pascal Context $59\!\to\!458$ & .10 & 16.7 / 3.7 & 7.6 / 11.5 & 7.6 / 14.9 & -- \\
Pascal Context $59\!\to\!458$ & .05 & 9.8 / 7.1 & 2.8 / 68.4 & 2.7 / 92.8 & -- \\
COCO $171\!\to\!1{,}283$ & .10 & 17.7 / 5.7 & 8.9 / 11.9 & 9.0 / 13.6 & -- \\
COCO $171\!\to\!1{,}283$ & .05 & 10.3 / 10.4 & 4.0 / 30.2 & 4.0 / 42.1 & -- \\
\midrule
\multicolumn{6}{@{}l}{\emph{benchmark's own}} \\
ADE20K $150\!\to\!845$ & .10 & 34.2 / 4.5 & 22.2 / 9.0 & 22.6 / 10.1 & 33.8 / 5.1 \\
Pascal Context $59\!\to\!458$ & .10 & 24.7 / 2.5 & 22.3 / 2.7 & 26.5 / 2.6 & 39.9 / 1.6 \\
COCO $171\!\to\!1{,}283$ & .10 & 32.1 / 3.3 & 17.6 / 6.3 & 21.2 / 5.6 & 29.2 / 3.9 \\
\bottomrule
\end{tabular}
\end{table*}

\Cref{tab:labelexch} tests \cref{prop:labels}. The \emph{added-label risk} is the label loss $\ell_c(\hat t_\ell)$ averaged over added labels, whose benchmark annotations are used for evaluation only. The source vocabulary is a random subset of the grown one (exchangeable labels) or the benchmark's own classes.

\paragraph{Random source vocabulary.}
The image-level threshold of \cref{sec:main} misses $\LxAdeImgRisk\%$ of an added label's pixels on ADE20K, with $\LxAdeImgSize$ labels per pixel where the label-level threshold needs $\LxAdeLabSize$. When the source vocabulary is redrawn in every trial (\cref{tab:labelmodels}), the label-level threshold on the anchored scores keeps the added-label risk on the calibration images below the target for all five models on all three benchmarks ($\LxModelsMinRisk$--$\LxModelsMaxRisk\%$; $\alpha=5\%$ and held-out images in \cref{sec:labelmodels}). The native model, calibrated the same way and deployed at $845$ names, misses $\LxAdeFrozenRisk\%$, since its label losses change with the vocabulary. Native label-level CRC at the grown vocabulary is valid but needs that vocabulary in advance and gives larger sets ($\LxAdeRecalSize$ labels per pixel).

\paragraph{The benchmark's own source classes.}
These are the frequent classes, and label-level CRC leaves an added-label risk of $\LxCocSplitLabRisk$--$\LxPasSplitLabRisk\%$. The decomposition of \cref{prop:shift} explains the excess. Source labels stay below the target, while added labels have a loss higher by $\Delta(\hat t_\ell)=\LxDCocSplitLabDelta$--$\LxDPasSplitLabDelta$ points, against $|\Delta(\hat t_\ell)|\le\LxDExchDeltaMax$ with a random source vocabulary. Annotating $100$ random added names on the calibration images restores exchangeability and keeps the ADE20K added-label risk at $\LsAdeVitLMHundredRisk\%$ for ViT-L and $\LsAdeSedMHundredRisk\%$ for SED (\cref{sec:weights}).

\subsection{Open-vocabulary detection}
\label{sec:detection}
We also convert three Grounding DINO detectors \cite{liu2024groundingdino,zhao2024mmgdino} on COCO \cite{lin2014coco} $80\!\to1{,}204$ (COCO $\cup$ LVIS v1). OWLv2 \cite{minderer2023owlv2}, candidate-independent by design, is a control. In the anchored detector, the image branch sees only an $80$-name COCO bank, and each candidate is scored against the object queries computed from the bank. At LVIS scale the reused native threshold meets the target for every detector (\cref{tab:ovd}). At $\alpha=10\%$ anchored sets are no larger than those of native recalibration, and inference is $\OvdRtSpeedupLo$--$\OvdRtSpeedupHi\times$ faster, as the image branch runs twice instead of once per chunk of names (\cref{tab:runtime}). Anchoring leaves more added-class objects with no overlapping box, often setting the label-level threshold to the all-label endpoint (\cref{sec:detectionfull}), and cuts their top-1 accuracy from $\OvdAccNativeLo$--$\OvdAccNativeHi\%$ to $\OvdAccAnchALo$--$\OvdAccAnchAHi\%$.

\subsection{Choice of reference bank}
\label{sec:bank}

Every bank we tried meets the target and changes only the set size (\cref{tab:bankfull}). For anchored CAT-Seg ViT-B on ADE20K (\BankSweepImages\ images), the sets contain $\BankCandidateOnlyZeroSize$ labels per pixel with the candidate alone, $\BankCocoOneSevenOneSize$ with the COCO-171 bank and $\BankCocoOiSixEightThreeSize$ with the $683$-name bank, and grow slowly with larger banks. Five redraws of the Open Images names differ by at most $\BankSeedsSizeSpread$ labels (\cref{tab:bankseeds}). On COCO-Stuff, a smaller bank gives smaller sets. Banks without the training-vocabulary names (WordNet \cite{miller1995wordnet} nouns, non-words) give larger sets.

\paragraph{Segmentation accuracy.}
On full images of ADE20K (A-150, A-847) and Pascal Context (PC-59, PC-459), anchored CAT-Seg and SED with the $683$-name bank are at most $\MiouGapHiPcFiftyNine$ mIoU points below the native models. FC-CLIP loses $\MiouGapFcclipAnchoredBankAdeOneFifty$ on A-150, $\MiouGapFcclipAnchoredBankAdeEightFortySeven$ on A-847 and none on Pascal Context (\cref{sec:miouproto}). FC-CLIP, whose class softmax runs over the bank, loses added-class accuracy. CAT-Seg and SED do not (\cref{tab:pixaccadded}).

\subsection{Confirmation on reserved images}
\label{sec:reserve}
We scored $2{,}000$ ADE20K training images not used in development, once, with CAT-Seg ViT-L, fixing the $683$-name bank and the $845$-name split in advance. The reused native threshold has $\ConfPrimaryNativeFrozenRisk\%$ risk and the anchored one $\ConfPrimaryAnchRisk\%$, with $\ConfPrimaryPrice$ more labels per pixel than native recalibration. The added-label risk with random source labels is $\LxRsvLabRisk\%$ (\cref{sec:reservefull}).

\section{Conclusion}
\label{sec:conclusion}\label{sec:limitations}
A risk threshold calibrated for a released open-vocabulary segmenter does not survive adding names, and the failure shows little in its prediction sets. Scoring each label against a fixed reference bank makes its score independent of the other names. One calibration then keeps the source-label guarantee as names are added, including names chosen per image, without retraining. Across our benchmarks, the cost at $\alpha=10\%$ is at most one extra label per pixel for CAT-Seg and SED. Anchored CAT-Seg ViT-L takes $\RtLeanVitLBankRatio\times$ the native inference time on an RTX 5070 Ti. On fixed images, calibrating over labels gives an added-label guarantee when annotated names are sampled at random. With about $20$ labelled images per name, a threshold per name, capped at the image-level threshold, bounds its miss rate after names are added, with sets $\PnHybRatioLo$--$\PnHybRatioHi\times$ larger (\cref{sec:pername}).

The source-label guarantee assumes exchangeable calibration and test images. It holds on average over images, not for each scene or each per-image vocabulary \cite{gibbs2025conditional}. The conversion covers linear attention across labels, softmaxes over the vocabulary and Grounding DINO's text--image fusion. Softmax attention across labels is untested. Native recalibration is valid when the vocabulary is fixed in advance and labelled images exist. Anchoring is for vocabularies that arrive during use, such as the names a tagger supplies for each image \cite{ren2024groundedsam,gu2024conceptgraphs,vocabfree2025}.

\section*{LLM Usage Statement}
The research question and the experimental design are the authors'. An AI coding assistant, Claude Code (Anthropic), helped write and debug the experiment and figure code, which the authors ran after review. It also helped search for and summarise related work and edited the text. A multimodal language model, Claude Opus 5.5, is also evaluated in the experiments (\cref{sec:mllm,sec:freeform}). The authors reviewed all AI-assisted work, and every number in the text is generated from stored results by the released code. We take responsibility for the final content.
{
    \small
    \nocite{chan2021smiyc,blumenstiel2023mess,yang2023som,giuffrida2022phisat,swan2021ai4mars}%
    \bibliography{refs}
}
\clearpage
\newcommand{\supptable}[1]{\adjustbox{max width=\textwidth}{\footnotesize\setlength{\tabcolsep}{4pt}\InputIfFileExists{tables/#1}{}{\errmessage{missing file tables/#1}}}}
\newcommand{\suppcoltable}[1]{\adjustbox{max width=\linewidth}{\footnotesize\setlength{\tabcolsep}{4pt}\InputIfFileExists{tables/#1}{}{\errmessage{missing file tables/#1}}}}
\newcommand{\supppair}[2]{\adjustbox{max width=\textwidth}{\footnotesize\setlength{\tabcolsep}{4pt}\adjustbox{valign=t}{\InputIfFileExists{tables/#1}{}{\errmessage{missing file tables/#1}}}\hspace{14pt}\adjustbox{valign=t}{\InputIfFileExists{tables/#2}{}{\errmessage{missing file tables/#2}}}}}
\renewcommand{\topfraction}{.9}
\renewcommand{\dbltopfraction}{.9}
\renewcommand{\textfraction}{.1}
\renewcommand{\floatpagefraction}{.8}
\renewcommand{\dblfloatpagefraction}{.8}
\setcounter{topnumber}{4}
\setcounter{dbltopnumber}{8}%
\setcounter{totalnumber}{8}
\makeatletter%
\setlength{\@fptop}{0pt}\setlength{\@dblfptop}{0pt}\setlength{\@fpsep}{18pt}\setlength{\@dblfpsep}{18pt}
\setlength{\@fpbot}{0pt plus 1fil}\setlength{\@dblfpbot}{0pt plus 1fil}\makeatother
\setlength{\dblfloatsep}{14pt plus 4pt minus 2pt}\setlength{\dbltextfloatsep}{16pt plus 4pt minus 3pt}
\maketitlesupplementary
\def\theHtable{S\arabic{table}}\def\theHfigure{S\arabic{figure}}\def\theHproposition{S\arabic{proposition}}
\appendix
\setcounter{proposition}{0}\renewcommand{\theproposition}{S\arabic{proposition}}
\setcounter{table}{0}\renewcommand{\thetable}{S\arabic{table}}
\setcounter{figure}{0}\renewcommand{\thefigure}{S\arabic{figure}}

This supplement provides proofs, implementation details and complete experimental results. \Cref{sec:sampling,sec:labelsproof,sec:newimagesproof,sec:impossibilityproof} complete the theory of \cref{sec:setting,sec:addedtheory}. \Cref{sec:impl} completes the experimental setup of \cref{sec:results}. \Cref{sec:fullresults} extends \cref{sec:main,sec:adaptive}, and \cref{sec:addedfull,sec:detectionfull,sec:bankfull,sec:reservefull} extend \cref{sec:labelguarantee,sec:detection,sec:bank,sec:reserve}; \cref{sec:beyondfull} contains the further analyses summarised at the end of \cref{sec:results}, and \cref{sec:deploy} describes deployments in which the vocabulary grows.

\section{Sampled pixels and the guarantee}
\label{sec:sampling}
This section shows that calibrating on a sample of pixels bounds the full-image risk, as stated in \cref{sec:crc}. The stored scores cover $k=4{,}096$ pixels per image, drawn uniformly without replacement. The seed of each image is its index, which does not depend on the model or the labels, and we model it as independent of the image content. Let $P_x$ be the full-image set of annotated source-label pixels, as in \eqref{eq:loss}, $S$ the sample and $m=|S\cap P_x|$. Images with no source-labelled pixel ($P_x=\emptyset$), for which \eqref{eq:loss} is undefined, are excluded. The sampled loss is
\begin{equation*}
  \tilde L_t(x,S) = \frac{1}{m}\sum_{p\in S\cap P_x} \ind\big[y_p \notin C_t(x,p)\big], \qquad m\ge 1 .
\end{equation*}

\paragraph{Unbiasedness.} Given $x$ and $m$, the set $S\cap P_x$ is uniform among the $m$-subsets of $P_x$, so $\mathbb E[\tilde L_t(x,S)\mid x,m] = L_t(x)$ for every fixed $t$.

\paragraph{Exchangeability.} $S_i$ depends only on the image size and the seed, which we model as independent of the image content. The pairs $(x_i,S_i)$ are therefore exchangeable whenever the images are. They remain exchangeable after images with $m=0$ are excluded, since the exclusion is a fixed rule applied to each pair. $\tilde L_t$ is bounded in $[0,1]$ and non-decreasing in $t$, so CRC on the calibration pairs gives $\mathbb E[\tilde L_{\hat t}(x_{n+1},S_{n+1})]\le\alpha$.

\paragraph{Full-image loss.} $\hat t$ is a function of the calibration pairs, which are independent of $S_{n+1}$ given $x_{n+1}$. Conditioning on the calibration pairs, $x_{n+1}$ and $m_{n+1}$ and applying unbiasedness gives $\mathbb E[L_{\hat t}(x_{n+1})] = \mathbb E[\tilde L_{\hat t}(x_{n+1},S_{n+1})] \le \alpha$.

The bound therefore holds for the full-image miss rate, although calibration reads only the sampled pixels. The expectation is over images with $m\ge1$, \ie the image distribution reweighted by $\Pr(m\ge1\mid x)$. With $4{,}096$ pixels no development image has $m=0$; one of the $2{,}000$ reserved images does and is excluded, leaving $\ConfImages$. Sampling adds variance to the loss of each image, which the marginal bound absorbs. The argument uses linearity in the pixels; the stricter losses of \cref{sec:losses} are not linear, so for them the guarantee holds for the losses as evaluated on the sample.

\section{Proof of the added-label guarantee}
\label{sec:labelsproof}
This section proves \cref{prop:labels}. The calibration images are fixed, and the only randomness is the draw of the labels. For a label $c$ present in these images, let $P_c$ be the pixels annotated with $c$ and
\begin{equation*}
  \ell_c(t) = \frac{1}{|P_c|}\sum_{(x,p)\in P_c} \ind\big[\score(x,p,c) < t\big].
\end{equation*}

\paragraph{Exchangeable losses.} Because $\score$ is candidate-independent, $\ell_c$ is the same function of $c$ whether $c$ is requested in $\Vsrc$ or in $\Vdep$. The loss curves $\ell_i=\ell_{c_i}$, $i\le m+1$, are therefore one fixed map applied to each coordinate of an exchangeable sequence, and they are exchangeable. This is the only step that uses candidate independence. For a candidate-dependent score, the source losses are computed with $\score_{\Vsrc}$ and the added label's with $\score_{\Vdep}$, which are two different maps, so exchangeable names do not give exchangeable losses. Each $\ell_i$ takes values in $[0,1]$, is non-decreasing in $t$, and is zero at the lowest grid point, where every label is included.

\paragraph{Risk bound.} Let $t'$ be the largest grid threshold with $\frac1{m+1}\sum_{i\le m+1}\ell_i(t)\le\alpha$; it exists because every loss is zero at the lowest grid point. Since $\ell_{m+1}\le1$, $\hat t_\ell$ satisfies this inequality, so $\hat t_\ell\le t'$; this also holds when $\hat t_\ell$ is the all-label endpoint. Since $\ell_{m+1}$ is non-decreasing, $\ell_{m+1}(\hat t_\ell)\le\ell_{m+1}(t')$. The threshold $t'$ is a symmetric function of $(\ell_1,\dots,\ell_{m+1})$, so exchangeability gives $\mathbb E[\ell_i(t')]=\mathbb E[\ell_{m+1}(t')]$ for every $i$, and
\begin{equation*}
  \mathbb E[\ell_{m+1}(\hat t_\ell)] \le \mathbb E[\ell_{m+1}(t')] = \mathbb E\Big[\frac{1}{m+1}\sum_{i\le m+1}\ell_i(t')\Big] \le \alpha .
\end{equation*}

\paragraph{Random source vocabularies.} Let $N$ be the names that will be requested and are present in the calibration images, let $\Vsrc$ be a uniformly random $m$-subset of $N$, and let $c_{m+1}$ be any added label drawn uniformly from $N\setminus\Vsrc$. Listing $\Vsrc$ in uniformly random order makes $(c_1,\dots,c_{m+1})$ a uniformly random ordered draw without replacement from $N$, which is exchangeable. The order does not matter, since $\hat t_\ell$ depends on $\Vsrc$ only as a set. A uniformly chosen added label is such a draw, so by linearity the bound also holds for the risk averaged over all added labels, which is the quantity in \cref{tab:labelexch}.

\paragraph{Scope of the assumption.} The argument uses exchangeability of the loss curves $\ell_{c_i}$, which is weaker than exchangeability of the names. Exchangeability is also not necessary for the bound. By \cref{prop:shift}, any added-label distribution with $\mathbb E[\Delta(\hat t_\ell)]\le0$, \ie one whose labels are no harder at the calibrated threshold, satisfies it as well. Without a restriction on added-label losses, uniform validity requires the all-label threshold with probability at least $1-\alpha$ (\cref{prop:impossible}).

\section{Added labels on new images}
\label{sec:newimagesproof}
\begin{proposition}[Added labels on new images]
\label{prop:newimages}
In \cref{prop:labels}, let each label come with $k$ images annotated with it, such that these units are exchangeable and, conditional on the labels, the joint distribution is invariant to permutations of images within each unit. Replace $\ell_c(t)$ by $\bar\ell_c(t)$, the mean over these images of the fraction of $c$'s pixels scored below $t$. Then an image $x$ of the added label, drawn as its $k$ images are, satisfies $\mathbb E[L^{c_{m+1}}_{\hat t_\ell}(x)]\le\alpha$, where $L^c_t(x)$ is the fraction of $c$'s pixels in $x$ scored below $t$.
\end{proposition}

This section proves \cref{prop:newimages}. For a label $c$ and an image $x$ annotated with it, let $L^c_t(x)\in[0,1]$ be the fraction of $c$'s pixels in $x$ whose score for $c$ is below $t$; it is non-decreasing in $t$ and, for a candidate-independent score, does not depend on the vocabulary. Label $c_i$ comes with images $X_i=(x_{i1},\dots,x_{ik})$, and $\bar\ell_i(t)=\frac1k\sum_j L^{c_i}_t(x_{ij})$. The units $U_i=(c_i,X_i)$, $i\le m+1$, are exchangeable, and their joint distribution is invariant to permuting the images within $U_{m+1}$ while leaving the other units fixed; the test image is $x=x_{m+1,1}$, and the other images of $U_{m+1}$ enter the argument only.

Let $t'$ be the largest grid threshold with $\frac1{m+1}\sum_{i\le m+1}\bar\ell_i(t)\le\alpha$; it exists because every loss is zero at the lowest grid point. Since $\bar\ell_{m+1}\le1$, every $t$ that is feasible for $\hat t_\ell$, \ie $(\sum_{i\le m}\bar\ell_i(t)+1)/(m+1)\le\alpha$, is feasible for $t'$, so $\hat t_\ell\le t'$ and $L^{c_{m+1}}_{\hat t_\ell}(x)\le L^{c_{m+1}}_{t'}(x)$. The threshold $t'$ is a symmetric function of the units and, within $U_{m+1}$, of the images. Exchangeability of the images gives $\mathbb E[L^{c_{m+1}}_{t'}(x)]=\mathbb E[\bar\ell_{m+1}(t')]$, and exchangeability of the units gives $\mathbb E[\bar\ell_{m+1}(t')]=\mathbb E[\frac1{m+1}\sum_i\bar\ell_i(t')]\le\alpha$. Hence $\mathbb E[L^{c_{m+1}}_{\hat t_\ell}(x)]\le\alpha$.

The number of images may differ between labels if it is part of the unit (\eg drawn independently for each label). Sharing images does not itself preclude exchangeability. However, a random split of benchmark images does not implement the per-label sampling design assumed here. \Cref{tab:labelimages} therefore repeats the label-level analysis of \cref{tab:labelheld} with the image-averaged loss, and reports it on held-out images, where it estimates the expected miss rate of an added label on a new image. The trials are those of \cref{tab:labelheld}; with the pooled loss the replay reproduces that table. On held-out images the added-label risk is $\LxImgMinHeld$--$\LxImgMaxHeld\%$ at $\alpha=10\%$ and $\LxImgBMinHeld$--$\LxImgBMaxHeld\%$ at $\alpha=5\%$, within $\LxImgMaxGap$ points of its value on the calibration images, for every model and benchmark. As for \cref{prop:labels}, the bound holds in expectation over the label draw, and $\LxImgMinAbove$--$\LxImgMaxAbove\%$ of single draws exceed $\alpha=10\%$. \LxImgPending
\begin{table*}[tp]
  \centering
  \caption{\textbf{Added labels on new images} (\%): anchored model, label-level threshold with the image-averaged label loss of \cref{prop:newimages}, random source vocabulary, \LxTrials\ trials. \emph{cal.}\ and \emph{held-out}: mean added-label risk on the calibration and held-out images; \emph{draws $>\alpha$}: percentage of draws whose held-out risk exceeds $\alpha$.}
  \label{tab:labelimages}
  \adjustbox{max width=\textwidth}{\footnotesize\setlength{\tabcolsep}{4pt}\begin{tabular}{@{}llrrrrrr@{}}
\toprule
& & \multicolumn{3}{c}{$\alpha=.10$} & \multicolumn{3}{c}{$\alpha=.05$} \\
\cmidrule(lr){3-5}\cmidrule(l){6-8}
Model & benchmark & cal. & held-out & draws $>\alpha$ & cal. & held-out & draws $>\alpha$ \\
\midrule
CAT-Seg ViT-B & ADE20K & 9.5 & 9.6 & 40 & 4.4 & 4.4 & 34 \\
CAT-Seg ViT-B & Pascal Context & 7.3 & 7.3 & 24 & 2.4 & 2.4 & 10 \\
CAT-Seg ViT-B & COCO-Stuff & 9.0 & 9.1 & 35 & 4.0 & 4.1 & 27 \\
CAT-Seg ViT-L & ADE20K & 9.4 & 9.5 & 39 & 4.5 & 4.5 & 35 \\
CAT-Seg ViT-L & Pascal Context & 7.5 & 7.4 & 23 & 2.5 & 2.5 & 9 \\
CAT-Seg ViT-L & COCO-Stuff & 8.9 & 8.9 & 34 & 3.9 & 4.0 & 27 \\
CAT-Seg + DeCLIP & ADE20K & 9.5 & 9.4 & 40 & 4.4 & 4.4 & 31 \\
CAT-Seg + DeCLIP & Pascal Context & 7.4 & 7.4 & 24 & 2.6 & 2.6 & 13 \\
CAT-Seg + DeCLIP & COCO-Stuff & 9.0 & 9.0 & 32 & 4.0 & 4.1 & 26 \\
SED & ADE20K & 9.5 & 9.5 & 40 & 4.5 & 4.5 & 37 \\
SED & Pascal Context & 7.5 & 7.5 & 25 & 2.6 & 2.6 & 12 \\
SED & COCO-Stuff & 9.0 & 9.0 & 35 & 4.0 & 4.0 & 27 \\
FC-CLIP & ADE20K & 9.5 & 9.5 & 41 & 4.5 & 4.5 & 35 \\
FC-CLIP & Pascal Context & 7.4 & 7.5 & 24 & 2.4 & 2.4 & 11 \\
FC-CLIP & COCO-Stuff & 9.0 & 9.0 & 31 & 4.0 & 3.9 & 25 \\
\bottomrule
\end{tabular}}}
\end{table*}

\section{Worst-case added-label losses}
\label{sec:impossibilityproof}
\begin{proposition}[No assumption-free guarantee]
\label{prop:impossible}
Fix a threshold rule that uses no annotations of the added label, with $\hat t$ on a grid of minimum $t_{\min}$. Uniformly over all added-label loss functions $\ell:[t_{\min},t_{\max}]\to[0,1]$ that are non-decreasing and satisfy $\ell(t_{\min})=0$, including loss functions chosen after seeing $\hat t$, the bound $\mathbb E[\ell(\hat t)]\le\alpha$ holds if and only if $\Pr(\hat t>t_{\min})\le\alpha$.
\end{proposition}
The admissible loss $\ell(t)=\ind[t>t_{\min}]$, fixed in advance, proves necessity. Every admissible loss, however it is chosen, is bounded by this indicator at every $t$, which proves sufficiency. An adversary who selects the added label after seeing the threshold is therefore no stronger than one who selects it in advance. This result applies to any threshold rule in the stated class, including native recalibration and anchored scoring. Randomly annotating added names (\cref{sec:weights}) supplies an exchangeable sample for the remaining names, but does not protect against adversarial selection from them. The result concerns worst-case loss functions; a particular model need not contain a label attaining that worst case.

\section{Experimental setup}
\label{sec:impl}
This section describes the parts of the setup of \cref{sec:results} that are specific to this paper: the COCO-Stuff benchmark and the fixed choices that make the experiments reproducible.

\subsection{COCO-Stuff grown with LVIS}
\label{sec:cocoall}
The third benchmark starts from the training vocabulary of CAT-Seg and SED. All choices below were fixed before any score was computed.
\begin{itemize}
\item \emph{Source vocabulary:} the $171$ COCO-Stuff classes \cite{caesar2018cocostuff}, named as in CAT-Seg's release.
\item \emph{Added labels:} the $1{,}112$ LVIS v1 categories \cite{gupta2019lvis} that are not COCO-Stuff concepts, for $1{,}283$ labels in total. An LVIS category counts as a COCO-Stuff concept when its synset is the one LVIS assigns to a COCO thing class, or when a synonym without a parenthetical qualifier, with underscores read as spaces, equals a COCO-Stuff name after its suffix \emph{-other}, \emph{-stuff} or \emph{-merged} is dropped.
\item \emph{Images:} the first $2{,}000$ COCO val2017 images that LVIS annotates. None of the five models was trained on them.
\item \emph{Added-label annotation:} a sampled pixel takes an added label when the smallest LVIS instance covering it belongs to an added category and its COCO-Stuff label is a stuff class or none. COCO thing masks are never overridden, since LVIS also annotates parts and worn items.
\end{itemize}

\subsection{Reproducibility}
\label{sec:repro}
All thresholds use a score grid of step $1/64$ on $[-30,30]$ with the all-label endpoint; the CRC rule selects the largest grid threshold whose corrected calibration loss $\big(\sum_i L + 1\big)/(n+1)$ is at most $\alpha$. Random splits use a fixed seed. The reserved images are listed in a manifest fixed before any experiment, and label names, nested addition orders and prompt templates are fixed in hashed manifests. Scores are stored in float16, clamped to $[-30,30]$. They are logits, except for FC-CLIP, whose stored score is the logarithm of its mask-weighted class score. Every number in the paper and this supplement is generated from the stored results by a script. Code, manifests and results are available at \url{https://github.com/munim110/open-vocab-crc}. The stored scores are too large to host; the code regenerates them from the datasets and the released checkpoints.

\section{Calibration reuse: complete results}
\label{sec:fullresults}
\begin{figure}[t]
  \centering
  \includegraphics[width=\linewidth]{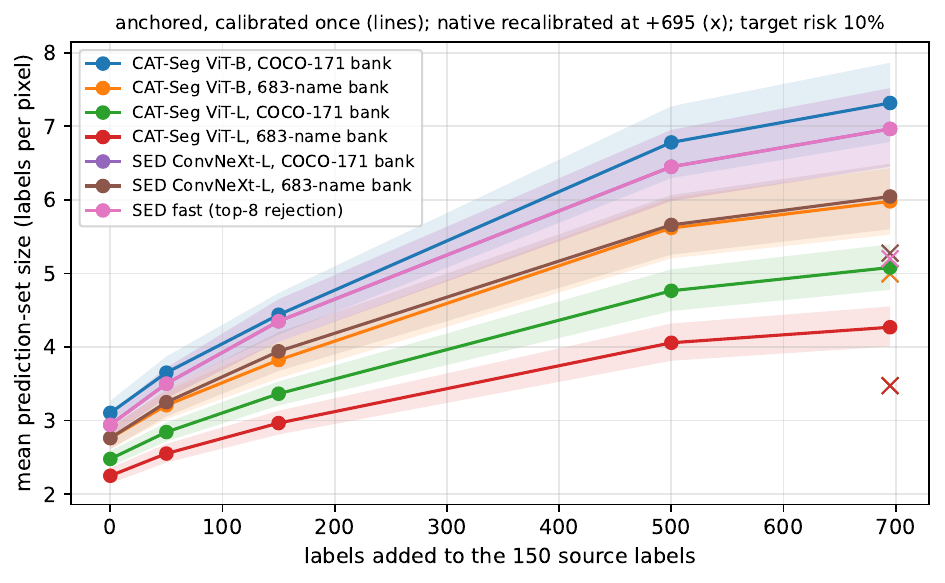}
  \caption{\textbf{Set size against the number of added labels} on ADE20K for every configuration.}
  \label{fig:curve}
\end{figure}

This section gives the complete results behind \cref{fig:anchored}. \Cref{tab:main} lists its values, and \cref{tab:mainb} repeats the comparison at $\alpha=5\%$. \Cref{tab:adeall,tab:pascalmain,tab:cocomain} give every main configuration on each benchmark, including the COCO-171 bank and SED's fast mode, with the fraction of splits on which each threshold exceeds $\alpha$. \Cref{tab:curve} and \cref{fig:curve} give the set size at each step of the nested ADE20K growth. The section ends with the distribution of the set size over pixels (\cref{tab:setsize}), native references (\cref{tab:nativerefs}) and a general multimodal model.

The conclusions of \cref{sec:main} hold at $\alpha=5\%$. The reused native threshold exceeds the target for CAT-Seg and FC-CLIP on every benchmark and stays below it for SED. The anchored threshold meets the target in every case, with an overhead of at most $\OverviewCostAggPriceMaxB$ labels per pixel for CAT-Seg and SED. The guarantee holds in expectation over the calibration draw: at $\alpha=10\%$ single splits exceed the target in $\OverviewAnchFracMin$--$\OverviewAnchFracMax\%$ of cases for the anchored threshold and in $\OverviewReplayFracMin$--$\OverviewReplayFracMax\%$ for native recalibration. With the smaller COCO-171 bank the anchored threshold remains valid, but at $\alpha=10\%$ the overhead roughly doubles for CAT-Seg and SED ($\VitLPrice$ instead of $\VitLBankPrice$ labels per pixel for ViT-L on ADE20K). Models whose scores are candidate-independent by construction are valid but far less efficient (\cref{tab:indep}).

\paragraph{Native references.}\label{sec:nativerefs}
\Cref{tab:nativerefs} adds two native references on ADE20K, from the stored scores and splits of \cref{tab:adeall}. Calibrated and deployed at the $150$ source names, the native model gives smaller sets than the anchored model ($\NrVitLNatSrcSize$ against $\NrVitLAnchSrcSize$ labels per pixel for CAT-Seg ViT-L at $\alpha=10\%$, and $\NrSedNatSrcSize$ against $\NrSedAnchSrcSize$ for SED), so part of the overhead of anchoring is present before the vocabulary grows. The reused native threshold can also be corrected by calibrating at the source vocabulary with a level $\alpha^\prime$ in place of $\alpha$. We take the largest $\alpha^\prime$, on a grid of step $0.25\%$, whose reused threshold has mean risk at most $\alpha$ at $845$ names on the evaluation images. The corrected sets are as small as those of native recalibration ($\NrVitLTightSize$ against $\NrVitLRecalSize$ for ViT-L). Selecting $\alpha^\prime$ requires labelled evaluation images at the grown vocabulary, and the direction of correction depends on the model: $\alpha^\prime=\NrVitLAlphaPrime\%$ for ViT-L, whose reused threshold exceeds the target, and $\NrSedAlphaPrime\%$ for SED, whose reused threshold is over-conservative. This is therefore an oracle comparison: a level chosen without those labels has no corresponding guarantee. \paragraph{Agreement with native predictions.} The native top-1 label over the $845$ names lies outside the anchored set at $\NrVitLOutside\%$ of the source-labelled evaluation pixels for ViT-L and $\NrSedOutside\%$ for SED at $\alpha=10\%$, and at $\NrVitLBOutside\%$ and $\NrSedBOutside\%$ at $\alpha=5\%$.
\paragraph{A general multimodal model.}\label{sec:mllm} A general multimodal model can name image regions from a list without a segmentation head. We asked Claude Opus 5.5 (\texttt{claude-opus-5-5}, Anthropic API, November 2026) about \ClaudeImages\ ADE20K images drawn at random. Each image carries $15$ numbered markers, placed on stored sample pixels with a source label (Set-of-Mark prompting \cite{yang2023som}). For each marker the model returns up to $10$ names from the given list, each with a probability, and these probabilities are its scores; unlisted names score $0$, since the API gives no token probabilities. Each image is asked twice, with the $150$ source names and with all $845$ names in the same relative order. The top name is correct at $\ClaudeTopOneSrc\%$ of the markers with $150$ names and at $\ClaudeTopOneFull\%$ with $845$. The scores are candidate-dependent: adding the names changes the probability of the true label at $\ClaudeProbChanged\%$ of the markers and the top name at $\ClaudeTopChanged\%$. The true label is missing from the answer at $\ClaudeUnlistedSrc\%$ of the markers with $150$ names and at $\ClaudeUnlistedFull\%$ with $845$. Both exceed $\alpha=10\%$, so CRC, reused or recalibrated, falls to the all-label endpoint and includes all $\ClaudeCrcSize$ labels. On the same markers and \ClaudeSplits\ random halves of the images, CAT-Seg ViT-L with the $683$-name bank has $\ClaudeCatReusedRisk\%$ risk with the reused native threshold; the anchored threshold has $\ClaudeCatAnchRisk\%$ with $\ClaudeCatAnchSize$ labels per marker, against $\ClaudeCatRecalSize$ for native recalibration. The queries cost about \$\ClaudeCostImage\ per image for both vocabularies. Under the API terms, inputs and outputs are kept for up to $30$ days and are not used for training; the ADE20K terms permit non-commercial research use. The prompts and every response are released with the code (\cref{sec:repro}).
\begin{table*}[tp]
  \centering
  \caption{\textbf{Native references} (ADE20K $150\!\to\!845$, \NumSplits\ splits). Size at $150$: labels per pixel when calibrated and deployed at the source names. Reused: risk (\%) at $845$ names of the native threshold calibrated at $150$. Reused at $\alpha^\prime$: the native model calibrated at $150$ names with the level $\alpha^\prime$ (\%) that meets $\alpha$ at $845$ names on the evaluation images (an oracle), with its risk (\%) and set size at $845$ names. Recal., anchored: set size at $845$ names, as in \cref{tab:adeall}. Top-1 outside: percentage of source-labelled pixels whose native top-1 label at $845$ names is not in the anchored set.}
  \label{tab:nativerefs}
  \adjustbox{max width=\textwidth}{\footnotesize\setlength{\tabcolsep}{4pt}\setlength{\tabcolsep}{3pt}\begin{tabular}{@{}llrrrrrrrrr@{}}
\toprule
& & \multicolumn{2}{c}{size at $150$} & reused & \multicolumn{3}{c}{reused at $\alpha^\prime$} & recal. & anchored & top-1 \\ \cmidrule(lr){3-4}\cmidrule(lr){6-8}
Model & $\alpha$ & native & anch. & risk & $\alpha^\prime$ & risk & size & size & size & outside \\
\midrule
CAT-Seg ViT-L & $0.10$ & 1.91 & 2.25 & 15.2 & 6.00 & 9.9 & 3.48 & 3.47 & 4.27 & 0.34 \\
 & $0.05$ & 3.28 & 3.66 & 8.4 & 2.50 & 4.8 & 7.95 & 7.81 & 8.63 & 0.01 \\
SED & $0.10$ & 2.22 & 2.76 & 7.1 & 15.25 & 10.0 & 5.20 & 5.27 & 6.04 & 0.20 \\
 & $0.05$ & 4.70 & 5.30 & 3.8 & 6.50 & 4.8 & 16.41 & 16.14 & 15.58 & 0.00 \\
\bottomrule
\end{tabular}}}
\end{table*}

\begin{table}[tp]
  \centering
  \caption{\textbf{Scores that are candidate-independent by construction} (ADE20K $150\!\to\!845$, $\alpha=10\%$, threshold calibrated at $150$ names): risk at $845$ names (\%) and set size at $150$ and $845$ names. SCLIP: CLIP ViT-B patch--text similarities with the correlative self-attention and sliding-window inference of SCLIP \cite{wang2024sclip}. The CLIPSeg row uses \ClipsegImages\ development images and \ClipsegSplits\ random splits into halves; the other rows use the \NumSplits\ splits of \cref{tab:adeall}.}
  \label{tab:indep}
  \footnotesize\begin{tabular}{@{}lrr@{}}
    \toprule
    Model & risk & set size \\
    \midrule
    SCLIP (CLIP ViT-B) & $\VitBSclipRisk$ & $\VitBSclipSrc\to\VitBSclipFull$ \\
    CLIPSeg \cite{luddecke2022clipseg} & $\ClipsegRisk$ & $\ClipsegSrc\to\ClipsegFull$ \\
    anchored CAT-Seg ViT-B & $\VitBBankAnchRiskOne$ & $\VitBBankAnchSrc\to\VitBBankAnchFull$ \\
    \bottomrule
  \end{tabular}
\end{table}
\begin{table*}[tp]
  \centering
  \caption{\textbf{Risk and set size after vocabulary growth} (values of \cref{fig:anchored}; $\alpha=10\%$, mean over \NumSplits\ splits). Risk: native threshold reused from the source vocabulary, anchored calibrated there (ours). Set size (labels per pixel at the grown vocabulary): native recalibrated there, anchored.}
  \label{tab:main}
  \adjustbox{max width=\textwidth}{\footnotesize\setlength{\tabcolsep}{4pt}\setlength{\tabcolsep}{4pt}\begin{tabular}{@{}lrrrrrrrrrrrr@{}}
\toprule
& \multicolumn{4}{c}{ADE20K $150\!\to\!845$} & \multicolumn{4}{c}{Pascal Context $59\!\to\!458$} & \multicolumn{4}{c}{COCO-Stuff $171\!\to\!1{,}283$} \\
\cmidrule(lr){2-5}\cmidrule(lr){6-9}\cmidrule(l){10-13}
& \multicolumn{2}{c}{risk (\%)} & \multicolumn{2}{c}{set size} & \multicolumn{2}{c}{risk (\%)} & \multicolumn{2}{c}{set size} & \multicolumn{2}{c}{risk (\%)} & \multicolumn{2}{c}{set size} \\
\cmidrule(lr){2-3}\cmidrule(lr){4-5}\cmidrule(lr){6-7}\cmidrule(lr){8-9}\cmidrule(lr){10-11}\cmidrule(l){12-13}
Model & reused & ours & recal. & ours & reused & ours & recal. & ours & reused & ours & recal. & ours \\
\midrule
CAT-Seg ViT-B & 15.4 & 9.9 & 4.99 & 5.98 & 15.2 & 9.8 & 2.52 & 2.82 & 12.9 & 9.9 & 3.11 & 3.84 \\
CAT-Seg ViT-L & 15.2 & 9.9 & 3.47 & 4.27 & 15.0 & 9.8 & 2.07 & 2.34 & 12.8 & 9.8 & 2.66 & 3.22 \\
CAT-Seg + DeCLIP & 14.7 & 9.9 & 3.39 & 4.38 & 14.7 & 9.8 & 2.09 & 2.42 & 12.9 & 9.9 & 2.64 & 3.19 \\
SED & 7.1 & 9.9 & 5.28 & 6.05 & 4.6 & 9.8 & 2.68 & 2.80 & 3.6 & 9.9 & 3.74 & 3.55 \\
FC-CLIP & 13.8 & 9.9 & 8.18 & 11.18 & 16.2 & 9.9 & 5.30 & 5.80 & 17.1 & 9.9 & 26.50 & 26.82 \\
\bottomrule
\end{tabular}}}
\end{table*}
\begin{table*}[tp]
  \centering
  \caption{\textbf{Risk and set size after vocabulary growth at $\alpha=5\%$} (columns as in \cref{tab:main}).}
  \label{tab:mainb}
  \adjustbox{max width=\textwidth}{\footnotesize\setlength{\tabcolsep}{4pt}\setlength{\tabcolsep}{4pt}\begin{tabular}{@{}lrrrrrrrrrrrr@{}}
\toprule
& \multicolumn{4}{c}{ADE20K $150\!\to\!845$} & \multicolumn{4}{c}{Pascal Context $59\!\to\!458$} & \multicolumn{4}{c}{COCO-Stuff $171\!\to\!1{,}283$} \\
\cmidrule(lr){2-5}\cmidrule(lr){6-9}\cmidrule(l){10-13}
& \multicolumn{2}{c}{risk (\%)} & \multicolumn{2}{c}{set size} & \multicolumn{2}{c}{risk (\%)} & \multicolumn{2}{c}{set size} & \multicolumn{2}{c}{risk (\%)} & \multicolumn{2}{c}{set size} \\
\cmidrule(lr){2-3}\cmidrule(lr){4-5}\cmidrule(lr){6-7}\cmidrule(lr){8-9}\cmidrule(lr){10-11}\cmidrule(l){12-13}
Model & reused & ours & recal. & ours & reused & ours & recal. & ours & reused & ours & recal. & ours \\
\midrule
CAT-Seg ViT-B & 8.1 & 4.9 & 13.29 & 14.25 & 8.4 & 4.9 & 4.66 & 4.95 & 6.9 & 4.9 & 5.96 & 7.12 \\
CAT-Seg ViT-L & 8.4 & 4.9 & 7.81 & 8.63 & 8.5 & 4.9 & 3.38 & 3.66 & 6.9 & 4.9 & 4.89 & 5.75 \\
CAT-Seg + DeCLIP & 7.9 & 4.9 & 7.64 & 8.74 & 8.4 & 4.9 & 3.58 & 3.85 & 6.8 & 4.9 & 4.84 & 5.67 \\
SED & 3.8 & 4.9 & 16.19 & 15.62 & 2.1 & 4.9 & 4.63 & 4.77 & 1.6 & 4.9 & 6.47 & 6.17 \\
FC-CLIP & 6.8 & 4.9 & 25.65 & 31.19 & 8.6 & 4.9 & 12.77 & 13.99 & 9.8 & 4.9 & 82.70 & 77.21 \\
\bottomrule
\end{tabular}}}
\end{table*}
\begin{table*}[tp]
  \centering
  \caption{\textbf{ADE20K $150\!\to\!845$, every main configuration} ($\alpha=10\%$, \NumSplits\ splits). Risk in \%; set size in labels per pixel, anchored at the source $\to$ grown vocabulary; overhead = anchored minus recalibrated native size, computed before rounding. ``Splits $>\alpha$'': fraction of splits on which the risk of that arm exceeds the target. In the SED fast-mode row only the native arms use fast mode; the anchored arm is SED in its default mode with the COCO-171 bank, identical to the row above.}
  \label{tab:adeall}
  \adjustbox{max width=\textwidth}{\footnotesize\setlength{\tabcolsep}{4pt}\begin{tabular}{@{}llrrrrrrrrr@{}}
\toprule
& & \multicolumn{2}{c}{native, reused} & \multicolumn{3}{c}{native, recalibrated} & \multicolumn{4}{c}{anchored (ours)} \\
\cmidrule(lr){3-4}\cmidrule(lr){5-7}\cmidrule(lr){8-11}
Model & bank & risk & splits $>\alpha$ & risk & splits $>\alpha$ & size & risk & splits $>\alpha$ & size & overhead \\
\midrule
CAT-Seg ViT-B & COCO-171 bank & 15.4 & 100\% & 9.9 & 41\% & 4.99 & 9.9 & 39\% & 3.10$\to$7.32 & +2.32 \\
CAT-Seg ViT-B & 683-name bank & 15.4 & 100\% & 9.9 & 41\% & 4.99 & 9.9 & 41\% & 2.76$\to$5.98 & +0.98 \\
CAT-Seg ViT-L & COCO-171 bank & 15.2 & 100\% & 9.9 & 43\% & 3.47 & 9.9 & 40\% & 2.48$\to$5.08 & +1.61 \\
CAT-Seg ViT-L & 683-name bank & 15.2 & 100\% & 9.9 & 43\% & 3.47 & 9.9 & 42\% & 2.25$\to$4.27 & +0.79 \\
CAT-Seg + DeCLIP EVA-L & 683-name bank & 14.7 & 100\% & 9.9 & 41\% & 3.39 & 9.9 & 42\% & 2.18$\to$4.38 & +0.99 \\
SED ConvNeXt-L & COCO-171 bank & 7.1 & 0\% & 9.9 & 43\% & 5.28 & 9.9 & 42\% & 2.94$\to$6.96 & +1.69 \\
SED ConvNeXt-L & fast mode (top-8 rejection) & 12.8 & 100\% & 9.9 & 41\% & 5.20 & 9.9 & 42\% & 2.94$\to$6.96 & +1.76 \\
SED ConvNeXt-L & 683-name bank & 7.1 & 0\% & 9.9 & 43\% & 5.28 & 9.9 & 43\% & 2.76$\to$6.05 & +0.77 \\
FC-CLIP ConvNeXt-L & 133-name bank & 13.8 & 100\% & 9.9 & 41\% & 8.18 & 9.9 & 41\% & 4.11$\to$13.56 & +5.37 \\
FC-CLIP ConvNeXt-L & 701-name bank & 13.8 & 100\% & 9.9 & 41\% & 8.18 & 9.9 & 38\% & 3.80$\to$11.18 & +2.99 \\
\bottomrule
\end{tabular}}}
\end{table*}
\begin{table*}[tp]
  \centering
  \caption{\textbf{Pascal Context $59\!\to\!458$, every configuration} (columns as in \cref{tab:adeall}).}
  \label{tab:pascalmain}
  \adjustbox{max width=\textwidth}{\footnotesize\setlength{\tabcolsep}{4pt}\begin{tabular}{@{}llrrrrrrrrr@{}}
\toprule
& & \multicolumn{2}{c}{native, reused} & \multicolumn{3}{c}{native, recalibrated} & \multicolumn{4}{c}{anchored (ours)} \\
\cmidrule(lr){3-4}\cmidrule(lr){5-7}\cmidrule(lr){8-11}
Model & bank & risk & splits $>\alpha$ & risk & splits $>\alpha$ & size & risk & splits $>\alpha$ & size & overhead \\
\midrule
CAT-Seg ViT-B & COCO-171 bank, Pascal 59$\to$458 & 15.2 & 100\% & 9.9 & 46\% & 2.52 & 9.8 & 42\% & 1.78$\to$3.35 & +0.83 \\
CAT-Seg ViT-B & 683-name bank, Pascal 59$\to$458 & 15.2 & 100\% & 9.9 & 46\% & 2.52 & 9.8 & 42\% & 1.67$\to$2.82 & +0.30 \\
CAT-Seg ViT-L & 683-name bank, Pascal 59$\to$458 & 15.0 & 100\% & 9.9 & 43\% & 2.07 & 9.8 & 40\% & 1.51$\to$2.34 & +0.27 \\
CAT-Seg + DeCLIP EVA-L & 683-name bank, Pascal 59$\to$458 & 14.7 & 100\% & 9.8 & 39\% & 2.09 & 9.8 & 40\% & 1.51$\to$2.42 & +0.33 \\
SED ConvNeXt-L & 683-name bank, Pascal 59$\to$458 & 4.6 & 0\% & 9.8 & 43\% & 2.68 & 9.8 & 41\% & 1.67$\to$2.80 & +0.13 \\
FC-CLIP ConvNeXt-L & 701-name bank, Pascal 59$\to$458 & 16.2 & 100\% & 9.9 & 40\% & 5.30 & 9.9 & 41\% & 2.01$\to$5.80 & +0.50 \\
\bottomrule
\end{tabular}}}
\end{table*}
\begin{table*}[tp]
  \centering
  \caption{\textbf{COCO-Stuff $171\!\to\!1{,}283$, every configuration} (columns as in \cref{tab:adeall}).}
  \label{tab:cocomain}
  \adjustbox{max width=\textwidth}{\footnotesize\setlength{\tabcolsep}{4pt}\begin{tabular}{@{}llrrrrrrrrr@{}}
\toprule
& & \multicolumn{2}{c}{native, reused} & \multicolumn{3}{c}{native, recalibrated} & \multicolumn{4}{c}{anchored (ours)} \\
\cmidrule(lr){3-4}\cmidrule(lr){5-7}\cmidrule(lr){8-11}
Model & bank & risk & splits $>\alpha$ & risk & splits $>\alpha$ & size & risk & splits $>\alpha$ & size & overhead \\
\midrule
CAT-Seg ViT-B & 683-name bank, COCO 171$\to$1,283 & 12.9 & 100\% & 9.9 & 38\% & 3.11 & 9.9 & 41\% & 2.33$\to$3.84 & +0.73 \\
CAT-Seg ViT-L & 683-name bank, COCO 171$\to$1,283 & 12.8 & 100\% & 9.8 & 40\% & 2.66 & 9.8 & 40\% & 2.06$\to$3.22 & +0.56 \\
CAT-Seg + DeCLIP EVA-L & 683-name bank, COCO 171$\to$1,283 & 12.9 & 100\% & 9.9 & 41\% & 2.64 & 9.9 & 42\% & 1.94$\to$3.19 & +0.55 \\
SED ConvNeXt-L & 683-name bank, COCO 171$\to$1,283 & 3.6 & 0\% & 9.9 & 43\% & 3.74 & 9.9 & 43\% & 2.19$\to$3.55 & -0.20 \\
FC-CLIP ConvNeXt-L & 701-name bank, COCO 171$\to$1,283 & 17.1 & 100\% & 9.9 & 44\% & 26.50 & 9.9 & 43\% & 5.86$\to$26.82 & +0.32 \\
\bottomrule
\end{tabular}}}
\end{table*}
\begin{table}[tp]
  \centering
  \caption{\textbf{Anchored set size at each nested addition} on ADE20K (labels per pixel, $\alpha=10\%$).}
  \label{tab:curve}
  \adjustbox{max width=\linewidth}{\footnotesize\setlength{\tabcolsep}{4pt}\begin{tabular}{@{}lrrrrr@{}}
\toprule
Configuration & +0 & +50 & +150 & +500 & +695 \\
\midrule
CAT-Seg ViT-B, COCO-171 bank & 3.10 & 3.65 & 4.44 & 6.78 & 7.32 \\
CAT-Seg ViT-B, 683-name bank & 2.76 & 3.21 & 3.82 & 5.62 & 5.98 \\
CAT-Seg ViT-L, COCO-171 bank & 2.48 & 2.84 & 3.36 & 4.76 & 5.08 \\
CAT-Seg ViT-L, 683-name bank & 2.25 & 2.55 & 2.97 & 4.06 & 4.27 \\
CAT-Seg + DeCLIP EVA-L, 683-name bank & 2.18 & 2.53 & 2.94 & 4.19 & 4.38 \\
SED ConvNeXt-L, COCO-171 bank & 2.94 & 3.50 & 4.35 & 6.45 & 6.96 \\
SED ConvNeXt-L, 683-name bank & 2.76 & 3.25 & 3.94 & 5.66 & 6.05 \\
FC-CLIP ConvNeXt-L, 133-name bank & 4.11 & 5.36 & 7.13 & 11.96 & 13.56 \\
FC-CLIP ConvNeXt-L, 701-name bank & 3.80 & 4.84 & 6.21 & 10.01 & 11.18 \\
\bottomrule
\end{tabular}}}
\end{table}

\begin{figure}[tp]
  \centering
  \includegraphics[width=\linewidth]{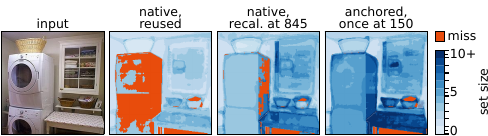}
  \caption{\textbf{Prediction sets at $845$ names} (CAT-Seg ViT-L, $\alpha=10\%$, median split, selected ADE20K image). Blue: labels per pixel. Orange: missed source-labelled pixels. From left to right, the maps miss $\QualReusedMiss$, $\QualRecalMiss$ and $\QualAnchMiss\%$ of them in this image and $\QualPoolReusedMiss$, $\QualPoolRecalMiss$ and $\QualPoolAnchMiss\%$ over the \QualPoolImages\ test images of this split with stored full-image outputs. Random images are in \cref{fig:qualpanel}.}
  \label{fig:qual}
\end{figure}

\begin{figure*}[tp]
  \centering
  \includegraphics[width=.9\textwidth]{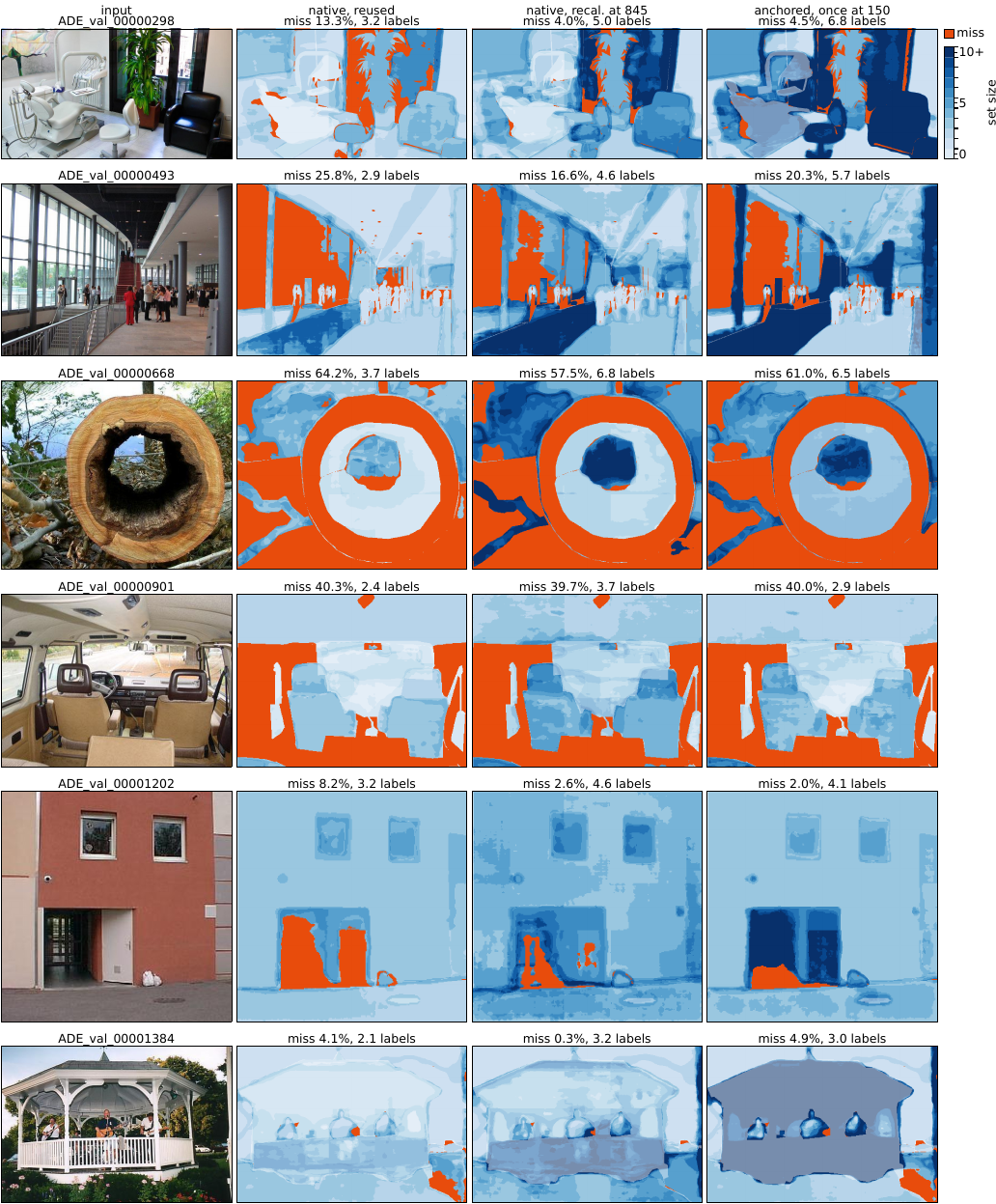}
  \caption{\textbf{Prediction sets at $845$ names on random images} (CAT-Seg ViT-L, $\alpha=10\%$; the thresholds of the median split as in \cref{fig:qual}; \QualPanelImages\ images drawn at random with a fixed seed from the same pool of \QualPanelPool\ test images). Blue: labels per pixel; orange: missed source-labelled pixels; faded: unlabelled pixels. Above each panel: missed source-labelled pixels and mean labels per source-labelled pixel on that image.}
  \label{fig:qualpanel}
\end{figure*}

\paragraph{Distribution of the set size.}\label{sec:setsize}
The set sizes in \cref{tab:main} are means over source-labelled pixels. \Cref{tab:setsize} gives the distribution behind them on the same splits and thresholds for \SsdConfigs\ of the configurations of \cref{tab:main}; \cref{tab:setsizeb} repeats it at $\alpha=5\%$. Compared with the recalibrated native sets, the anchored sets have the same or a smaller median in every configuration. They also hold a single label at more pixels in every configuration but one (on ADE20K, $\SsdVitLBankAnchOne\%$ of pixels against $\SsdVitLBankRecalOne\%$ for CAT-Seg ViT-L and $\SsdSedBankAnchOne\%$ against $\SsdSedBankRecalOne\%$ for SED); the exception is CAT-Seg ViT-L on Pascal Context ($\SsdPascalVitLBankAnchOne\%$ against $\SsdPascalVitLBankRecalOne\%$). Their larger mean comes from a tail of pixels with more than ten labels: $\SsdVitLBankAnchGtTen\%$ against $\SsdVitLBankRecalGtTen\%$ for CAT-Seg ViT-L and $\SsdSedBankAnchGtTen\%$ against $\SsdSedBankRecalGtTen\%$ for SED on ADE20K. The reused native SED threshold, which is over-conservative after growth, places $\SsdSedBankReusedGtTen\%$ of the ADE20K pixels above ten labels. Empty sets occur at under $1\%$ of pixels under every threshold. \Cref{fig:qual} adds the anchored threshold to the image of \cref{fig:problem}. \Cref{fig:qualpanel} shows the sets on \QualPanelImages\ images drawn at random, with the same thresholds. Over these images, the reused native threshold misses $\QualPanelReusedMiss\%$ of the source-labelled pixels, recalibration $\QualPanelRecalMiss\%$ and the anchored threshold $\QualPanelAnchMiss\%$, at $\QualPanelReusedSize$, $\QualPanelRecalSize$ and $\QualPanelAnchSize$ labels per pixel. Two of the six images are hard for every threshold; the guarantee is an average over images ($\QualPoolReusedMiss$, $\QualPoolRecalMiss$ and $\QualPoolAnchMiss\%$ over the \QualPoolImages\ images of the pool).
\begin{table*}[tp]
  \centering
  \caption{\textbf{Distribution of the per-pixel set size} ($\alpha=10\%$, \NumSplits\ splits as in \cref{tab:main}): percentage of source-labelled test pixels whose prediction set is empty, has one label, at most three, at most ten or more than ten; median of the split-averaged distribution; mean as in \cref{tab:main}. \emph{native at source}: calibrated and deployed at the source names.}
  \label{tab:setsize}
  \adjustbox{max width=\textwidth}{\footnotesize\setlength{\tabcolsep}{4pt}\begin{tabular}{@{}llrrrrrrrr@{}}
\toprule
& & & \multicolumn{5}{c}{pixels with $|C|$ (\%)} & \multicolumn{2}{c}{labels per pixel} \\
\cmidrule(lr){4-8}\cmidrule(l){9-10}
Configuration & threshold & names & $=0$ & $=1$ & $\le 3$ & $\le 10$ & $>10$ & median & mean \\
\midrule
ADE20K, CAT-Seg ViT-L & native at source & 150 & 0.2 & 46.6 & 90.5 & 100.0 & 0.0 & 2 & 1.91 \\
 & native, reused & 845 & 0.3 & 38.0 & 85.5 & 100.0 & 0.0 & 2 & 2.16 \\
 & native, recalibrated & 845 & 0.0 & 22.1 & 58.0 & 99.4 & 0.6 & 3 & 3.47 \\
 & anchored & 845 & 0.3 & 23.5 & 51.8 & 94.5 & 5.5 & 3 & 4.27 \\
\addlinespace
ADE20K, SED & native at source & 150 & 0.1 & 38.9 & 83.3 & 100.0 & 0.0 & 2 & 2.22 \\
 & native, reused & 845 & 0.0 & 5.9 & 15.2 & 67.0 & 33.0 & 8 & 8.80 \\
 & native, recalibrated & 845 & 0.0 & 10.5 & 29.8 & 95.2 & 4.8 & 5 & 5.28 \\
 & anchored & 845 & 0.1 & 19.4 & 39.7 & 82.9 & 17.1 & 5 & 6.05 \\
\addlinespace
Pascal Context, CAT-Seg ViT-L & native at source & 59 & 0.2 & 70.7 & 99.2 & 100.0 & 0.0 & 1 & 1.37 \\
 & native, reused & 458 & 0.5 & 63.6 & 97.8 & 100.0 & 0.0 & 1 & 1.49 \\
 & native, recalibrated & 458 & 0.0 & 47.4 & 85.1 & 100.0 & 0.0 & 2 & 2.07 \\
 & anchored & 458 & 0.4 & 44.6 & 79.3 & 99.8 & 0.2 & 2 & 2.34 \\
\addlinespace
Pascal Context, SED & native at source & 59 & 0.1 & 62.7 & 97.2 & 100.0 & 0.0 & 1 & 1.53 \\
 & native, reused & 458 & 0.0 & 7.5 & 36.0 & 95.8 & 4.2 & 4 & 4.80 \\
 & native, recalibrated & 458 & 0.2 & 17.6 & 77.2 & 100.0 & 0.0 & 3 & 2.68 \\
 & anchored & 458 & 0.8 & 27.0 & 72.2 & 99.6 & 0.4 & 2 & 2.80 \\
\addlinespace
COCO-Stuff, SED & native at source & 171 & 0.1 & 41.2 & 89.6 & 100.0 & 0.0 & 2 & 2.01 \\
 & native, reused & 1,283 & 0.0 & 1.1 & 14.8 & 72.7 & 27.3 & 8 & 8.21 \\
 & native, recalibrated & 1,283 & 0.3 & 7.2 & 49.4 & 100.0 & 0.0 & 4 & 3.74 \\
 & anchored & 1,283 & 0.2 & 17.9 & 61.5 & 98.2 & 1.8 & 3 & 3.55 \\
\bottomrule
\end{tabular}}}
\end{table*}
\begin{table*}[tp]
  \centering
  \caption{\textbf{Distribution of the per-pixel set size at $\alpha=5\%$} (columns as in \cref{tab:setsize}).}
  \label{tab:setsizeb}
  \adjustbox{max width=\textwidth}{\footnotesize\setlength{\tabcolsep}{4pt}\begin{tabular}{@{}llrrrrrrrr@{}}
\toprule
& & & \multicolumn{5}{c}{pixels with $|C|$ (\%)} & \multicolumn{2}{c}{labels per pixel} \\
\cmidrule(lr){4-8}\cmidrule(l){9-10}
Configuration & threshold & names & $=0$ & $=1$ & $\le 3$ & $\le 10$ & $>10$ & median & mean \\
\midrule
ADE20K, CAT-Seg ViT-L & native at source & 150 & 0.0 & 23.9 & 62.4 & 99.4 & 0.6 & 3 & 3.28 \\
 & native, reused & 845 & 0.0 & 17.0 & 48.4 & 97.0 & 3.0 & 4 & 4.21 \\
 & native, recalibrated & 845 & 0.0 & 6.5 & 23.5 & 74.2 & 25.8 & 7 & 7.81 \\
 & anchored & 845 & 0.0 & 11.0 & 27.9 & 69.1 & 30.9 & 7 & 8.63 \\
\addlinespace
ADE20K, SED & native at source & 150 & 0.0 & 14.8 & 43.6 & 93.9 & 6.1 & 4 & 4.71 \\
 & native, reused & 845 & 0.0 & 0.2 & 1.9 & 15.7 & 84.3 & 22 & 24.84 \\
 & native, recalibrated & 845 & 0.0 & 1.9 & 5.9 & 32.0 & 68.0 & 15 & 16.19 \\
 & anchored & 845 & 0.0 & 7.2 & 16.5 & 44.4 & 55.6 & 12 & 15.62 \\
\addlinespace
Pascal Context, CAT-Seg ViT-L & native at source & 59 & 0.0 & 52.6 & 89.7 & 100.0 & 0.0 & 1 & 1.87 \\
 & native, reused & 458 & 0.0 & 43.3 & 79.8 & 100.0 & 0.0 & 2 & 2.30 \\
 & native, recalibrated & 458 & 0.0 & 31.1 & 62.4 & 98.1 & 1.9 & 3 & 3.38 \\
 & anchored & 458 & 0.1 & 30.2 & 61.1 & 96.0 & 4.0 & 3 & 3.66 \\
\addlinespace
Pascal Context, SED & native at source & 59 & 0.0 & 42.4 & 82.9 & 100.0 & 0.0 & 2 & 2.20 \\
 & native, reused & 458 & 0.0 & 3.0 & 16.3 & 72.4 & 27.6 & 7 & 8.45 \\
 & native, recalibrated & 458 & 0.0 & 7.9 & 37.9 & 96.7 & 3.3 & 4 & 4.63 \\
 & anchored & 458 & 0.3 & 12.2 & 44.1 & 93.3 & 6.7 & 4 & 4.77 \\
\addlinespace
COCO-Stuff, SED & native at source & 171 & 0.0 & 25.0 & 64.7 & 99.3 & 0.7 & 3 & 3.21 \\
 & native, reused & 1,283 & 0.0 & 0.2 & 5.4 & 38.3 & 61.7 & 13 & 14.67 \\
 & native, recalibrated & 1,283 & 0.0 & 2.0 & 20.9 & 88.1 & 11.9 & 6 & 6.47 \\
 & anchored & 1,283 & 0.0 & 7.1 & 34.3 & 84.8 & 15.2 & 5 & 6.17 \\
\bottomrule
\end{tabular}}}
\end{table*}

\paragraph{Under the label-level threshold.}\label{sec:setsizelabel}
\Cref{tab:setsizelabel} gives the same distribution for the thresholds of \cref{tab:labelexch,tab:labelmodels}, on their trials and, as there, over the labelled held-out pixels, source or added. Here the sets are larger at most pixels, not only in a tail. For CAT-Seg ViT-L on ADE20K at $\alpha=10\%$ the median pixel has $\SsdLVitLBankLabMedian$ labels and $\SsdLVitLBankLabGtTen\%$ of pixels have more than ten, against $\SsdLVitLBankImgMedian$ and $\SsdLVitLBankImgGtTen\%$ under the image-level threshold, which gives no guarantee for the added labels. On Pascal Context, where fewer names are added, the median is $\SsdLPascalVitLBankLabMedian$ against $\SsdLPascalVitLBankImgMedian$. Each label counts equally in the label-level loss, so rare added labels with low scores at their pixels pull the threshold down, and the lower threshold applies at every pixel. In each of the \SsdLConfigs\ configurations of \cref{tab:setsizelabel} the anchored sets are smaller than those of native recalibration with the label-level threshold (ADE20K, CAT-Seg ViT-L: median $\SsdLVitLBankNatLabMedian$, $\SsdLVitLBankNatLabGtTen\%$ above ten). The one exception in \cref{tab:labelmodels}, FC-CLIP on COCO-Stuff, is not among these configurations. With the benchmark's own source classes the sets are smaller (median $\SsdLSplitVitLBankLabMedian$, $\SsdLSplitVitLBankLabGtTen\%$ above ten on ADE20K) because the rare labels are not among the calibrated ones, which is also why the guarantee fails there (\cref{tab:labelexch}).
\begin{table*}[tp]
  \centering
  \caption{\textbf{Distribution of the per-pixel set size under the label-level threshold} (random source vocabulary, the trials of \cref{tab:labelexch,tab:labelmodels}; for CAT-Seg ViT-L on ADE20K one of \LxAdeSets\ draws, as in \cref{tab:labelexch}): percentage of labelled held-out pixels whose prediction set has one label, at most three, at most ten or more than ten; median of the trial-averaged distribution; mean as in those tables. \emph{anchored, image-level}: the threshold of \cref{tab:main}, which gives no guarantee for the added labels.}
  \label{tab:setsizelabel}
  \adjustbox{max width=\textwidth}{\footnotesize\setlength{\tabcolsep}{4pt}\begin{tabular}{@{}llrrrrrr@{}}
\toprule
& & \multicolumn{4}{c}{pixels with $|C|$ (\%)} & \multicolumn{2}{c}{labels per pixel} \\
\cmidrule(lr){3-6}\cmidrule(l){7-8}
Configuration & threshold & $=1$ & $\le 3$ & $\le 10$ & $>10$ & median & mean \\
\midrule
\multicolumn{8}{@{}l}{\emph{$\alpha=10\%$}} \\
ADE20K, CAT-Seg ViT-L & anchored, image-level & 15.3 & 35.8 & 79.7 & 20.3 & 5 & 6.79 \\
 & anchored, label-level & 1.1 & 4.8 & 24.7 & 75.3 & 20 & 28.85 \\
 & native recal., label-level & 0.1 & 0.8 & 12.6 & 87.4 & 27 & 35.80 \\
\addlinespace
ADE20K, SED & anchored, image-level & 10.8 & 22.9 & 55.4 & 44.6 & 9 & 16.79 \\
 & anchored, label-level & 0.6 & 2.9 & 12.9 & 87.1 & 34 & 45.10 \\
\bottomrule
\end{tabular}}}
\end{table*}

\subsection{Per-image vocabularies}
\label{sec:adaptivesed}
\Cref{tab:adaptive} gives the per-image results of \cref{sec:adaptive} for CAT-Seg ViT-L. The worst case in these tables is over the five fixed rules, not over all vocabularies, and which rules a native threshold fails depends on the architecture (\cref{tab:adaptivesed,tab:adaptivefcclip}). These rules vary both the number of added names and their relation to the image. Anchored source-label scores remain identical across rules, while native scores can change. An external tagger such as RAM \cite{zhang2024ram} would leave the anchored risk unchanged; only the anchored set sizes and the native results would change. Of the native thresholds in \cref{tab:adaptivemain}, the one reused from the source names exceeds the target under the worst rule for every model, as does the one calibrated on the source names plus $500$ random added names (column \emph{random}). Recalibrating at all grown names meets the target for CAT-Seg ViT-L but exceeds it for FC-CLIP ($\AdaptFcAdvRecalRisk\%$) and SED ($\AdaptSedAdvRecalRisk\%$). Recalibrating CAT-Seg ViT-L at a \AdaptLexSize-name lexicon gives $\AdaptAdvLexRisk\%$ risk at $\AdaptAdvLexSize$ labels per pixel, about twice the anchored set size. At the lexicon, the FC-CLIP threshold meets the target ($\AdaptFcAdvLexRisk\%$) at $\AdaptFcAdvLexSize$ labels per pixel against $\AdaptFcAdvAnchSize$ for the anchored threshold, and the SED threshold exceeds it further ($\AdaptSedAdvLexRisk\%$). On Pascal Context the threshold recalibrated at all $458$ names is at the target under the worst rule, with $\AdaptPasAdvRecalAbove\%$ of the splits above it; for FC-CLIP on ADE20K $\AdaptFcAdvRecalAbove\%$ of the splits lie above it.
\begin{table}[tp]
  \centering
  \caption{\textbf{Per-image vocabularies, CAT-Seg ViT-L} (ADE20K, $\alpha=10\%$): source-label risk (\%) / set size on source-labelled pixels. Native thresholds: reused, recalibrated at $845$ names or at a \AdaptLexSize-name lexicon, calibrated on vocabularies of the source names plus $500$ random added names (\emph{random}), or calibrated under the row's own rule (matched; \cref{prop:adaptive} for the worst case).}
  \label{tab:adaptive}
  \adjustbox{max width=\linewidth}{\footnotesize\setlength{\tabcolsep}{4pt}\setlength{\tabcolsep}{4pt}\begin{tabular}{@{}lrrrrrr@{}}
\toprule
& \multicolumn{1}{c}{anchored} & \multicolumn{5}{c}{native} \\ \cmidrule(lr){2-2}\cmidrule(l){3-7}Added names & \multicolumn{1}{c}{once} & \multicolumn{1}{c}{reused} & \multicolumn{1}{c}{recal.\ 845} & \multicolumn{1}{c}{random} & \multicolumn{1}{c}{recal.\ 17k} & \multicolumn{1}{c}{matched} \\
\midrule
present in image & 9.9/2.3 & 10.3/1.9 & 6.2/2.9 & 6.8/2.6 & 2.3/6.0 & 9.9/2.0 \\
50 random & 9.9/2.4 & 10.3/2.0 & 6.2/2.9 & 6.8/2.7 & 2.2/6.5 & 9.9/2.0 \\
200 random & 9.9/2.8 & 11.8/2.0 & 7.3/3.1 & 8.0/2.9 & 2.9/7.3 & 9.9/2.4 \\
500 random & 9.9/3.7 & 14.1/2.1 & 9.1/3.4 & 9.9/3.1 & 3.9/8.3 & 9.9/3.1 \\
tagger, top 50 & 9.9/4.2 & 14.8/2.2 & 9.4/3.7 & 10.3/3.3 & 4.0/9.2 & 9.9/3.5 \\
worst, per image & 9.9/4.2 & 14.9/2.2 & 9.5/3.6 & 10.4/3.3 & 4.1/8.9 & 9.9/3.4 \\
\bottomrule
\end{tabular}}}
\end{table}
\Cref{tab:adaptivesed} repeats the experiment for SED in its default mode, on the same \AdaptSedImages\ ADE20K images and with the same rules; the added names of the tagger rule are those that SED scores highest. The anchored threshold again has $\AdaptSedAdvAnchRisk\%$ risk under every rule. The native threshold reused from $150$ names stays below the target when random names are added ($\AdaptSedRandFiftyFrozenRisk\%$ with $50$, $\AdaptSedRandFiveHundredFrozenRisk\%$ with $500$), as in \cref{tab:main}: at a fixed threshold, SED misses fewer source pixels as random names are added. It exceeds the target when the added names are those annotated in the image ($\AdaptSedPresentFrozenRisk\%$) or those SED scores highest ($\AdaptSedTaggerFrozenRisk\%$). For CAT-Seg, the native threshold recalibrated at the $845$ grown names meets the target under every tested rule. For SED it exceeds the target under every rule ($\AdaptSedRandFiveHundredRecalRisk$--$\AdaptSedAdvRecalRisk\%$): it is calibrated where SED misses least, and each per-image vocabulary is a subset of the grown one. At a lexicon the threshold moves further in the same direction: under the worst rule the risk rises from $\AdaptSedAdvLexOneRisk\%$ at \AdaptSedLexSmall\ names to $\AdaptSedAdvLexFourRisk\%$ at \AdaptSedLexSize\ names, with $\AdaptSedAdvLexSize$ labels per pixel (column \emph{recal.\ 17k}). Recalibrating at a lexicon therefore relies on how the model's scores move as names are added, which differs between architectures. The native thresholds that meet the target under every rule are those calibrated under the deployed rule, or under the worst case over all rules (\cref{prop:adaptive}), which gives $\AdaptSedAdvWorstSize$ labels per pixel against $\AdaptSedAdvAnchSize$ for the anchored threshold.
\begin{table}[tp]
  \centering
  \caption{\textbf{Per-image vocabularies, SED} (ADE20K, $\alpha=10\%$): source-label risk (\%) / set size on source-labelled pixels, as in \cref{tab:adaptive}. \emph{matched}: native threshold calibrated under the row's own rule; for the last row, under the worst case over all rules.}
  \label{tab:adaptivesed}
  \adjustbox{max width=\linewidth}{\footnotesize\setlength{\tabcolsep}{4pt}\setlength{\tabcolsep}{4pt}\begin{tabular}{@{}lrrrrrr@{}}
\toprule
& \multicolumn{1}{c}{anchored} & \multicolumn{5}{c}{native} \\ \cmidrule(lr){2-2}\cmidrule(l){3-7}Added names & \multicolumn{1}{c}{once} & \multicolumn{1}{c}{reused} & \multicolumn{1}{c}{recal.\ 845} & \multicolumn{1}{c}{random} & \multicolumn{1}{c}{recal.\ 17k} & \multicolumn{1}{c}{matched} \\
\midrule
present in image & 9.9/2.9 & 10.4/2.3 & 15.5/1.6 & 15.1/1.6 & 21.7/1.2 & 9.9/2.4 \\
50 random & 9.9/3.0 & 9.0/2.6 & 13.6/1.8 & 13.3/1.8 & 19.4/1.4 & 9.9/2.4 \\
200 random & 9.9/3.7 & 8.0/4.0 & 11.4/2.6 & 11.1/2.7 & 15.9/1.8 & 9.9/3.0 \\
500 random & 9.9/5.1 & 7.3/6.8 & 10.1/4.2 & 9.9/4.3 & 13.8/2.8 & 9.9/4.3 \\
tagger, top 50 & 9.9/5.9 & 12.7/3.4 & 18.5/2.1 & 18.1/2.1 & 25.0/1.4 & 9.9/4.8 \\
worst, per image & 9.9/5.9 & 12.8/3.4 & 18.8/2.0 & 18.4/2.1 & 25.4/1.4 & 9.9/5.0 \\
\bottomrule
\end{tabular}}}
\end{table}
\Cref{tab:adaptivefcclip} repeats the experiment for FC-CLIP, whose tagger rule adds the names FC-CLIP scores highest. The anchored threshold has $\AdaptFcAdvAnchRisk\%$ risk under every rule. As for CAT-Seg, the reused native threshold exceeds the target under every rule ($\AdaptFcRandFiftyFrozenRisk\%$ with $50$ random names, $\AdaptFcAdvFrozenRisk\%$ in the worst case). The native threshold recalibrated at the $845$ grown names lies between CAT-Seg and SED. It is below the target for random names ($\AdaptFcRandFiftyRecalRisk$--$\AdaptFcRandFiveHundredRecalRisk\%$) and for the names present in the image ($\AdaptFcPresentRecalRisk\%$). It is slightly above the target for the tagger rule ($\AdaptFcTaggerRecalRisk\%$, $\AdaptFcTaggerRecalAbove\%$ of the splits above it) and in the worst case ($\AdaptFcAdvRecalRisk\%$, $\AdaptFcAdvRecalAbove\%$ of the splits). Recalibrated at a lexicon it meets the target under every rule: the worst-rule risk falls from $\AdaptFcAdvLexOneRisk\%$ at \AdaptFcLexSmall\ names to $\AdaptFcAdvLexFourRisk\%$ at \AdaptFcLexSize\ names, while the set size grows from $\AdaptFcAdvLexOneSize$ to $\AdaptFcAdvLexSize$ labels per pixel, against $\AdaptFcAdvAnchSize$ for the anchored threshold (column \emph{recal.\ 17k}). FC-CLIP's source-label scores mostly fall as names are added, so a threshold set at the lexicon admits more labels on every per-image vocabulary; SED's rise, and the same recalibration fails. As for SED, the native threshold calibrated under the worst case over all rules gives smaller sets than the anchored one ($\AdaptFcAdvWorstSize$ against $\AdaptFcAdvAnchSize$ labels per pixel), but it requires knowing the rules.
\begin{table}[tp]
  \centering
  \caption{\textbf{Per-image vocabularies, FC-CLIP} (ADE20K, $\alpha=10\%$): source-label risk (\%) / set size on source-labelled pixels, as in \cref{tab:adaptivesed}.}
  \label{tab:adaptivefcclip}
  \adjustbox{max width=\linewidth}{\footnotesize\setlength{\tabcolsep}{4pt}\setlength{\tabcolsep}{4pt}\begin{tabular}{@{}lrrrrrr@{}}
\toprule
& \multicolumn{1}{c}{anchored} & \multicolumn{5}{c}{native} \\ \cmidrule(lr){2-2}\cmidrule(l){3-7}Added names & \multicolumn{1}{c}{once} & \multicolumn{1}{c}{reused} & \multicolumn{1}{c}{recal.\ 845} & \multicolumn{1}{c}{random} & \multicolumn{1}{c}{recal.\ 17k} & \multicolumn{1}{c}{matched} \\
\midrule
present in image & 9.9/4.0 & 10.5/3.2 & 7.6/4.7 & 8.0/4.4 & 4.3/10.4 & 9.9/3.4 \\
50 random & 9.9/4.3 & 10.3/3.4 & 7.3/5.1 & 7.8/4.8 & 4.1/11.9 & 9.9/3.6 \\
200 random & 9.9/5.9 & 11.6/3.9 & 8.2/6.0 & 8.7/5.6 & 4.6/14.7 & 9.9/4.8 \\
500 random & 9.9/9.0 & 13.0/4.7 & 9.4/7.4 & 9.9/6.9 & 5.1/19.3 & 9.9/6.9 \\
tagger, top 50 & 9.9/9.2 & 14.0/4.9 & 10.2/7.4 & 10.7/6.9 & 5.6/15.9 & 9.9/7.6 \\
worst, per image & 9.9/9.7 & 14.1/4.8 & 10.3/7.4 & 10.8/6.9 & 5.6/17.4 & 9.9/7.7 \\
\bottomrule
\end{tabular}}}
\end{table}
\paragraph{Pascal Context.}\label{sec:adaptivepascal}
\Cref{tab:adaptivepascal} repeats the experiment for CAT-Seg ViT-L and SED on the \AdaptPasImages\ Pascal Context validation images, whose $59$ source names grow to $458$, under a protocol fixed before the per-image scores were computed. Each rule adds one of: the names annotated in the image ($\AdaptPasPresentNames$ on average); $50$, $100$ or $300$ of the $399$ added names at random ($300$ is about the same share of the added names as $500$ on ADE20K); the $50$ added names the model scores highest; or the worst of these per image. The anchored threshold has $\AdaptPasAdvAnchRisk\%$ risk under every rule for both models. For CAT-Seg, the reused native threshold exceeds the target under every rule ($\AdaptPasPresentFrozenRisk$--$\AdaptPasAdvFrozenRisk\%$). The threshold recalibrated at all $458$ names is below the target for random names and the names present in the image ($\AdaptPasPresentRecalRisk$--$\AdaptPasRandThreeHundredRecalRisk\%$) and at the target under the tagger rule and in the worst case ($\AdaptPasAdvRecalRisk\%$, $\AdaptPasAdvRecalAbove\%$ of the splits above it). For SED, the reused threshold stays below the target for random names ($\AdaptPasSedRandFiftyFrozenRisk\%$ with $50$, $\AdaptPasSedRandThreeHundredFrozenRisk\%$ with $300$) and for the names present in the image ($\AdaptPasSedPresentFrozenRisk\%$), and exceeds it under the tagger rule ($\AdaptPasSedTaggerFrozenRisk\%$) and in the worst case ($\AdaptPasSedAdvFrozenRisk\%$). Recalibrating at $458$ names exceeds the target under every rule ($\AdaptPasSedRandThreeHundredRecalRisk$--$\AdaptPasSedAdvRecalRisk\%$), as on ADE20K. The native threshold calibrated under the worst case over all rules gives $\AdaptPasAdvWorstSize$ and $\AdaptPasSedAdvWorstSize$ labels per pixel, against $\AdaptPasAdvAnchSize$ and $\AdaptPasSedAdvAnchSize$ for the anchored threshold. At $\alpha=5\%$, in the worst case, the anchored threshold has $\AdaptPasBAdvAnchRisk$ and $\AdaptPasSedBAdvAnchRisk\%$ risk, and recalibration at $458$ names $\AdaptPasBAdvRecalRisk$ and $\AdaptPasSedBAdvRecalRisk\%$.
\begin{table}[tp]
  \centering
  \caption{\textbf{Per-image vocabularies on Pascal Context} ($59\to458$, $\alpha=10\%$): source-label risk (\%) / set size on source-labelled pixels, as in \cref{tab:adaptivesed}. \emph{random}: calibrated on vocabularies of the source names plus $300$ random added names.}
  \label{tab:adaptivepascal}
  \adjustbox{max width=\linewidth}{\footnotesize\setlength{\tabcolsep}{4pt}\setlength{\tabcolsep}{4pt}\begin{tabular}{@{}lrrrrr@{}}
\toprule
& \multicolumn{1}{c}{anchored} & \multicolumn{4}{c}{native} \\ \cmidrule(lr){2-2}\cmidrule(l){3-6}Added names & \multicolumn{1}{c}{once} & \multicolumn{1}{c}{reused} & \multicolumn{1}{c}{recal.\ 458} & \multicolumn{1}{c}{random} & \multicolumn{1}{c}{matched} \\
\midrule
\multicolumn{6}{@{}l}{\emph{CAT-Seg ViT-L}} \\
present in image & 9.8/1.5 & 10.1/1.4 & 6.0/1.7 & 6.6/1.7 & 9.8/1.4 \\
50 random & 9.8/1.6 & 10.9/1.4 & 6.4/1.8 & 7.1/1.7 & 9.8/1.5 \\
100 random & 9.8/1.7 & 11.5/1.4 & 7.0/1.8 & 7.7/1.7 & 9.8/1.5 \\
300 random & 9.8/2.1 & 14.0/1.5 & 8.9/2.0 & 9.8/1.9 & 9.8/1.9 \\
tagger, top 50 & 9.8/2.3 & 15.2/1.5 & 10.0/2.0 & 10.9/1.9 & 9.8/2.1 \\
worst, per image & 9.8/2.3 & 15.2/1.5 & 10.0/2.0 & 10.9/1.9 & 9.8/2.1 \\
\midrule
\multicolumn{6}{@{}l}{\emph{SED}} \\
present in image & 9.8/1.7 & 9.9/1.5 & 18.9/1.1 & 17.6/1.1 & 9.8/1.6 \\
50 random & 9.8/1.8 & 8.2/1.8 & 15.2/1.2 & 14.2/1.3 & 9.8/1.6 \\
100 random & 9.8/2.0 & 7.3/2.1 & 13.7/1.4 & 12.8/1.4 & 9.8/1.7 \\
300 random & 9.8/2.5 & 5.0/3.9 & 10.6/2.2 & 9.8/2.3 & 9.8/2.3 \\
tagger, top 50 & 9.8/2.8 & 11.8/2.1 & 19.4/1.3 & 18.4/1.3 & 9.8/2.4 \\
worst, per image & 9.8/2.8 & 12.3/1.9 & 21.1/1.1 & 19.8/1.2 & 9.8/2.4 \\
\bottomrule
\end{tabular}}}
\end{table}
\paragraph{Names from a language model.}\label{sec:freeform}
The rules above draw the added names from the $845$ evaluated names. A tagger or language model in deployment is not limited to a list. Under a protocol fixed before any score was computed, we asked Claude Opus 5.5 to list what it sees in each of the \FfImages\ ADE20K images, without showing it any list of names. It gave $\FfAnswered$ names per image. After removing the source names, their ADE20K synonyms and repeats, $\FfAdded$ added names remain per image, \FfUnique\ distinct names in all; $\FfOutsideAde\%$ of the added names are not among the ADE20K-847 names and $\FfOutsideLex\%$ are not in the $17{,}001$-name lexicon. Each image deploys the $150$ source names followed by its own added names. \Cref{tab:freeform} gives the source-label risk. The anchored threshold, calibrated once at the source names, has $\FfAnchRiskLo\%$ risk for every model (\cref{prop:transfer}). The native threshold reused from the source names has $\FfReusedRiskLo$--$\FfReusedRiskHi\%$ risk, and the one recalibrated at all $845$ names $\FfRecalRiskLo$--$\FfRecalRiskHi\%$; both exceed the target for every model in every split. For SED, recalibrating at a lexicon moves the risk further from the target ($\FfSedLexRisk\%$ at $17{,}001$ names), as under the fixed rules. For FC-CLIP, the threshold recalibrated at the $17{,}001$-name lexicon meets the target with $\FfFcLexSize$ labels per pixel, $\FfFcLexRatio$ times the anchored set size, although $\FfOutsideLex\%$ of the added names are outside the lexicon, where no guarantee applies. Calibrating the native model under the language model's names on the calibration images is valid by exchangeability, but needs the language model at calibration time and again whenever the model or the prompt changes; its sets are larger than the anchored ones for every model ($\FfVitLMatchedSize$, $\FfSedMatchedSize$ and $\FfFcMatchedSize$ against $\FfVitLAnchSize$, $\FfSedAnchSize$ and $\FfFcAnchSize$ labels per pixel). Names outside ADE20K-847 have no annotations, so only the source-label risk is evaluated; one language model and one prompt were used.
\begin{table}[tp]
  \centering
  \caption{\textbf{Names from a language model} (ADE20K, \FfImages\ images; the $150$ source names plus $\FfAdded$ names per image on average from Claude Opus 5.5, most outside ADE20K-847): source-label risk (\%) / set size on source-labelled pixels, $500$ random half/half splits. Native thresholds: reused from the source names, recalibrated at the $845$ names or at a lexicon of $1{,}690$ or $17{,}001$ names (SED and FC-CLIP), or calibrated under the language model's names on the calibration images (matched).}
  \label{tab:freeform}
  \adjustbox{max width=\linewidth}{\footnotesize\setlength{\tabcolsep}{4pt}\setlength{\tabcolsep}{4pt}\begin{tabular}{@{}lrrrrrr@{}}
\toprule
& \multicolumn{1}{c}{anchored} & \multicolumn{5}{c}{native} \\ \cmidrule(lr){2-2}\cmidrule(l){3-7}Model & \multicolumn{1}{c}{once} & \multicolumn{1}{c}{reused} & \multicolumn{1}{c}{recal.\ 845} & \multicolumn{1}{c}{recal.\ 1.7k} & \multicolumn{1}{c}{recal.\ 17k} & \multicolumn{1}{c}{matched} \\
\midrule
\multicolumn{7}{@{}l}{$\alpha=10\%$} \\
CAT-Seg ViT-L & 9.9/4.6 & 20.6/2.7 & 12.6/4.1 & -- & -- & 9.9/4.8 \\
SED & 9.9/5.7 & 20.5/3.4 & 30.3/2.3 & 35.4/1.9 & 41.2/1.5 & 9.9/6.4 \\
FC-CLIP & 9.9/8.0 & 21.5/4.4 & 15.6/5.9 & 13.6/6.6 & 7.8/11.0 & 9.9/8.8 \\
\midrule
\multicolumn{7}{@{}l}{$\alpha=5\%$} \\
CAT-Seg ViT-L & 4.9/7.0 & 10.2/4.7 & 5.1/7.3 & -- & -- & 4.9/7.5 \\
SED & 4.9/9.8 & 8.9/7.0 & 12.7/5.2 & 15.5/4.4 & 18.9/3.7 & 4.9/10.9 \\
FC-CLIP & 4.9/15.1 & 10.4/8.5 & 7.0/12.2 & 6.2/13.6 & 4.1/22.0 & 4.9/18.0 \\
\bottomrule
\end{tabular}}}
\end{table}

\section{Added labels}
\label{sec:addedfull}
This section extends \cref{sec:labelguarantee}. \Cref{sec:labelmodels} repeats the label-level analysis for every model. \Cref{sec:weights} analyses the benchmark's own source classes, including calibration on randomly drawn added names. \Cref{sec:added} evaluates the added labels under the image-level thresholds of \cref{sec:main}, which carry no guarantee for them. Per-name thresholds (\cref{sec:pername}) bound the risk of each added name with enough labelled images, without exchangeable labels.

\subsection{Every model}
\label{sec:labelmodels}
\Cref{tab:labelmodels} applies the label-level threshold of \cref{prop:labels} to every model with a random source vocabulary, drawn afresh in each trial. \LxModelsPending The added-label risk is $\LxModelsMinRisk$--$\LxModelsMaxRisk\%$ at $\alpha=10\%$ and $\LxModelsBMinRisk$--$\LxModelsBMaxRisk\%$ at $\alpha=5\%$. The label-level threshold on the native model at the grown vocabulary is also valid, and the anchored sets are smaller in every comparison except FC-CLIP on COCO-Stuff (\cref{tab:labelmodels}). On the benchmark's own split every model shows an excess of the kind analysed in \cref{sec:weights}.
\begin{table*}[tp]
  \centering
  \caption{\textbf{Added-label risk for every model} (\%) / set size at the grown vocabulary, label-level threshold, \LxTrials\ trials. Random source vocabulary: anchored model calibrated at the source vocabulary, and native model calibrated at the grown vocabulary. Benchmark's own: anchored model, with $\Delta(\hat t)$ of \cref{prop:shift} in points.}
  \label{tab:labelmodels}
  \adjustbox{max width=\textwidth}{\footnotesize\setlength{\tabcolsep}{4pt}\begin{tabular}{@{}llrrrrrr@{}}
\toprule
& & \multicolumn{4}{c}{random source vocabulary} & \multicolumn{2}{c}{benchmark\rq{}s own, $\alpha=.10$} \\
\cmidrule(lr){3-6}\cmidrule(l){7-8}
& & \multicolumn{2}{c}{$\alpha=.10$} & \multicolumn{2}{c}{$\alpha=.05$} & & \\
\cmidrule(lr){3-4}\cmidrule(lr){5-6}
Model & benchmark & anchored & native recal. & anchored & native recal. & anchored & $\Delta(\hat t)$ \\
\midrule
CAT-Seg ViT-B & ADE20K & 9.5 / 46.4 & 9.6 / 59.5 & 4.5 / 102.6 & 3.4 / 430.1 & 22.6 & 13.2 \\
CAT-Seg ViT-B & Pascal Context & 7.5 / 23.9 & 7.6 / 32.1 & 2.6 / 98.4 & 2.5 / 160.7 & 27.4 & 19.0 \\
CAT-Seg ViT-B & COCO-Stuff & 9.1 / 21.7 & 9.1 / 26.6 & 4.1 / 58.8 & 4.0 / 177.3 & 22.5 & 13.1 \\
CAT-Seg ViT-L & ADE20K & 9.5 / 27.6 & 9.5 / 34.2 & 4.5 / 76.4 & 4.0 / 260.1 & 22.2 & 12.8 \\
CAT-Seg ViT-L & Pascal Context & 7.6 / 11.5 & 7.6 / 14.9 & 2.8 / 68.4 & 2.7 / 92.8 & 22.3 & 13.9 \\
CAT-Seg ViT-L & COCO-Stuff & 8.9 / 11.9 & 9.0 / 13.6 & 4.0 / 30.2 & 4.0 / 42.1 & 17.6 & 8.2 \\
CAT-Seg + DeCLIP & ADE20K & 9.5 / 19.0 & 9.4 / 22.2 & 4.5 / 52.3 & 4.2 / 194.3 & 20.7 & 11.3 \\
CAT-Seg + DeCLIP & Pascal Context & 7.5 / 10.4 & 7.6 / 17.6 & 2.8 / 62.4 & 2.8 / 132.0 & 18.1 & 9.6 \\
CAT-Seg + DeCLIP & COCO-Stuff & 9.0 / 9.5 & 9.1 / 10.1 & 4.1 / 22.5 & 4.1 / 46.8 & 14.6 & 5.2 \\
SED & ADE20K & 9.5 / 45.1 & 9.5 / 51.4 & 4.6 / 118.5 & 4.6 / 139.8 & 22.2 & 12.8 \\
SED & Pascal Context & 7.5 / 17.5 & 7.6 / 20.1 & 2.8 / 80.4 & 2.8 / 86.9 & 26.2 & 17.8 \\
SED & COCO-Stuff & 9.0 / 21.2 & 9.1 / 23.9 & 4.0 / 60.1 & 4.1 / 66.6 & 24.3 & 14.9 \\
FC-CLIP & ADE20K & 9.6 / 61.0 & 9.6 / 65.6 & 4.5 / 142.2 & 4.6 / 152.4 & 16.4 & 7.0 \\
FC-CLIP & Pascal Context & 7.5 / 43.4 & 7.5 / 49.7 & 2.5 / 132.3 & 2.5 / 158.7 & 22.9 & 14.5 \\
FC-CLIP & COCO-Stuff & 9.0 / 82.9 & 9.1 / 66.1 & 4.0 / 227.6 & 4.1 / 186.4 & 16.6 & 7.2 \\
\bottomrule
\end{tabular}}}
\end{table*}
\Cref{prop:labels} bounds the added-label risk on the calibration images. \Cref{tab:labelheld} compares it with the risk of the same added labels on the held-out half of the images, which a deployment would see: the two differ by at most $\LxModelsHeldMaxGap$ points for every model and benchmark. Since the bound holds in expectation over the label draw, a single draw can exceed $\alpha$: $\LxModelsMinAbove$--$\LxModelsMaxAbove\%$ of the draws do at $\alpha=10\%$ ($\LxModelsHeldMinAbove$--$\LxModelsHeldMaxAbove\%$ on held-out images) and $\LxModelsBMinAbove$--$\LxModelsBMaxAbove\%$ at $\alpha=5\%$.
\begin{table*}[tp]
  \centering
  \caption{\textbf{Added-label risk on calibration and held-out images} (\%), anchored model, label-level threshold, random source vocabulary, \LxTrials\ trials: mean risk, and the percentage of draws whose risk exceeds $\alpha$. \emph{cal.}: the calibration images, on which \cref{prop:labels} holds; \emph{held-out}: the other half of the images.}
  \label{tab:labelheld}
  \adjustbox{max width=\textwidth}{\footnotesize\setlength{\tabcolsep}{4pt}\begin{tabular}{@{}llrrrrrrrr@{}}
\toprule
& & \multicolumn{4}{c}{$\alpha=.10$} & \multicolumn{4}{c}{$\alpha=.05$} \\
\cmidrule(lr){3-6}\cmidrule(l){7-10}
& & \multicolumn{2}{c}{risk} & \multicolumn{2}{c}{draws $>\alpha$} & \multicolumn{2}{c}{risk} & \multicolumn{2}{c}{draws $>\alpha$} \\
\cmidrule(lr){3-4}\cmidrule(lr){5-6}\cmidrule(lr){7-8}\cmidrule(l){9-10}
Model & benchmark & cal. & held-out & cal. & held-out & cal. & held-out & cal. & held-out \\
\midrule
CAT-Seg ViT-B & ADE20K & 9.5 & 9.6 & 39 & 39 & 4.5 & 4.5 & 34 & 35 \\
CAT-Seg ViT-B & Pascal Context & 7.5 & 7.5 & 25 & 26 & 2.6 & 2.6 & 14 & 15 \\
CAT-Seg ViT-B & COCO-Stuff & 9.1 & 9.2 & 36 & 37 & 4.1 & 4.1 & 29 & 28 \\
CAT-Seg ViT-L & ADE20K & 9.5 & 9.5 & 40 & 40 & 4.5 & 4.5 & 36 & 37 \\
CAT-Seg ViT-L & Pascal Context & 7.6 & 7.5 & 23 & 25 & 2.8 & 2.7 & 16 & 16 \\
CAT-Seg ViT-L & COCO-Stuff & 8.9 & 9.0 & 33 & 34 & 4.0 & 4.1 & 27 & 28 \\
CAT-Seg + DeCLIP & ADE20K & 9.5 & 9.4 & 39 & 38 & 4.5 & 4.5 & 34 & 34 \\
CAT-Seg + DeCLIP & Pascal Context & 7.5 & 7.5 & 25 & 25 & 2.8 & 2.8 & 19 & 19 \\
CAT-Seg + DeCLIP & COCO-Stuff & 9.0 & 9.1 & 35 & 33 & 4.1 & 4.1 & 27 & 29 \\
SED & ADE20K & 9.5 & 9.5 & 42 & 42 & 4.6 & 4.6 & 36 & 37 \\
SED & Pascal Context & 7.5 & 7.5 & 25 & 27 & 2.8 & 2.8 & 17 & 19 \\
SED & COCO-Stuff & 9.0 & 9.1 & 34 & 36 & 4.0 & 4.1 & 27 & 29 \\
FC-CLIP & ADE20K & 9.6 & 9.5 & 40 & 40 & 4.5 & 4.5 & 36 & 36 \\
FC-CLIP & Pascal Context & 7.5 & 7.5 & 23 & 22 & 2.5 & 2.6 & 12 & 13 \\
FC-CLIP & COCO-Stuff & 9.0 & 9.0 & 32 & 33 & 4.0 & 4.0 & 26 & 28 \\
\bottomrule
\end{tabular}}}
\end{table*}

\subsection{The benchmark's own source classes}
\label{sec:weights}
\paragraph{Decomposition.} On the calibration images, the added-label risk equals the mean loss of the source labels at the calibrated threshold plus $\Delta(\hat t)$, the difference in \cref{prop:shift}. At $\alpha=10\%$ the source labels are at or below the target on every benchmark, and the gap $\Delta(\hat t)$ accounts for the added-label excess (\cref{sec:labelguarantee}). Native recalibration at the grown vocabulary shows the same pattern, with an added-label risk of $\LxCocSplitRecalRisk$--$\LxPasSplitRecalRisk\%$ across the benchmarks ($\Delta(\hat t)=\LxDAdeSplitRecalDelta$ points on ADE20K). For anchored CAT-Seg ViT-L on ADE20K, the worst-case bound $\max_t\Delta(t)$ is $\LxDAdeSplitLabDeltaRatio$ times the realised excess ($\LxDAdeSplitLabDeltaMax$ against $\LxDAdeSplitLabDelta$ points). For the native model with the threshold reused from the source vocabulary, $\Delta(\hat t)$ also contains candidate dependence: it is $\LxDAdeExchFrozenDelta$ points on ADE20K even under a random source vocabulary, and $\LxDAdeSplitFrozenDelta$ points on the benchmark's own split, roughly the sum of the two effects.

\paragraph{Added names selected by difficulty.}\label{sec:hardness}
\Cref{tab:hardness} measures the excess of \cref{prop:shift} when the added names are selected by how hard they are. Each of \HdTrials\ trials follows the exchangeable design of \cref{sec:labelmodels}: a random calibration half of the ADE20K images and $150$ uniformly drawn calibration names. The remaining names present in both halves are ranked by their loss on the held-out half, at the label-level threshold calibrated on that half, and split into fifths; $Q$ is uniform over one fifth. The ranking uses the held-out images and the drawn names, which also determine $\hat t_\ell$, so $Q$ is not independent of $\hat t_\ell$ as \cref{prop:shift} assumes. The reported bounds do not need this independence. Let $P$ be uniform over all names present in the calibration images, source names included. For any $Q$, $\bar\ell_Q(\hat t_\ell)=\bar\ell_P(\hat t_\ell)+\Delta(\hat t_\ell)$ in every realisation. The CRC constraint bounds the mean loss of the source names by $\alpha$, and \cref{prop:labels}, applied to a uniformly chosen remaining name, bounds the expected mean loss of the remaining names by $\alpha$, so $\mathbb E[\bar\ell_P(\hat t_\ell)]\le\alpha$. The columns $\alpha+\Delta(\hat t_\ell)$ and $\alpha+\max\Delta$ therefore bound the expected risk, and so does $\alpha+\varepsilon$ below. All risks are measured on the calibration images.

Uniform over all remaining names, the added-label risk for CAT-Seg ViT-L at $\alpha=10\%$ is $\HdVitLQAllRisk\%$. The four easier fifths have mean risk below the target, whereas the hardest fifth reaches $\HdVitLQFiveRisk\%$ ($\HdSedQFiveRisk\%$ for SED), within the worst-case bound $\alpha+\max_t\Delta(t)=\HdVitLQFiveBound\%$. The guarantee thus does not extend to names chosen because they are difficult. 

When $P$ is contaminated with a fraction $\varepsilon$ of the hardest fifth, the risk grows linearly in $\varepsilon$ from $\HdVitLPRisk\%$ at $\varepsilon=0$, with slope equal to the gap between the hardest fifth and $P$. It stays well inside the bound $\alpha+\varepsilon$: at $\varepsilon=0.1$ it is $\HdVitLEpsTenRisk\%$ against $20\%$. At $\alpha=5\%$ the hardest fifth reaches $\HdVitLBQFiveRisk\%$ for ViT-L and $\HdSedBQFiveRisk\%$ for SED.
\begin{table*}[tp]
  \centering
  \caption{\textbf{Added names selected by difficulty} (ADE20K $150\!\to\!845$, $\alpha=10\%$, \HdTrials\ trials of the exchangeable design, label-level threshold $\hat t_\ell$ from $150$ uniformly drawn names). Added names: uniform over one fifth of the remaining names, ranked by their loss on the held-out images (the first three fifths have held-out loss of at most $\HdFirstThreeMaxHard\%$, so their order is largely arbitrary), or over all of them. Held-out: that loss (\%). Risk: added-label risk on the calibration images (\%), next to the bound of \cref{prop:shift} at the realised threshold and its worst case. Bottom: the uniform draw $P$ over all names present in the calibration images, source names included, contaminated with a fraction $\varepsilon$ of the hardest fifth.}
  \label{tab:hardness}
  \adjustbox{max width=\textwidth}{\footnotesize\setlength{\tabcolsep}{4pt}\setlength{\tabcolsep}{4pt}\begin{tabular}{@{}lrrrrrrrr@{}}
\toprule
& \multicolumn{4}{c}{CAT-Seg ViT-L} & \multicolumn{4}{c}{SED} \\ \cmidrule(lr){2-5}\cmidrule(l){6-9}
Added names & held-out & risk & $\alpha+\Delta(\hat t_\ell)$ & $\alpha+\max\Delta$ & held-out & risk & $\alpha+\Delta(\hat t_\ell)$ & $\alpha+\max\Delta$ \\
\midrule
fifth 1 (easiest) & 0.0 & 1.7 & 2.3 & 10.0 & 0.0 & 1.3 & 1.8 & 10.0 \\
fifth 2 & 0.0 & 6.1 & 6.7 & 13.5 & 0.0 & 6.3 & 6.9 & 12.9 \\
fifth 3 & 0.3 & 4.4 & 5.0 & 10.7 & 0.2 & 5.2 & 5.7 & 11.4 \\
fifth 4 & 3.3 & 6.2 & 6.8 & 12.7 & 3.3 & 6.1 & 6.7 & 15.1 \\
fifth 5 (hardest) & 39.6 & 24.9 & 25.4 & 36.1 & 41.9 & 25.6 & 26.2 & 38.5 \\
\midrule all remaining & 8.6 & 8.7 & 9.2 & 10.3 & 9.0 & 8.9 & 9.4 & 10.6 \\
\midrule
\multicolumn{9}{@{}l}{\emph{$(1-\varepsilon)$ uniform draw $+\ \varepsilon$ hardest fifth: risk, and the bound $\alpha+\varepsilon$}} \\
$\varepsilon=0.05$ &  & 10.2 & 15 &  &  & 10.3 & 15 &  \\
$\varepsilon=0.1$ &  & 11.0 & 20 &  &  & 11.1 & 20 &  \\
$\varepsilon=0.25$ &  & 13.3 & 35 &  &  & 13.5 & 35 &  \\
$\varepsilon=0.5$ &  & 17.2 & 60 &  &  & 17.5 & 60 &  \\
\bottomrule
\end{tabular}}}
\end{table*}

\paragraph{Prevalence weights.}
If source and added labels differ only through a label feature that can be computed without annotations, weighted conformal prediction \cite{tibshirani2019weighted} restores a guarantee. As the feature we take the predicted prevalence of a label: the number of sampled pixels, over all images, at which its anchored score exceeds a logit of $0$. We divide it into $B$ bins, at quantiles over all names or with equal numbers of source names per bin, and weight each source label in bin $b$ by $w_b = (\text{share of added names in } b)/(\text{share of source names in } b)$. For an added label in bin $b$ the threshold is the largest $t$ with $\big(\sum_i w_i\ell_{c_i}(t)+w_b\big)/\big(\sum_i w_i+w_b\big)\le\alpha$; if $w_b/(\sum_i w_i+w_b)>\alpha$ no such $t$ exists and the label receives the all-label threshold. As a diagnostic we also use the true prevalence, which is not available at deployment.

\Cref{tab:weights,tab:weightscoco} give the results for anchored CAT-Seg ViT-L. With predicted prevalence, three bins over all names reach the target on ADE20K at moderate set size ($\LxWAdePAllThreeRisk\%$ at $\LxWAdePAllThreeSize$ labels per pixel). The same rule, however, assigns the all-label threshold to many added labels at $\alpha=5\%$ and on Pascal Context, whose $59$ source labels are too few per bin, and it is over-conservative on COCO-Stuff. Weights from true prevalence do not improve the results either, suggesting that prevalence alone does not capture the difference in difficulty between source and added classes. The binning rules were explored on ADE20K and then run once on Pascal Context and COCO-Stuff.

\paragraph{Annotating randomly drawn added names.}
A deployment can instead restore exchangeability by construction. It draws $m$ names uniformly at random from the names that may be added, annotates the calibration images for them, and calibrates the label-level threshold of \cref{prop:labels} on the $m'$ drawn names that occur in these images. Given $m'$, these form a uniformly random subset of the added labels present, so a remaining added label is exchangeable with them and \cref{prop:labels} applies. The source classes keep the image-level threshold, whose guarantee is unaffected, since with candidate-independent scores the loss at a pixel depends only on the threshold of its true label. \Cref{tab:labelsample} gives results for CAT-Seg ViT-L and SED on ADE20K, Pascal Context and COCO-Stuff, with a fresh image split and draw in each of \LxTrials\ trials and the drawn names nested across $m$. \LsPending Mean added-label risk on the calibration images stays below the target for every tested $m$ in every run. At $m=100$ and $\alpha=10\%$, ViT-L gives $\LsAdeVitLMHundredRisk\%$ risk on calibration images and $\LsAdeVitLMHundredHeld\%$ on held-out images, with $\LsAdeVitLMHundredSize$ labels per pixel against $\LsAdeVitLMHundredNatSize$ for native recalibration. The held-out result is empirical; the fixed-image guarantee concerns the calibration images. 

Set sizes at this budget approach the random-source references: $\LsAdeVitLMHundredSize$ versus $\LsAdeVitLRandSize$ for ViT-L and $\LsAdeSedMHundredSize$ versus $\LsAdeSedRandSize$ for SED. This is budget-dependent: at $m=20$, sizes are $\LsAdeVitLMTwentySize$ and $\LsAdeSedMTwentySize$, respectively. At $m=10$ and $\alpha=10\%$, the finite-sample correction forces the all-label threshold in $\LsAdeVitLMTenFloor\%$ of ViT-L draws and $\LsAdeSedMTenFloor\%$ of SED draws. 

On Pascal Context and COCO-Stuff, fewer of the drawn names occur in the calibration images ($m'=\LsPasVitLMHundredPresent$ and $\LsCocVitLMHundredPresent$ at $m=100$), so small budgets reach the all-label threshold more often: at $m=20$, in $\LsPasVitLMTwentyFloor\%$ of Pascal and $\LsCocVitLMTwentyFloor\%$ of COCO-Stuff draws for ViT-L. At $m=100$, ViT-L gives $\LsPasVitLMHundredSize$ labels per pixel on Pascal Context and $\LsCocVitLMHundredSize$ on COCO-Stuff, against $\LsPasVitLMHundredNatSize$ and $\LsCocVitLMHundredNatSize$ for native recalibration, and SED gives $\LsPasSedMHundredSize$ on Pascal Context and $\LsCocSedMHundredSize$ on COCO-Stuff, against $\LsPasSedMHundredNatSize$ and $\LsCocSedMHundredNatSize$. The cost is annotating $m$ names on the calibration images, only $m'$ of which are present.
\begin{table*}[tp]
  \centering
  \caption{\textbf{Added-label guarantee from $m$ randomly drawn added names} (ADE20K $150\!\to\!845$, Pascal Context $59\!\to\!458$, COCO-Stuff $171\!\to\!1{,}283$; benchmark's own source classes, \LxTrials\ trials). $m'$: drawn names present in the calibration images. Risk: added-label risk (\%) on the calibration images, and percentage of draws above $\alpha$. Size: labels per pixel at the grown vocabulary, with the source classes at the image-level threshold and the added names at the label-level threshold; \emph{native}: the same calibration on the native model at the grown vocabulary. Reference rows: label-level threshold from the benchmark's source classes, and from a random source vocabulary (one threshold for all names, as in \cref{tab:labelmodels}).}
  \label{tab:labelsample}
  \adjustbox{max width=\textwidth}{\footnotesize\setlength{\tabcolsep}{4pt}\adjustbox{valign=t}{\setlength{\tabcolsep}{4pt}\begin{tabular}{@{}lrrrrrrr@{}}
\toprule
& & \multicolumn{4}{c}{$\alpha=.10$} & \multicolumn{2}{c}{$\alpha=.05$} \\ \cmidrule(lr){3-6}\cmidrule(l){7-8}
Calibration labels & $m'$ & risk & draws $>\alpha$ & size & native & risk & size \\
\midrule
\multicolumn{8}{@{}l}{\emph{CAT-Seg ViT-L, ADE20K}} \\
benchmark source classes & 149 & 22.2 & 100 & 7.6 & -- & 13.1 & 17.5 \\
random source vocabulary & 113 & 9.5 & 40 & 27.6 & -- & 4.5 & 76.4 \\
$m=10$ random added names & 7 & 0.3 & 0 & 627.7 & 666.8 & 0.0 & 698.7 \\
$m=20$ random added names & 14 & 5.8 & 17 & 102.1 & 335.6 & 0.0 & 695.2 \\
$m=30$ random added names & 21 & 7.0 & 22 & 60.3 & 232.3 & 1.9 & 297.7 \\
\bottomrule
\end{tabular}}}\hspace{14pt}\adjustbox{valign=t}{\setlength{\tabcolsep}{4pt}\begin{tabular}{@{}lrrrrrrr@{}}
\toprule
& & \multicolumn{4}{c}{$\alpha=.10$} & \multicolumn{2}{c}{$\alpha=.05$} \\ \cmidrule(lr){3-6}\cmidrule(l){7-8}
Calibration labels & $m'$ & risk & draws $>\alpha$ & size & native & risk & size \\
\multicolumn{8}{@{}l}{\emph{SED, ADE20K}} \\
benchmark source classes & 149 & 22.2 & 100 & 12.0 & -- & 12.7 & 30.7 \\
random source vocabulary & 113 & 9.5 & 42 & 45.1 & -- & 4.6 & 118.5 \\
$m=10$ random added names & 7 & 0.3 & 0 & 637.7 & 641.3 & 0.0 & 700.4 \\
$m=20$ random added names & 14 & 5.8 & 16 & 157.7 & 181.3 & 0.0 & 696.4 \\
$m=30$ random added names & 21 & 7.0 & 24 & 102.5 & 122.9 & 1.8 & 370.8 \\
\bottomrule
\end{tabular}}}}
\end{table*}
\begin{table*}[tp]
  \centering
  \caption{\textbf{Prevalence-weighted label-level CRC on the benchmark's own split} (anchored CAT-Seg ViT-L, \LxTrials\ trials): added-label risk (\%) / labels per pixel at the grown vocabulary; in parentheses, the share (\%) of added labels that receive the all-label threshold, where it is not zero. Predicted: prevalence from the anchored scores; true: from the annotations (a diagnostic).}
  \label{tab:weights}
  \adjustbox{max width=\textwidth}{\footnotesize\setlength{\tabcolsep}{4pt}\begin{tabular}{@{}llrrrr@{}}
\toprule
& & \multicolumn{2}{c}{ADE20K $150\!\to\!845$} & \multicolumn{2}{c}{Pascal Context $59\!\to\!458$} \\
\cmidrule(lr){3-4}\cmidrule(l){5-6}
Prevalence bins & weights from & $\alpha=.10$ & $\alpha=.05$ & $\alpha=.10$ & $\alpha=.05$ \\
\midrule
unweighted (label-level CRC) & -- & 22.2 / 9.0 & 13.1 / 20.6 & 22.3 / 2.7 & 12.6 / 5.0 \\
\midrule
all names, 2 bins & predicted & 10.1 / 26.8 & 5.1 / 58.9 & 8.1 / 232.6 (41) & 5.9 / 234.2 (41) \\
all names, 2 bins & true & 11.5 / 426.6 (56) & 7.4 / 428.8 (56) & 18.1 / 231.2 (14) & 10.5 / 232.8 (14) \\
all names, 3 bins & predicted & 9.6 / 27.4 & 1.9 / 322.4 (38) & 4.2 / 309.3 (65) & 2.9 / 310.5 (65) \\
all names, 3 bins & true & 4.3 / 569.5 (78) & 2.2 / 575.5 (78) & 17.5 / 307.5 (47) & 10.5 / 308.5 (47) \\
all names, 5 bins & predicted & 10.1 / 25.6 & 4.4 / 62.5 & 4.8 / 278.9 (55) & 2.1 / 283.1 (55) \\
all names, 5 bins & true & 4.5 / 511.9 (69) & 2.2 / 516.9 (69) & 7.8 / 368.3 (80) & 5.0 / 369.0 (80) \\
source names, 2 bins & predicted & 16.1 / 14.5 & 9.2 / 32.7 & 16.8 / 3.5 & 7.5 / 9.2 \\
source names, 2 bins & true & 20.3 / 10.1 & 12.7 / 21.0 & 19.5 / 3.0 & 8.5 / 7.9 \\
source names, 3 bins & predicted & 13.5 / 18.8 & 7.4 / 42.7 & 13.3 / 4.3 & 1.8 / 55.5 \\
source names, 3 bins & true & 19.2 / 11.0 & 11.0 / 25.2 & 15.9 / 3.6 & 1.1 / 85.3 \\
source names, 5 bins & predicted & 11.6 / 22.9 & 6.1 / 51.6 & 10.4 / 5.5 & 0.8 / 379.2 (88) \\
source names, 5 bins & true & 15.1 / 15.6 & 7.6 / 41.0 & 9.7 / 6.4 & 1.1 / 406.1 (96) \\
\bottomrule
\end{tabular}}}
\end{table*}
\begin{table}[tp]
  \centering
  \caption{\textbf{Prevalence-weighted label-level CRC on COCO-Stuff} $171\!\to\!1{,}283$, benchmark's own split (as in \cref{tab:weights}).}
  \label{tab:weightscoco}
  \adjustbox{max width=\linewidth}{\footnotesize\setlength{\tabcolsep}{4pt}\begin{tabular}{@{}llrr@{}}
\toprule
Prevalence bins & weights from & $\alpha=.10$ & $\alpha=.05$ \\
\midrule
unweighted (label-level CRC) & -- & 17.6 / 6.3 & 8.9 / 13.2 \\
\midrule
all names, 2 bins & predicted & 2.7 / 54.9 & 0.6 / 689.8 (37) \\
all names, 2 bins & true & 16.6 / 647.1 (3) & 8.4 / 652.0 (3) \\
all names, 3 bins & predicted & 3.9 / 34.0 & 1.2 / 484.9 (17) \\
all names, 3 bins & true & 9.6 / 860.2 (39) & 6.8 / 862.5 (39) \\
all names, 5 bins & predicted & 3.0 / 277.5 (19) & 1.1 / 365.4 (20) \\
all names, 5 bins & true & 2.2 / 1031.1 (78) & 1.5 / 1032.2 (78) \\
source names, 2 bins & predicted & 11.7 / 9.7 & 5.3 / 21.8 \\
source names, 2 bins & true & 11.9 / 9.5 & 5.3 / 21.8 \\
source names, 3 bins & predicted & 8.9 / 12.7 & 3.8 / 29.5 \\
source names, 3 bins & true & 11.4 / 9.7 & 4.5 / 25.9 \\
source names, 5 bins & predicted & 6.4 / 17.5 & 2.3 / 53.5 \\
source names, 5 bins & true & 6.6 / 17.5 & 1.9 / 71.9 \\
\bottomrule
\end{tabular}}}
\end{table}

\subsection{Added labels under image-level thresholds}
\label{sec:added}
\begin{figure*}[t]
  \centering
  \includegraphics[width=.8\textwidth]{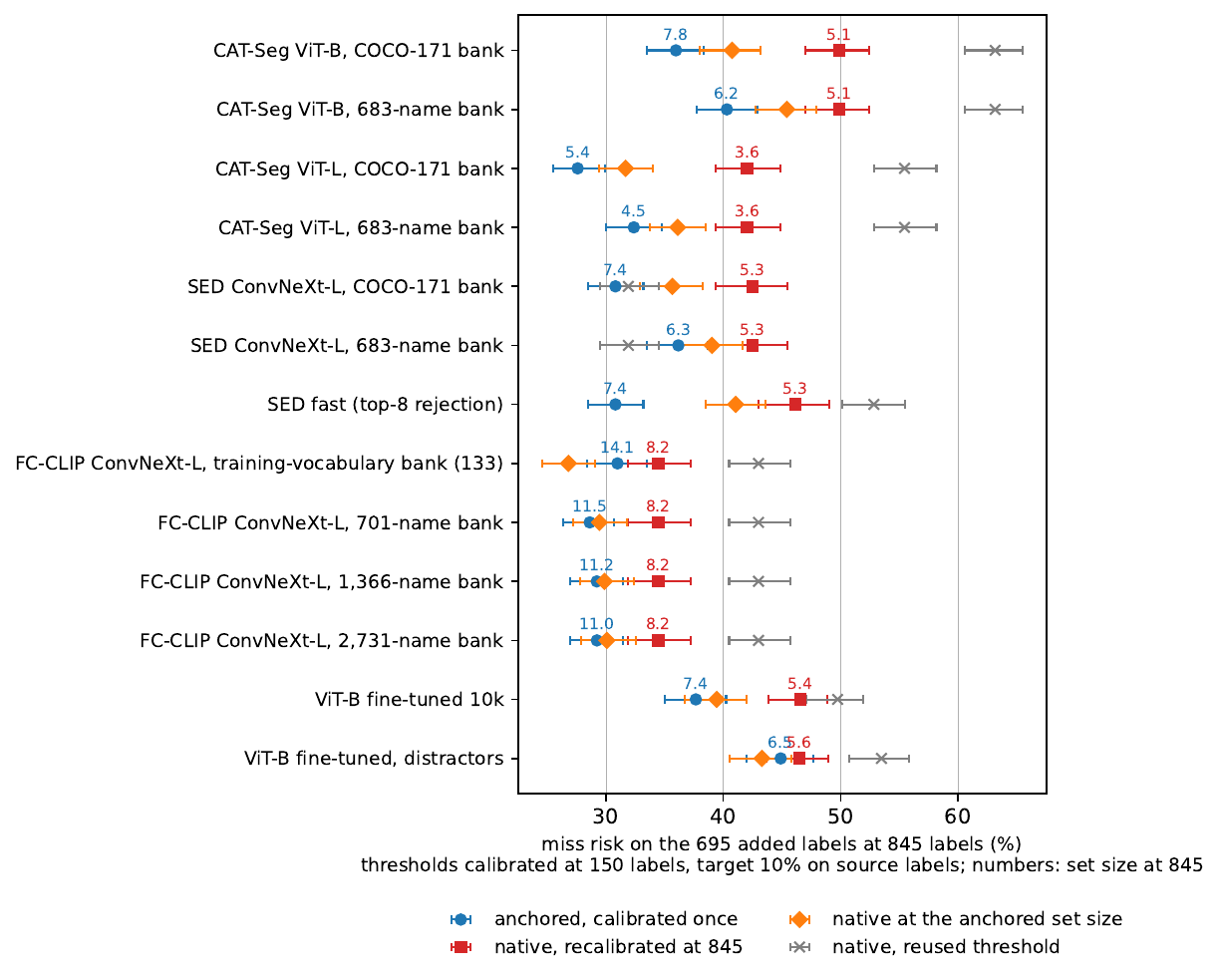}
  \caption{\textbf{Added-label risk} on ADE20K at $845$ labels, per configuration (mean $\pm$ s.d.\ over splits). Numbers: set size at $845$ labels.}
  \label{fig:added}
\end{figure*}

The image-level thresholds of \cref{sec:main} are calibrated on the source labels only and give no guarantee for the added labels. We compare their empirical miss rates at equal set size to measure how each model distributes its predictions between source and added labels. We compare the anchored threshold with the native threshold moved on the calibration images until its set size matches the anchored one (\cref{tab:added,tab:addedpascal,tab:addedcoco}, \cref{fig:added}). An oracle image-level calibration using annotations for all $845$ labels leaves an added-label risk of $\AddedOracleMin$--$\AddedOracleMax\%$ on ADE20K across the configurations of \cref{tab:added}, anchored or native, so these labels are hard for every method.

For CAT-Seg ViT-L on ADE20K, the anchored threshold misses $\VitLBankAddedAnch\%$ of the added-label pixels and the size-matched native threshold $\VitLBankAddedMatched\%$; the latter lowers the source-label risk to $\VitLBankMatchedSourceRisk\%$, \ie the native model spends the additional set size on the source labels. With the released weights, the anchored threshold misses fewer added-label pixels for every CAT-Seg and SED configuration on ADE20K and Pascal Context, and on COCO-Stuff for ViT-B, ViT-L and SED; for DeCLIP on COCO-Stuff the two are about equal ($\CocoDeclipBankAddedAnch$ against $\CocoDeclipBankAddedMatched\%$). The ViT-B aggregator fine-tuned with distractor names is an exception ($\FtDistractAddedAnch$ against $\FtDistractAddedMatched\%$; \cref{sec:finetune}). For FC-CLIP the comparison depends on the bank and the benchmark ($\FcclipBankAddedAnch$ against $\FcclipBankAddedMatched\%$ on ADE20K, $\PascalFcclipBankAddedAnch$ against $\PascalFcclipBankAddedMatched\%$ on Pascal Context, $\CocoFcclipBankAddedAnch$ against $\CocoFcclipBankAddedMatched\%$ on COCO-Stuff). Lower added-label risk at equal set size is therefore a property of CAT-Seg and SED rather than of the conversion in general.
\begin{table*}[tp]
  \centering
  \caption{\textbf{Added-label risk under image-level thresholds} on ADE20K ($\alpha=10\%$, \NumSplits\ splits, risk \% / set size at $845$ labels). Thresholds are calibrated on source labels only; the $695$ added labels are evaluated with the ADE20K-847 annotations at the same pixels ($\AddedPixelFraction\%$ of the sampled pixels). ``Native at anchored size'': the native threshold moved on the calibration images until its set size matches that of the anchored model. In the SED fast-mode row only the native arms use fast mode; the anchored arm is SED in its default mode with the COCO-171 bank.}
  \label{tab:added}
  \adjustbox{max width=\textwidth}{\footnotesize\setlength{\tabcolsep}{4pt}\begin{tabular}{@{}lrrrr@{}}
\toprule
Configuration & anchored & native, recalibrated & native at anchored size & native, reused \\
\midrule
CAT-Seg ViT-B, COCO-171 bank & 36.0 / 7.8 & 49.9 / 5.1 & 40.8 / 7.8 & 63.2 / 2.8 \\
CAT-Seg ViT-B, 683-name bank & 40.3 / 6.2 & 49.9 / 5.1 & 45.4 / 6.2 & 63.2 / 2.8 \\
CAT-Seg ViT-L, COCO-171 bank & 27.6 / 5.4 & 42.0 / 3.6 & 31.7 / 5.4 & 55.4 / 2.2 \\
CAT-Seg ViT-L, 683-name bank & 32.4 / 4.5 & 42.0 / 3.6 & 36.1 / 4.5 & 55.4 / 2.2 \\
CAT-Seg + DeCLIP EVA-L, 683-name bank & 27.1 / 4.6 & 36.9 / 3.5 & 30.0 / 4.6 & 49.6 / 2.2 \\
SED ConvNeXt-L, COCO-171 bank & 30.8 / 7.4 & 42.5 / 5.3 & 35.7 / 7.4 & 31.9 / 8.9 \\
SED ConvNeXt-L, fast mode (top-8 rejection) & 30.8 / 7.4 & 46.2 / 5.3 & 41.1 / 7.4 & 52.8 / 3.6 \\
SED ConvNeXt-L, 683-name bank & 36.2 / 6.3 & 42.5 / 5.3 & 39.0 / 6.3 & 31.9 / 8.9 \\
FC-CLIP ConvNeXt-L, 133-name bank & 31.0 / 14.1 & 34.5 / 8.2 & 26.8 / 14.1 & 43.0 / 5.2 \\
FC-CLIP ConvNeXt-L, 701-name bank & 28.6 / 11.5 & 34.5 / 8.2 & 29.4 / 11.5 & 43.0 / 5.2 \\
FC-CLIP ConvNeXt-L, 1,366-name bank & 29.2 / 11.2 & 34.5 / 8.2 & 29.9 / 11.2 & 43.0 / 5.2 \\
FC-CLIP ConvNeXt-L, 2,731-name bank & 29.2 / 11.0 & 34.5 / 8.2 & 30.1 / 11.0 & 43.0 / 5.2 \\
ViT-B, aggregator fine-tuned 10k & 37.7 / 7.4 & 46.6 / 5.4 & 39.4 / 7.4 & 49.7 / 4.7 \\
ViT-B, fine-tuned with distractors & 44.9 / 6.5 & 46.5 / 5.6 & 43.3 / 6.5 & 53.5 / 4.1 \\
\bottomrule
\end{tabular}}}
\end{table*}
\begin{table*}[tp]
  \centering
  \caption{\textbf{Added-label risk on Pascal Context} (columns as in \cref{tab:added}; risk on the $399$ added labels / set size at $458$ labels; $\PascalAddedPixelFraction\%$ of the sampled pixels carry an added label).}
  \label{tab:addedpascal}
  \adjustbox{max width=\textwidth}{\footnotesize\setlength{\tabcolsep}{4pt}\begin{tabular}{@{}lrrrr@{}}
\toprule
Configuration & anchored & native, recalibrated & native at anchored size & native, reused \\
\midrule
CAT-Seg ViT-B, COCO-171 bank, Pascal 59$\to$458 & 28.3 / 3.7 & 44.3 / 2.6 & 35.1 / 3.7 & 58.7 / 1.7 \\
CAT-Seg ViT-B, 683-name bank, Pascal 59$\to$458 & 36.5 / 3.0 & 44.3 / 2.6 & 40.5 / 3.0 & 58.7 / 1.7 \\
CAT-Seg ViT-L, 683-name bank, Pascal 59$\to$458 & 30.7 / 2.5 & 40.6 / 2.1 & 34.9 / 2.5 & 55.2 / 1.5 \\
CAT-Seg + DeCLIP EVA-L, 683-name bank, Pascal 59$\to$458 & 22.6 / 2.6 & 32.7 / 2.1 & 27.2 / 2.6 & 47.2 / 1.5 \\
SED ConvNeXt-L, 683-name bank, Pascal 59$\to$458 & 43.6 / 2.9 & 49.8 / 2.7 & 47.3 / 2.9 & 31.2 / 4.9 \\
FC-CLIP ConvNeXt-L, 701-name bank, Pascal 59$\to$458 & 36.5 / 6.1 & 37.0 / 5.4 & 34.9 / 6.1 & 49.9 / 2.9 \\
\bottomrule
\end{tabular}}}
\end{table*}
\begin{table*}[tp]
  \centering
  \caption{\textbf{Added-label risk on COCO-Stuff} (columns as in \cref{tab:added}; risk on the $1{,}112$ added labels / set size at $1{,}283$ labels; $\CocoAddedPixelFraction\%$ of the sampled pixels, in $\CocoAddedImages$ images, carry an added label).}
  \label{tab:addedcoco}
  \adjustbox{max width=\textwidth}{\footnotesize\setlength{\tabcolsep}{4pt}\begin{tabular}{@{}lrrrr@{}}
\toprule
Configuration & anchored & native, recalibrated & native at anchored size & native, reused \\
\midrule
CAT-Seg ViT-B, 683-name bank, COCO 171$\to$1,283 & 38.0 / 3.9 & 45.9 / 3.1 & 38.9 / 3.9 & 54.1 / 2.4 \\
CAT-Seg ViT-L, 683-name bank, COCO 171$\to$1,283 & 30.2 / 3.3 & 38.9 / 2.7 & 33.2 / 3.3 & 46.4 / 2.1 \\
CAT-Seg + DeCLIP EVA-L, 683-name bank, COCO 171$\to$1,283 & 25.2 / 3.2 & 30.8 / 2.6 & 24.7 / 3.2 & 38.6 / 2.1 \\
SED ConvNeXt-L, 683-name bank, COCO 171$\to$1,283 & 41.2 / 3.6 & 42.0 / 3.8 & 43.5 / 3.6 & 22.2 / 8.2 \\
FC-CLIP ConvNeXt-L, 701-name bank, COCO 171$\to$1,283 & 23.1 / 27.2 & 18.3 / 26.5 & 18.1 / 27.2 & 29.3 / 9.8 \\
\bottomrule
\end{tabular}}}
\end{table*}

\section{Reference bank and cost}
\label{sec:bankfull}
This section completes \cref{sec:bank}: the full bank sweep for CAT-Seg and the bank size for FC-CLIP (\cref{sec:banksweep}), segmentation accuracy (\cref{sec:miouproto}), runtime and memory (\cref{sec:runtime}), and fine-tuning the aggregator for anchored scoring (\cref{sec:finetune}).

\subsection{Bank sweep}
\label{sec:banksweep}
\Cref{tab:bankfull} gives the sweep summarised in \cref{sec:bank}, with intermediate sizes, banks built from Pascal Context names (evaluated on ADE20K only) and banks that combine the COCO-Stuff names with abstract WordNet \cite{miller1995wordnet} nouns or non-words (evaluated on COCO-Stuff only). Top-10 accuracy barely changes beyond $20$ names, suggesting that the bank acts mainly on the scale of the scores, not on their ranking. For FC-CLIP (\cref{tab:fcclip}) the overhead decreases with the bank size up to about $1{,}400$ names and then levels off.
\paragraph{Images that chose the bank.} The sweep ran on the first $\DisjBankImages$ of the $2{,}000$ ADE20K validation images used in \cref{tab:main}. Repeating the half/half splits on the other \DisjImages\ images alone, with the $683$-name bank, gives an anchored risk of at most $\DisjAnchRiskHi\%$ at $\alpha=10\%$ for CAT-Seg ViT-L, SED and FC-CLIP, with set sizes within $\DisjSizeGapMax$ labels per pixel of those in \cref{tab:main}.
\begin{table*}[tp]
  \centering
  \caption{\textbf{Reference-bank sweep} (anchored CAT-Seg, $\alpha=10\%$): set size, risk (\%) and top-10 accuracy (\%). ADE20K $150\!\to\!845$ on a \BankSweepImages-image subset (sizes differ from \cref{tab:main}). COCO-Stuff: $171$ classes plus $695$ Open Images (OI) names that are in no bank (\BankCocoImages\ images). Abstract nouns: WordNet nouns with only abstract senses; non-words: pronounceable non-words; --: not run.}
  \label{tab:bankfull}
  \adjustbox{max width=\textwidth}{\footnotesize\setlength{\tabcolsep}{4pt}\begin{tabular}{@{}lrrrrrrr@{}}
\toprule
 & & \multicolumn{3}{c}{ADE20K, ViT-B} & \multicolumn{3}{c}{COCO-Stuff} \\
\cmidrule(lr){3-5}\cmidrule(l){6-8}
Reference bank & names & size & risk & top-10 & ViT-B size & ViT-B risk & ViT-L size \\
\midrule
native, recalibrated & -- & -- & -- & -- & 2.87 & 9.7 & 2.33 \\
\midrule
none (candidate only) & 0 & 82.34 & 9.4 & 49.3 & 108.50 & 9.7 & 139.25 \\
COCO-20 & 20 & 33.68 & 9.5 & 88.4 & 32.74 & 9.6 & 28.05 \\
COCO-85 & 85 & 11.18 & 9.5 & 90.9 & 6.18 & 9.6 & 4.81 \\
COCO-171 & 171 & 7.92 & 9.5 & 90.8 & 3.15 & 9.7 & 2.86 \\
Pascal-171 & 171 & 8.85 & 9.5 & 91.0 & -- & -- & -- \\
COCO + OI 255 & 255 & 7.65 & 9.5 & 90.9 & 2.99 & 9.7 & 2.71 \\
COCO + Pascal 255 & 255 & 7.39 & 9.6 & 90.9 & -- & -- & -- \\
COCO + OI 342 & 342 & 7.07 & 9.6 & 91.1 & 3.04 & 9.7 & 2.60 \\
COCO + Pascal 342 & 342 & 6.91 & 9.6 & 91.2 & -- & -- & -- \\
COCO + OI 683 & 683 & 6.44 & 9.5 & 91.5 & 3.32 & 9.7 & 2.81 \\
OI 683, no COCO names & 683 & 8.78 & 9.5 & 91.0 & 4.58 & 9.7 & 3.95 \\
COCO + OI 1,366 & 1,366 & 6.87 & 9.5 & 91.5 & 3.72 & 9.7 & 3.19 \\
COCO + OI 2,731 & 2,731 & 7.20 & 9.5 & 91.3 & 3.89 & 9.6 & 3.39 \\
COCO + OI 5,462 & 5,462 & 7.62 & 9.5 & 91.2 & 4.10 & 9.6 & 3.57 \\
\midrule
abstract nouns 683 & 683 & -- & -- & -- & 4.56 & 9.7 & -- \\
non-words 683 & 683 & -- & -- & -- & 5.06 & 9.7 & -- \\
COCO + abstract 683 & 683 & -- & -- & -- & 3.92 & 9.7 & -- \\
COCO + non-words 683 & 683 & -- & -- & -- & 4.01 & 9.7 & -- \\
\bottomrule
\end{tabular}}}
\end{table*}
\begin{table}[tp]
  \centering
  \caption{\textbf{FC-CLIP: bank size and overhead} (ADE20K, $845$ labels, $\alpha=10\%$): set sizes of the anchored and the recalibrated native model. ``Added risk'': risk on the $695$ added labels for the anchored threshold / for the native threshold moved to the anchored set size (\cref{sec:added}).}
  \label{tab:fcclip}
  \adjustbox{max width=\linewidth}{\footnotesize\setlength{\tabcolsep}{4pt}\begin{tabular}{@{}rrrrr@{}}
\toprule
bank names & anchored & native recal. & overhead & added risk \\
\midrule
133 & 13.56 & 8.18 & 5.37 & 31.0 / 26.8 \\
701 & 11.18 & 8.18 & 2.99 & 28.6 / 29.4 \\
1,366 & 10.83 & 8.18 & 2.65 & 29.2 / 29.9 \\
2,731 & 10.71 & 8.18 & 2.53 & 29.2 / 30.1 \\
\bottomrule
\end{tabular}
}}
\end{table}
\paragraph{Redrawing the Open Images names.} The $683$-name bank holds the COCO-171 names and $512$ names drawn at random from Open Images. We redrew the $512$ names five times, keeping the COCO-171 names, and scored the \BankSweepImages\ sweep images with each bank for SED and CAT-Seg ViT-L (\cref{tab:bankseeds}); a redraw shares $\BankSeedsSharedLo$--$\BankSeedsSharedHi$ names with the draw used in the paper. Every draw meets the target at both levels. At $\alpha=10\%$, the standard deviation of the set size over the redraws is $\BankSeedsSizeSdSed$ labels per pixel for SED and $\BankSeedsSizeSdVitL$ for ViT-L, less than its standard deviation over the calibration splits of one bank ($\BankSeedsSplitSdSed$ and $\BankSeedsSplitSdVitL$). The overhead stays between $\BankSeedsOverheadLo$ and $\BankSeedsOverheadHi$ labels, and top-1 accuracy at $845$ names changes by at most $\BankSeedsTopOneMaxChange$ points.
\begin{table}[tp]
  \centering
  \caption{\textbf{Redraws of the bank's Open Images names} (ADE20K $150\!\to\!845$, the \BankSweepImages\ images of \cref{tab:bankfull}, $\BankSeedsSplits$ half/half splits): risk (\%) and set size at $\alpha=10\%$, and set size at $\alpha=5\%$. Every bank holds the COCO-171 names and $512$ Open Images names.}
  \label{tab:bankseeds}
  \adjustbox{max width=\linewidth}{\footnotesize\setlength{\tabcolsep}{4pt}\begin{tabular}{@{}lrrrrrr@{}}
\toprule
& \multicolumn{3}{c}{SED} & \multicolumn{3}{c}{CAT-Seg ViT-L} \\
\cmidrule(lr){2-4}\cmidrule(l){5-7}
Open Images names & risk & size & size, $5\%$ & risk & size & size, $5\%$ \\
\midrule
native, recalibrated & 9.51 & 5.65 & 17.44 & 9.41 & 3.80 & 7.86 \\
\midrule
draw of the paper & 9.46 & 6.50 & 16.42 & 9.48 & 4.48 & 8.72 \\
redraw 1 & 9.49 & 6.23 & 15.98 & 9.53 & 4.34 & 8.65 \\
redraw 2 & 9.49 & 6.44 & 16.02 & 9.50 & 4.52 & 8.62 \\
redraw 3 & 9.44 & 6.42 & 16.75 & 9.46 & 4.62 & 8.94 \\
redraw 4 & 9.46 & 6.42 & 16.31 & 9.50 & 4.38 & 8.89 \\
redraw 5 & 9.46 & 6.50 & 16.27 & 9.51 & 4.44 & 9.03 \\
\midrule
mean of the redraws & 9.47 & 6.40 & 16.27 & 9.50 & 4.46 & 8.82 \\
s.d.\ of the redraws & 0.02 & 0.11 & 0.31 & 0.03 & 0.11 & 0.18 \\
\bottomrule
\end{tabular}}}
\end{table}

\subsection{Segmentation accuracy}
\label{sec:miouproto}
\Cref{tab:miou,tab:pixacc} give the accuracy summarised in \cref{sec:bank}. CAT-Seg runs the released sliding-window inference ($640$ square, kernel $384$, overlap $0.333$) with one global view, obtained by resizing the $640$ square to $384$ (the release resizes the input image instead). SED runs the released whole-image inference at $768\times768$. FC-CLIP runs the released ConvNeXt-L COCO-panoptic weights with its A-847 configuration (shortest edge $800$, $250$ queries) on all four benchmarks. Class names are those shipped with each model; for CAT-Seg we average the synonym embeddings, where the release uses the first synonym (on PC-59 this changes CAT-Seg ViT-B by at most $\MiouProtocolMaxChange$ points). Scores are resampled to the annotation grid, the arg-max is taken over the benchmark vocabulary, and mIoU follows the detectron2 convention; PC-459 keeps ``unknown'' as a class, as in the released preparation. All validation images are used ($2{,}000$ for ADE20K, $5{,}104$ for Pascal Context).

The native rows are our runs of the released code. They are within $0.4$ points of the published values on A-847, PC-459 and A-150 and $1.0$--$1.8$ points below on PC-59 for CAT-Seg, SED and DeCLIP; FC-CLIP is $1.3$--$3.4$ points below its published values on all four benchmarks, including A-847, whose released configuration we use throughout; we did not identify the cause. The anchored--native gaps use these same runs as the baseline; the FC-CLIP shortfall limits comparison with its published performance. For FC-CLIP the conversion raises pixel accuracy on every benchmark while lowering mIoU on ADE20K (\cref{tab:pixacc}): the softmax over the bank favours frequent classes.

\paragraph{Source and added classes.}
\Cref{tab:pixaccadded} splits the pixel accuracy on A-847 and PC-459 into the classes of A-150 or PC-59 (source) and the others (added: $\SegAddClassesAdeEightFortySeven$ classes and $\SegAddShareAdeEightFortySeven\%$ of the annotated pixels on A-847; $\SegAddClassesPcFourFiftyNine$ classes, $\SegAddPresentPcFourFiftyNine$ of them present, and $\SegAddSharePcFourFiftyNine\%$ of the pixels on PC-459, where ``unknown'' belongs to neither group). The prediction is still the arg-max over the whole benchmark vocabulary. For CAT-Seg and SED with the $683$-name bank, added-class pixel accuracy changes by at most $\SegAddPixMaxAbsChange$ points, and with the COCO-171 bank it rises by $\SegAddPixCocoGainLo$--$\SegAddPixCocoGainHi$ points; the small loss in overall pixel accuracy falls on the source classes. In these models the bank enters only the class attention, and each label keeps its own logit. For FC-CLIP with the $701$-name bank, added-class pixel accuracy falls from $\SegAddFcclipNativeAdeEightFortySevenAdd$ to $\SegAddFcclipAnchoredBankAdeEightFortySevenAdd\%$ on A-847 and from $\SegAddFcclipNativePcFourFiftyNineAdd$ to $\SegAddFcclipAnchoredBankPcFourFiftyNineAdd\%$ on PC-459, while source-class accuracy rises. Here the bank is the normaliser of each mask's class probabilities: an added candidate competes with every bank name, but no bank name competes with it. Grounding DINO loses added-class accuracy in the same way, with the bank fixing its boxes (\cref{tab:ovdacc}). Anchoring thus costs added-class accuracy where the bank fixes the output on which a candidate is scored, not where it only provides context.
\begin{table*}[tp]
  \centering
  \caption{\textbf{Full-resolution mIoU} of the released models and their anchored conversions, with the official inference of each model. Gaps quoted in the text are computed before rounding.}
  \label{tab:miou}
  \adjustbox{max width=\textwidth}{\footnotesize\setlength{\tabcolsep}{4pt}\begin{tabular}{@{}llrrrr@{}}
\toprule
Model & scoring & A-150 & PC-459 & A-847 & PC-59 \\
\midrule
CAT-Seg ViT-B & native (released) & 31.7 & 19.0 & 12.2 & 56.4 \\
 & anchored, COCO-171 bank & 30.4 & 18.4 & 11.6 & 54.3 \\
 & anchored, 683-name bank & 31.0 & 18.7 & 11.9 & 55.0 \\
\midrule
CAT-Seg ViT-L & native (released) & 37.8 & 23.7 & 15.9 & 62.0 \\
 & anchored, COCO-171 bank & 36.1 & 23.1 & 15.4 & 60.1 \\
 & anchored, 683-name bank & 36.8 & 23.2 & 15.2 & 60.4 \\
\midrule
CAT-Seg + DeCLIP EVA-L & native (released) & 40.6 & 25.8 & 17.9 & 62.8 \\
 & anchored, COCO-171 bank & 38.5 & 25.1 & 16.9 & 60.1 \\
 & anchored, 683-name bank & 39.7 & 25.2 & 16.9 & 61.1 \\
\midrule
SED ConvNeXt-L & native (released) & 34.9 & 22.2 & 13.7 & 59.1 \\
 & anchored, COCO-171 bank & 33.5 & 22.1 & 13.6 & 57.5 \\
 & anchored, 683-name bank & 34.0 & 21.9 & 13.7 & 57.8 \\
\midrule
FC-CLIP ConvNeXt-L & native (released) & 32.0 & 16.6 & 13.5 & 55.0 \\
 & anchored, 133-name bank & 24.3 & 15.1 & 7.0 & 57.2 \\
 & anchored, 701-name bank & 28.0 & 16.8 & 10.7 & 55.7 \\
\bottomrule
\end{tabular}}}
\end{table*}
\begin{table*}[tp]
  \centering
  \caption{\textbf{Pixel accuracy (\%)} of the runs in \cref{tab:miou}.}
  \label{tab:pixacc}
  \adjustbox{max width=\textwidth}{\footnotesize\setlength{\tabcolsep}{4pt}\begin{tabular}{@{}llrrrr@{}}
\toprule
Model & scoring & A-150 & PC-459 & A-847 & PC-59 \\
\midrule
CAT-Seg ViT-B & native (released) & 72.9 & 69.4 & 62.1 & 79.4 \\
 & anchored, COCO-171 bank & 70.4 & 67.6 & 59.0 & 77.8 \\
 & anchored, 683-name bank & 72.1 & 68.8 & 60.8 & 78.6 \\
\midrule
CAT-Seg ViT-L & native (released) & 76.1 & 72.1 & 64.3 & 81.8 \\
 & anchored, COCO-171 bank & 73.4 & 70.6 & 60.9 & 80.4 \\
 & anchored, 683-name bank & 75.3 & 71.4 & 63.0 & 80.9 \\
\midrule
CAT-Seg + DeCLIP EVA-L & native (released) & 77.0 & 72.2 & 66.3 & 81.9 \\
 & anchored, COCO-171 bank & 74.4 & 70.6 & 63.3 & 80.4 \\
 & anchored, 683-name bank & 76.3 & 71.6 & 65.4 & 81.1 \\
\midrule
SED ConvNeXt-L & native (released) & 73.8 & 70.0 & 61.7 & 80.1 \\
 & anchored, COCO-171 bank & 71.6 & 69.1 & 59.1 & 79.0 \\
 & anchored, 683-name bank & 73.1 & 69.7 & 60.8 & 79.3 \\
\midrule
FC-CLIP ConvNeXt-L & native (released) & 67.7 & 61.5 & 51.6 & 76.0 \\
 & anchored, 133-name bank & 70.7 & 68.8 & 62.6 & 78.4 \\
 & anchored, 701-name bank & 69.9 & 65.5 & 59.0 & 76.7 \\
\bottomrule
\end{tabular}}}
\end{table*}

\begin{table*}[tp]
  \centering
  \caption{\textbf{Pixel accuracy (\%) on source and added classes} of the A-847 and PC-459 runs in \cref{tab:miou}, from their confusion matrices. Source: the classes of A-150 or PC-59; added: the other classes (``unknown'' belongs to neither). Source and added: pixel accuracy over the pixels of each group; class mean and mIoU: over the added classes with ground-truth pixels. The arg-max is over the whole benchmark vocabulary.}
  \label{tab:pixaccadded}
  \adjustbox{max width=\textwidth}{\footnotesize\setlength{\tabcolsep}{4pt}\setlength{\tabcolsep}{4pt}\begin{tabular}{@{}llrrrrrrrr@{}}
\toprule
& & \multicolumn{4}{c}{A-847 ($150$ source, $697$ added)} & \multicolumn{4}{c}{PC-459 ($59$ source, $399$ added)} \\ \cmidrule(lr){3-6}\cmidrule(l){7-10}
Model & scoring & source & added & class mean & mIoU & source & added & class mean & mIoU \\
\midrule
CAT-Seg ViT-B & native (released) & 64.8 & 25.7 & 19.2 & 8.9 & 73.7 & 35.4 & 29.0 & 12.6 \\
 & anchored, COCO-171 bank & 61.4 & 27.3 & 21.1 & 8.4 & 71.7 & 37.7 & 33.8 & 12.2 \\
 & anchored, 683-name bank & 63.4 & 25.9 & 18.7 & 8.7 & 73.1 & 36.4 & 29.3 & 12.4 \\
\midrule
CAT-Seg ViT-L & native (released) & 66.6 & 32.9 & 25.8 & 12.7 & 76.3 & 40.2 & 37.3 & 17.2 \\
 & anchored, COCO-171 bank & 62.8 & 35.2 & 29.7 & 12.3 & 74.5 & 42.2 & 42.7 & 16.7 \\
 & anchored, 683-name bank & 65.3 & 31.6 & 24.9 & 12.0 & 75.6 & 40.4 & 37.5 & 16.7 \\
\midrule
CAT-Seg + DeCLIP EVA-L & native (released) & 68.6 & 35.6 & 27.9 & 14.2 & 76.3 & 43.5 & 43.1 & 19.6 \\
 & anchored, COCO-171 bank & 65.2 & 37.3 & 30.2 & 13.2 & 74.5 & 45.3 & 47.5 & 19.0 \\
 & anchored, 683-name bank & 67.7 & 34.1 & 26.7 & 13.2 & 75.6 & 44.5 & 43.1 & 19.0 \\
\midrule
SED ConvNeXt-L & native (released) & 64.0 & 31.4 & 24.8 & 10.7 & 74.3 & 37.0 & 33.8 & 15.9 \\
 & anchored, COCO-171 bank & 60.9 & 34.9 & 28.7 & 10.8 & 73.2 & 38.3 & 38.9 & 15.8 \\
 & anchored, 683-name bank & 62.9 & 32.5 & 26.0 & 10.8 & 74.0 & 36.9 & 34.0 & 15.5 \\
\midrule
FC-CLIP ConvNeXt-L & native (released) & 52.7 & 37.5 & 29.0 & 10.3 & 65.2 & 35.5 & 36.7 & 10.1 \\
 & anchored, 133-name bank & 66.4 & 11.6 & 8.3 & 4.0 & 73.6 & 26.6 & 18.9 & 7.5 \\
 & anchored, 701-name bank & 61.7 & 23.2 & 17.0 & 7.7 & 69.8 & 31.4 & 31.5 & 9.8 \\
\bottomrule
\end{tabular}}}
\end{table*}

\subsection{Runtime and memory}
\label{sec:runtime}
\Cref{tab:runtime} gives the cost quoted in \cref{sec:conversion}. The native model is run once at the grown vocabulary, as at deployment; the pass at the source vocabulary is needed only once, for calibration. The segmentation block also includes a separate native pass over the bank and each candidate, to measure the cost of recomputing their joint context. The detection rows compare the native model run on all $1{,}204$ names in chunks of $40$ with the anchored model (\cref{sec:detectionfull}); here the ratio is below one, because the anchored model runs the image side twice, not once per chunk.
\begin{table*}[tp]
  \centering
  \caption{\textbf{Runtime and memory} (ratios from unrounded times): seconds per image and peak memory, fp32, on one GH200. Segmentation, $845$ labels: ``anchored'' is the bank pass plus all $845$ candidates, and ratio = (features + anchored) / (features + native); the rows were timed consecutively on an idle device over the same $100$ development images, except the one-label-at-a-time row (local RTX 5070 Ti). ViT-B and ViT-L share the aggregator, so only the CLIP features differ. Detection, $1{,}204$ names: ratio = anchored / native, timed consecutively on an idle device over the first $100$ benchmark images. Peak memory is the maximum over all passes of a run; \cref{tab:memory} gives each arm separately.}
  \label{tab:runtime}
  \adjustbox{max width=\textwidth}{\footnotesize\setlength{\tabcolsep}{4pt}\begin{tabular}{@{}lrrrrr@{}}
\toprule
Model & features & native & anchored & ratio & peak memory \\
\midrule
CAT-Seg ViT-B, COCO-171 bank, sliding window 640 & 0.03 & 0.27 & 0.88 & 3.0$\times$ & 8.3 GB \\
CAT-Seg ViT-B, 683-name bank & 0.03 & 0.27 & 1.29 & 4.3$\times$ & 8.4 GB \\
CAT-Seg ViT-L, COCO-171 bank & 0.07 & 0.28 & 0.88 & 2.8$\times$ & 9.3 GB \\
CAT-Seg ViT-L, 683-name bank & 0.07 & 0.28 & 1.29 & 4.0$\times$ & 9.4 GB \\
SED ConvNeXt-L, COCO-171 bank, whole image 768 & 0.03 & 0.34 & 0.40 & 1.2$\times$ & 8.0 GB \\
SED ConvNeXt-L, fast mode (top-8), native only & 0.03 & 0.07 & -- & -- & 5.5 GB \\
SED ConvNeXt-L, 683-name bank & 0.03 & 0.34 & 0.60 & 1.7$\times$ & 8.2 GB \\
FC-CLIP ConvNeXt-L, 133-name bank, shortest edge 800 & 0.31$^\dagger$ & 0.0003 & 0.0018 & -- & 3.7 GB \\
FC-CLIP ConvNeXt-L, 701-name bank & 0.40$^\dagger$ & 0.0003 & 0.0076 & $\geq$1.3$\times$ & 4.8 GB \\
native, one label at a time against the bank (ViT-B, RTX 5070 Ti) & -- & 350 & -- & -- & -- \\
\midrule
\multicolumn{6}{l}{\emph{Detection, $1{,}204$ names}} \\
Grounding DINO T, 80-name bank & -- & 2.97 & 0.50 & 0.17$\times$ (6.0$\times$ faster) & 9.3 GB \\
MM-Grounding DINO T, 80-name bank & -- & 3.04 & 0.51 & 0.17$\times$ (6.0$\times$ faster) & 9.3 GB \\
Grounding DINO B, 80-name bank & -- & 3.31 & 0.52 & 0.16$\times$ (6.3$\times$ faster) & 9.5 GB \\
OWLv2 B (control) & -- & 0.09 & -- & -- & 1.4 GB \\
\bottomrule
\multicolumn{6}{l}{\footnotesize $^\dagger$backbone, mask proposals and per-query logits, including the bank columns, which only the anchored arm needs.}\\
\multicolumn{6}{l}{\footnotesize No FC-CLIP pass without bank columns was timed; the $133$-name-bank pass bounds it from above, so the $701$-name-bank ratio is a lower bound.}
\end{tabular}}}
\end{table*}
\paragraph{A leaner bank pass.} The bank pass timed in \cref{tab:runtime} also computes outputs that no candidate reads: for CAT-Seg the bank's own logits and the output of the last attention layer, for SED the stage-3 attention output, the third fusion block and the prediction head. Stopping the pass once the key and value sums of every layer are recorded leaves each candidate score unchanged bit for bit (checked on $\LeanCheckImages$ images per model), so the calibration carries over as it is. \Cref{tab:runtimelean} compares the two passes on an RTX 5070 Ti (\cref{tab:runtime} was timed on a GH200). With the $683$-name bank, the lean pass removes $\RtLeanVitLBankSaving\%$ of the anchored time for CAT-Seg ViT-L and $\RtLeanSedBankSaving\%$ for SED; the ratio falls from $\RtLocalVitLBankRatio$ to $\RtLeanVitLBankRatio\times$ for ViT-L and from $\RtLocalSedBankRatio$ to $\RtLeanSedBankRatio\times$ for SED. CAT-Seg with the DeCLIP backbone, which has the ViT-L aggregator, runs at $\RtLocalDeclipBankRatio$ and $\RtLeanDeclipBankRatio\times$ the native time. Much of the remaining work comes from the number of labels processed: the anchored pass scores the bank names and all $845$ candidates, where native CAT-Seg scores $256$ per crop.
\begin{table*}[tp]
  \centering
  \caption{\textbf{Full against lean bank pass} (ratios from unrounded times): seconds per image, fp32, on one RTX 5070 Ti over the same $100$ development images as \cref{tab:runtime}, each pair timed back to back on an idle device. The lean pass stops once the key/value sums are recorded; candidate scores are identical. Features and the native pass are shared by both runs; peak memory is the maximum over all passes of the lean run.}
  \label{tab:runtimelean}
  \adjustbox{max width=\textwidth}{\footnotesize\setlength{\tabcolsep}{4pt}\begin{tabular}{@{}lrrrrrrr@{}}
\toprule
& & & \multicolumn{2}{c}{full bank pass} & \multicolumn{2}{c}{lean bank pass} & \\
\cmidrule(lr){4-5}\cmidrule(lr){6-7}
Model & features & native & anchored & ratio & anchored & ratio & peak memory \\
\midrule
CAT-Seg ViT-B, COCO-171 bank, sliding window 640 & 0.07 & 0.75 & 2.69 & 3.3$\times$ & 2.56 & 3.2$\times$ & 8.2 GB \\
CAT-Seg ViT-B, 683-name bank & 0.07 & 0.77 & 3.98 & 4.8$\times$ & 3.38 & 4.1$\times$ & 8.2 GB \\
CAT-Seg ViT-L, COCO-171 bank & 0.22 & 0.75 & 2.70 & 3.0$\times$ & 2.55 & 2.8$\times$ & 9.3 GB \\
CAT-Seg ViT-L, 683-name bank & 0.23 & 0.75 & 3.98 & 4.3$\times$ & 3.38 & 3.7$\times$ & 9.3 GB \\
SED ConvNeXt-L, COCO-171 bank, whole image 768 & 0.10 & 0.64 & 0.75 & 1.2$\times$ & 0.71 & 1.1$\times$ & 8.0 GB \\
SED ConvNeXt-L, 683-name bank & 0.15 & 0.63 & 1.13 & 1.6$\times$ & 0.95 & 1.4$\times$ & 8.1 GB \\
CAT-Seg + DeCLIP EVA-L, 683-name bank & 0.24 & 0.77 & 4.02 & 4.2$\times$ & 3.39 & 3.6$\times$ & 9.3 GB \\
\bottomrule
\end{tabular}}}
\end{table*}

\paragraph{Memory.} \Cref{tab:memory} gives the peak GPU memory of each arm on its own, as the vocabulary grows from the $150$ source names to $6{,}760$ (the $845$ evaluated names, then Open Images names). For CAT-Seg ViT-L, the anchored model with the lean bank pass needs $\MemVitLSrcExtra$\,GB more than the native model at the source vocabulary and $\MemVitLFullSaving$\,GB less from $845$ names on. Both stay flat beyond $845$ names, the native model because it keeps $256$ labels per crop. Native SED grows with every name: $\MemSedNatFull$\,GB at $845$ names, $\MemSedNatFour$\,GB at $3{,}380$, more than the card holds, and out of memory at $6{,}760$. Each of its stages attends over all labels, so the feature maps of every label are kept until the stage is complete. An anchored candidate needs only the bank's key and value sums, so candidates are processed in chunks, and the anchored peak stays between $\MemSedAnchMin$ and $\MemSedAnchMax$\,GB. FC-CLIP classifies a fixed set of mask queries, and both of its arms stay below $\MemFcMax$\,GB. CAT-Seg with the DeCLIP backbone follows ViT-L: the anchored model needs $\MemDeclipAnchMax$\,GB at every size, the native model $\MemDeclipNatFull$\,GB from $845$ names on. Besides the weights, the anchored model keeps the bank's text embeddings ($\MemBankVitLMB$\,MB for CAT-Seg ViT-L, $\MemBankSedMB$\,MB for SED). A recalibration needs the labelled calibration images ($\MemCalMB$\,MB for $\MemCalImages$ ADE20K images) and the memory of a native pass at the new vocabulary.
\begin{table}[tp]
  \centering
  \caption{\textbf{Peak GPU memory per arm} (GB), fp32, on one RTX 5070 Ti, maximum over $\MemImages$ ADE20K development images, with the peak reset before each arm. Weights, text embeddings and image features are included. Vocabularies: the $845$ evaluated names, then Open Images names. \emph{lean}: lean bank pass. Native CAT-Seg matches the released code to within $0.02$\,GB. $^\ddagger$More than the $\MemCard$\,GB of the card, allocated only through the driver's fallback to system memory. \emph{OOM}: out of memory.}
  \label{tab:memory}
  \adjustbox{max width=\linewidth}{\footnotesize\setlength{\tabcolsep}{4pt}\begin{tabular}{@{}llrrrrr@{}}
\toprule
Model & arm & \multicolumn{5}{c}{vocabulary size} \\
\cmidrule(l){3-7}
& & $150$ & $845$ & $1{,}690$ & $3{,}380$ & $6{,}760$ \\
\midrule
CAT-Seg ViT-B & native & 4.99 & 8.05 & 8.05 & 8.05 & 8.06 \\
 & anchored & 7.44 & 7.44 & 7.44 & 7.44 & 7.44 \\
 & anchored, lean & 7.39 & 7.39 & 7.39 & 7.39 & 7.39 \\
\addlinespace
CAT-Seg ViT-L & native & 6.04 & 9.10 & 9.10 & 9.11 & 9.12 \\
 & anchored & 8.49 & 8.49 & 8.49 & 8.49 & 8.49 \\
 & anchored, lean & 8.44 & 8.44 & 8.44 & 8.44 & 8.44 \\
\addlinespace
SED & native & 3.55 & 8.01 & 14.47 & 27.39$^\ddagger$ & \emph{OOM} \\
 & anchored & 6.62 & 6.62 & 6.62 & 6.62 & 6.62 \\
 & anchored, lean & 4.65 & 4.65 & 4.65 & 4.68 & 5.91 \\
\addlinespace
FC-CLIP & native & 2.59 & 2.60 & 2.61 & 2.63 & 2.67 \\
 & anchored & 2.60 & 2.61 & 2.62 & 2.64 & 2.67 \\
\addlinespace
CAT-Seg + DeCLIP & native & 6.04 & 9.10 & 9.10 & 9.11 & 9.12 \\
 & anchored & 8.49 & 8.49 & 8.49 & 8.49 & 8.49 \\
 & anchored, lean & 8.44 & 8.44 & 8.44 & 8.44 & 8.44 \\
\bottomrule
\end{tabular}}}
\end{table}

\subsection{When recalibration suffices}
\label{sec:recalcost}
Recalibrating the native model at a new vocabulary needs no new annotation, since the calibration images are labelled for the source vocabulary and the loss \eqref{eq:loss} counts only source-labelled pixels. It requires access to those images and a known vocabulary or selection rule. A fixed per-image rule can be replayed on the calibration images \cite{jin2025focal}; a fixed family of rules can be calibrated under its worst case (\cref{prop:adaptive}). Each vocabulary or rule requires a native pass over the calibration images. 

With $\RtCalImages$ calibration images, this pass takes $\RtVitLBankRecalMin$ minutes on one GH200 for CAT-Seg ViT-L, $\RtSedBankRecalMin$ for SED and at most $\RtFcclipBankRecalMinMax$ for FC-CLIP (\cref{tab:runtime}). The conversion instead adds (ratio${}-1$) native passes to every test image. Anchored inference therefore costs less total compute whenever fewer than $\RtCalImages/(\text{ratio}-1)$ test images are served per vocabulary: $\RtVitLBankBreakEven$ for CAT-Seg ViT-L, $\RtVitBBankBreakEven$ for ViT-B and $\RtSedBankBreakEven$ for SED, with the $683$-name bank. If the vocabulary-independent features of the calibration images are cached, the break-even falls to $\RtVitLBankBreakEvenCached$, $\RtVitBBankBreakEvenCached$ and $\RtSedBankBreakEvenCached$ images. 

For FC-CLIP the conversion adds the per-query logits of the $701$ bank names to every test image. No native pass without these columns was timed, but the pass with the $133$-name bank bounds it from above, so the ratio is at least $\RtFcclipBankRatioMin$ and the break-even is at most $\RtFcclipBankBreakEvenMax$ images. FC-CLIP's backbone and mask proposals do not depend on the vocabulary, so with these cached, recalibration reduces to the logits of the new names for the cached queries. This step was not timed separately. \Cref{tab:regimes} summarises which deployments each option serves.
\begin{table*}[tp]
  \centering
  \caption{\textbf{Calibrating once or recalibrating}, by deployment. All guarantees in the table assume exchangeable calibration and test images; per-image vocabularies are evaluated on ADE20K and Pascal Context (\cref{tab:adaptivemain}).}
  \label{tab:regimes}
  \footnotesize
  \begin{tabular}{@{}p{0.27\textwidth}p{0.47\textwidth}p{0.2\textwidth}@{}}
    \toprule
    Deployment & Native threshold & Anchored, calibrated once \\
    \midrule
    One vocabulary, fixed before deployment & Recalibrate at it: valid, usually smaller sets (\cref{tab:main}); cheaper above the break-even & Valid; cheaper below the break-even \\
    Vocabulary fixed per project before its images are seen & Recalibrate for each project: valid, one pass over the calibration images each time & Valid, no pass \\
    Vocabularies chosen per image by a rule replayable on the calibration images, \eg a fixed image tagger & Calibrate under the rule, or the worst case over a fixed set of rules: valid, smaller sets; repeated whenever a rule changes & Valid \\
    Names from a known lexicon, rule not known & Recalibrate at the lexicon: meets the target for CAT-Seg and FC-CLIP, with sets about twice the anchored size at a large lexicon, not for SED, which needs more memory than a $16$\,GB card from $3{,}380$ names (\cref{tab:memory}). Calibrate on random vocabularies from the lexicon: exceeds the target under the tagger rule for all three models & Valid \\
    Names listed per image by a language model, not limited to a list (\cref{tab:freeform}) & Calibrate under the model's names on the calibration images: valid, larger sets than anchored; repeated whenever the model or prompt changes. Recalibrating at the $845$ names exceeds the target for all three models; a $17{,}001$-name lexicon meets it for FC-CLIP at $\FfFcLexRatio\times$ the anchored size, not for SED & Valid \\
    Names written per image or per query by a user, or added during annotation & Not replayable; guaranteed if every vocabulary the process can produce is scored on every calibration image (\cref{prop:adaptive}) & Valid, with $\Vsrc$ appended to the names \\
    Vocabulary grown by a party without the labelled calibration images & Not possible & Valid \\
    \bottomrule
  \end{tabular}
\end{table*}

\subsection{Recalibration in deployment}
\label{sec:recaldeploy}
\Cref{tab:regimes} lists when recalibration is available. This section describes what it requires when the vocabulary changes during use.

\paragraph{Detecting the failure.} Recalibration starts only once the failure is noticed. Without labels at deployment, the prediction sets are the only signal, and in the \SilentNUnder\ of \SilentN\ combinations where the reused threshold exceeds the target, they grow by only $\SilentShareLo$--$\SilentShareHi\%$ of what the target needs (\cref{sec:failure}). The valid anchored sets grow more than the failing native ones in each of these combinations (\cref{sec:main}), so set growth does not separate a valid threshold from a failing one. Detecting the rise in risk requires labelled images at the grown vocabulary.

\paragraph{Until recalibration ends.} Recalibration takes one native pass over the $\RtCalImages$ calibration images, as long as segmenting them: $\RtVitLBankRecalMin$ minutes for CAT-Seg ViT-L on one GH200 and $\RtLocalVitLBankRecalMin$ on one RTX 5070 Ti (\cref{sec:recalcost}). On a single device the system either stops for this time or keeps running under the invalid threshold, and recalibrating afterwards does not cover the images already processed. When a tagger chooses the names for each frame, the vocabulary changes before any recalibration ends.

\paragraph{Data on the device.} Each recalibration needs the labelled calibration images where the model runs. Their licence or privacy may prevent copying them to the device, \eg medical images that cannot leave a site, and a party that receives a calibrated model without them cannot recalibrate at all. Segmentation also runs on board satellites, which send only selected images to the ground \cite{giuffrida2022phisat}, and is proposed for driving planetary rovers \cite{swan2021ai4mars}; there, recalibration needs the labelled images and a native pass over them on the device, with its memory at the new vocabulary (\cref{tab:memory}), or a round trip to the ground. The anchored model needs the bank names, a list of text. Its cost is inference time ($\RtLeanVitLBankRatio\times$ native for CAT-Seg ViT-L with the lean bank pass, \cref{tab:runtimelean}), which is the scope condition on constrained hardware.

\paragraph{Recalibrating many times.} The CRC guarantee holds on average over calibration sets, and for each recalibrated vocabulary separately. Vocabularies that arrive one after another, and are not known in advance, need a guarantee that holds for all $K$ of them at once with probability $1-\delta$ over the calibration set. By the union bound, this requires level $\delta/K$ for each. The anchored source-label scores do not depend on the vocabulary, so a single calibration at level $\delta$ covers every vocabulary. We compare the two with RCPS \cite{bates2021rcps} (Hoeffding--Bentkus bound) at $\alpha=\delta=\RrDelta\%$ on the splits of \cref{sec:main}, half of the images calibrating and half evaluating. The recalibrated native sets become larger than the anchored ones from $K=\RrVitLAdeCross$ for CAT-Seg ViT-L on ADE20K and Pascal Context; at $K=100$ on ADE20K they hold $\RrVitLAdeNatHundred$ against $\RrVitLAdeAnchSize$ labels per pixel. The crossover is at $K=\RrSedAdeCross$ for SED on ADE20K and Pascal Context, $K=\RrSedCocCross$ for SED on COCO-Stuff and $K=\RrFcAdeCross$ for FC-CLIP on ADE20K. When the vocabularies are known in advance, a single calibration under their worst case (\cref{prop:adaptive}) avoids the union bound. If the next vocabulary is chosen after inspecting results on the calibration images, for example by adding the names that fail there, the calibration images are no longer exchangeable with the test images, and recalibrating on them carries no guarantee.

\subsection{Fine-tuning the aggregator}
\label{sec:finetune}
The conversion requires no training, but the aggregator can be fine-tuned for anchored scoring (bank plus candidates, per-label sigmoid loss on COCO-Stuff, $10$k iterations). The plain recipe reduces the ViT-B set size at $845$ labels from $\VitBAnchFull$ to $\FtTenKAnchFull$. Adding hard-negative distractor names reduces it to $\FtDistractAnchFull$, but lowers the top-10 accuracy at $845$ labels on a fixed group of \FtHeldLabels\ ADE20K source labels from $\VitBHeldTopTen$ to $\FtDistractHeldTopTen\%$ and loses the advantage on added labels at equal set size (\cref{tab:finetune}). We retain the released weights with a large bank in the main experiments.
\begin{table*}[tp]
  \centering
  \caption{\textbf{Fine-tuning the CAT-Seg ViT-B aggregator for anchored scoring} (COCO-171 bank). Held-out top-10: accuracy (\%) of the true label among the ten highest anchored scores at $845$ labels, on a fixed group of \FtHeldLabels\ ADE20K source labels held out in an earlier pilot, over \FtProbeImages\ development images. Some of these labels are also COCO-Stuff classes, so the group is not outside the training vocabulary.}
  \label{tab:finetune}
  \adjustbox{max width=\textwidth}{\footnotesize\setlength{\tabcolsep}{4pt}\begin{tabular}{@{}lrrrr@{}}
\toprule
Aggregator & anchored size at +695 & native recal.\ size & held-out top-10 & added-label risk at matched size (anchored / native) \\
\midrule
CAT-Seg ViT-B, COCO-171 bank & 7.32 & 4.99 & 82.7 & 36.0 / 40.8 \\
ViT-B, aggregator fine-tuned 10k & 7.01 & 5.31 & 82.8 & 37.7 / 39.4 \\
ViT-B, fine-tuned with distractors & 6.15 & 5.47 & 80.1 & 44.9 / 43.3 \\
\bottomrule
\end{tabular}}}
\end{table*}

\begin{table}[tp]
  \centering\small
  \caption{\textbf{Detection label-loss surrogate} (COCO $80\!\to\!1{,}204$, COCO $\cup$ LVIS v1, chunks of $40$ names, $\alpha=10\%$): risk on COCO boxes (\%) / labels per ground-truth box. The all-label endpoint covers an unmatched box without localizing it. \OvdPending}
  \label{tab:ovd}
  \setlength{\tabcolsep}{4pt}\begin{tabular}{@{}lrrr@{}}
\toprule
& \multicolumn{2}{c}{native} & \multicolumn{1}{c}{anchored} \\ \cmidrule(lr){2-3}\cmidrule(l){4-4}
Detector & \multicolumn{1}{c}{reused} & \multicolumn{1}{c}{recal.} & \multicolumn{1}{c}{once} \\
\midrule
Grounding DINO T & 9.6/94.7 & 9.9/88.3 & 9.9/34.2 \\
MM-Grounding DINO T & 9.0/23.0 & 9.9/20.2 & 9.9/3.1 \\
Grounding DINO B & 9.8/3.7 & 9.9/3.7 & 9.9/3.6 \\
OWLv2 B (control) & 9.9/8.3 & -- & -- \\
\bottomrule
\end{tabular}
\end{table}

\section{Open-vocabulary detection}
\label{sec:detectionfull}
This section gives the protocol and the full results of \cref{sec:detection}. The protocol was fixed before any score was computed; its deviations are dated in the released protocol document.

\paragraph{Models.}
Grounding DINO~T and~B \cite{liu2024groundingdino} and MM-Grounding DINO~T \cite{zhao2024mmgdino}, a later training of the same architecture, all with their released weights and processor defaults (fp32). The Hugging Face Transformers implementation we use (version 5.17) does not load the decoder heads of two checkpoints as released: MM-Grounding DINO~T receives the box and class heads of the first decoder layer in every layer, and Grounding DINO~B a box head that differs from the checkpoint's. We copy the heads from the checkpoint files after loading; for Grounding DINO~T, whose heads load correctly, this leaves the scores unchanged. OWLv2~B \cite{minderer2023owlv2} is a control whose scores are candidate-independent by construction, since its queries do not see the text and each class is scored by its own sigmoid.

\paragraph{Growth.}
The source vocabulary is the $80$ COCO classes. The $1{,}124$ added labels are the LVIS categories that are not COCO classes, in a fixed seeded order. Two deployments are evaluated. In the first, the prompt is extended with added names until it reaches $256$ tokens ($\OvdGdTOneLabels$ names in total). In the second, all $1{,}204$ names are split into consecutive chunks of $40$, with the source names first, and a box is scored for a label by the chunk that holds it.

\paragraph{Scores and loss.}
The score of a query for a label is the largest token logit over the label's phrase tokens. The score of a ground-truth box for a label is the largest score over the predicted boxes that overlap it with IoU $\ge0.5$, and $-30$, the lowest point of the threshold grid, if there is none. The loss of an image is the fraction of its COCO boxes whose label is missing from the set. Images without a COCO box are excluded ($\OvdGdTImages$ images with $\OvdGdTBoxes$ boxes remain). The set size is the mean number of labels in the set of a COCO box. This loss is evaluated at annotated boxes and stands in for a detection loss: the minimum threshold includes all labels even for an unmatched true box. That endpoint does not produce a localized detection, so the guarantee is not one of object recall; we report unmatched-box rates separately. Labels per ground-truth box are therefore an evaluation metric, not the number of predictions emitted by the detector.

\paragraph{Conversion.}
The image side (feature fusion, query selection and decoder) sees only the bank tokens. Each candidate phrase passes through the text branch together with the bank but is hidden from the image side, and it is scored against the final queries by the model's contrastive head. The queries and boxes are those of the model with the bank as the whole prompt. Before any analysis, we checked on five images that the anchored scores do not change with the order of the candidates or with the other candidates (largest logit difference $\OvdGdTCheckMax$ for Grounding DINO~T, $\OvdGdBCheckMax$ for Grounding DINO~B and $\OvdMmTCheckMax$ for MM-Grounding DINO~T). The bank is either the $80$ COCO names (A, the main configuration) or $80$ Open Images names that are neither COCO nor LVIS concepts (B).

\paragraph{Results.}
\Cref{tab:ovdfull} reports all evaluated configurations. The native source scores move when names are added to the prompt. At LVIS scale the source names fill their own chunks, and the reused threshold stays below the target for every detector at both values of $\alpha$. With one prompt it does so for Grounding DINO~T and MM-Grounding DINO~T, but not for Grounding DINO~B ($\OvdGdBReuseOneRisk\%$ at $\alpha=10\%$, $\OvdGdBBReuseOneRisk\%$ at $\alpha=5\%$). The anchored model with bank A equals the native model at the source vocabulary, is valid for every detector, and gives the smallest sets at LVIS scale, except for Grounding DINO~B at $\alpha=5\%$ ($\OvdGdBBAnchASize$ against $\OvdGdBBRecalSize$ labels per box after recalibration). With bank B it is also valid, but its sets hold $\OvdGdBAnchBSrcSize$--$\OvdGdTAnchBSrcSize$ labels per box already at the source vocabulary; as for the segmenters, the bank changes only the set sizes. OWLv2, calibrated once at the $80$ names, has $\OvdOwlRisk\%$ risk with $\OvdOwlSize$ labels per box at LVIS scale. At $1{,}204$ names the anchored Grounding DINO models take $\OvdGdTRtAnch$--$\OvdGdBRtAnch$\,s per image against $\OvdGdTRtNative$--$\OvdGdBRtNative$\,s for the chunked native models (\cref{tab:runtime}), because the image side runs twice (bank pass and joint pass) rather than once per chunk.

\paragraph{Added labels.}
We apply label-level CRC with a random source vocabulary of $80$ of the $1{,}204$ names (\cref{prop:labels}) and the image-averaged label loss (\cref{prop:newimages}). For the anchored Grounding DINO models, the added-label risk is $\OvdMmTLxAnchExchCal$--$\OvdGdTLxAnchExchCal\%$ on the calibration images and $\OvdMmTLxAnchExchHeld$--$\OvdGdTLxAnchExchHeld\%$ on held-out images; for the native models at LVIS scale it is $\OvdGdTLxNativeExchHeld$--$\OvdMmTLxNativeExchHeld\%$ on held-out images. A box that no predicted box overlaps is covered only at the lowest grid point. With the anchored queries, $\OvdGdTAnchUnlocAdded$--$\OvdMmTAnchUnlocAdded\%$ of the added-label boxes are unmatched in this way, against $\OvdMmTNativeUnlocAdded$--$\OvdGdTNativeUnlocAdded\%$ for the native models (source boxes: $\OvdGdBAnchUnlocSrc$--$\OvdGdTAnchUnlocSrc\%$). Often these boxes alone exceed $\alpha$ for the drawn labels, and the calibrated threshold then includes every label. At $\alpha=10\%$ this happens in $\OvdGdTAnchTrivialImages$--$\OvdMmTAnchTrivialImages\%$ of the draws for the anchored models and $\OvdMmTNativeTrivialImages$--$\OvdGdTNativeTrivialImages\%$ for the native models, and at $\alpha=5\%$ in $\OvdMmTBNativeTrivialImages$--$\OvdMmTBAnchTrivialImages\%$ for both. For OWLv2 the guarantee is informative: $\OvdOwlUnlocAdded\%$ of the added-label boxes are unmatched, the threshold is trivial in $\OvdOwlTrivialImages\%$ of the draws at $\alpha=10\%$, and the added-label risk on held-out images is $\OvdOwlAddedHeld\%$. With the benchmark's own split (source = COCO), the added-label risk is $\OvdGdTLxAnchBenchHeld$--$\OvdGdBLxAnchBenchHeld\%$ for the anchored models and $\OvdGdTLxNativeBenchHeld$--$\OvdGdBLxNativeBenchHeld\%$ for the native models, as \cref{prop:shift} allows.

\paragraph{Accuracy.}
\Cref{tab:ovdacc} gives the detection counterpart of the pixel accuracy of \cref{tab:pixacc}, computed from the same stored scores: at each ground-truth box, the name with the largest box-level score among all $1{,}204$ is compared with the true label. A box that no predicted box overlaps counts as wrong, and ties count against the true label. On COCO boxes, top-1 is $\OvdAccAnchASrcLo$--$\OvdAccAnchASrcHi\%$ for the anchored models with bank A and $\OvdAccNativeSrcLo$--$\OvdAccNativeSrcHi\%$ for the chunked native models. On the $\OvdAccAddedBoxes$ boxes of added labels it falls from $\OvdAccNativeLo$--$\OvdAccNativeHi\%$ (native) to $\OvdAccAnchALo$--$\OvdAccAnchAHi\%$ (anchored). The drop is not only one of localisation: among localised boxes, top-1 falls from $\OvdAccNativeLocLo$--$\OvdAccNativeLocHi\%$ to $\OvdAccAnchALocLo$--$\OvdAccAnchALocHi\%$, since the queries are selected and decoded for the bank names alone. Over all boxes, COCO and added together, top-1 moves from $\OvdAccNativeAllLo$--$\OvdAccNativeAllHi\%$ to $\OvdAccAnchAAllLo$--$\OvdAccAnchAAllHi\%$ and rises for \OvdAccAllRiseCount\ of the \OvdAccAllDetectors\ detectors. Bank B lowers COCO top-1 to $\OvdAccAnchBSrcLo$--$\OvdAccAnchBSrcHi\%$. OWLv2, whose queries do not see the text, reaches $\OvdOwlAccOwlAddTop\%$ on the added boxes, so the cost comes from conditioning the image side on a fixed bank, not from candidate independence itself. LVIS labels are not exhaustive, so a box whose top name is correct but unannotated counts as wrong for every model alike.
\begin{table*}[tp]
  \centering
  \caption{\textbf{Detection, every arm} (COCO $80\!\to\!1{,}204$, COCO $\cup$ LVIS v1; mean over \NumSplits\ splits). Risk (\%) on COCO boxes; for native reused and recalibrated, the one-prompt and LVIS-scale deployments. Size: labels per COCO box at the $\OvdGdTOneLabels$-name prompt and at $1{,}204$ names. \OvdPending}
  \label{tab:ovdfull}
  \adjustbox{max width=\textwidth}{\footnotesize\setlength{\tabcolsep}{4pt}\setlength{\tabcolsep}{3pt}\begin{tabular}{@{}llrrrrrr@{}}
\toprule
& & \multicolumn{3}{c}{$\alpha=.10$} & \multicolumn{3}{c}{$\alpha=.05$} \\ \cmidrule(lr){3-5}\cmidrule(l){6-8}
Detector & arm & risk & size, one prompt & size, LVIS & risk & size, one prompt & size, LVIS \\
\midrule
Grounding DINO T & native, source & 9.9 & -- & -- & 4.9 & -- & -- \\
 & native reused & 9.6, 9.6 & 1.7 & 94.7 & 4.6, 4.8 & 3.8 & 291.1 \\
 & native recal. & 9.9, 9.9 & 1.6 & 88.3 & 4.9, 4.9 & 3.5 & 280.5 \\
 & anchored, bank A & 9.9 & 1.7 & 34.2 & 4.9 & 4.0 & 121.2 \\
 & anchored, bank B & 9.9 & 16.0 & 295.4 & 1.7 & 93.9 & 1158.5 \\
\midrule
MM-Grounding DINO T & native, source & 9.9 & -- & -- & 4.9 & -- & -- \\
 & native reused & 9.3, 9.0 & 1.7 & 23.0 & 4.3, 4.5 & 4.1 & 63.8 \\
 & native recal. & 9.9, 9.9 & 1.6 & 20.2 & 4.9, 4.9 & 3.4 & 55.8 \\
 & anchored, bank A & 9.9 & 1.2 & 3.1 & 4.9 & 2.0 & 9.5 \\
 & anchored, bank B & 9.9 & 8.5 & 84.1 & 4.9 & 42.6 & 474.1 \\
\midrule
Grounding DINO B & native, source & 9.9 & -- & -- & 4.9 & -- & -- \\
 & native reused & 18.3, 9.8 & 0.9 & 3.7 & 9.9, 4.9 & 1.2 & 9.7 \\
 & native recal. & 9.9, 9.9 & 1.2 & 3.7 & 4.9, 4.9 & 2.2 & 9.7 \\
 & anchored, bank A & 9.9 & 1.0 & 3.6 & 4.9 & 1.4 & 12.2 \\
 & anchored, bank B & 9.9 & 7.7 & 243.0 & 4.3 & 71.5 & 1026.7 \\
\midrule
OWLv2 B (control) & reused & 9.9 & -- & 8.3 & 4.9 & -- & 28.5 \\
\bottomrule
\end{tabular}}}
\end{table*}

\begin{table*}[tp]
  \centering
  \caption{\textbf{Detection accuracy at $1{,}204$ names} (COCO $80\!\to\!1{,}204$, COCO $\cup$ LVIS v1; from the stored box-level scores). Top-1: the true label has the largest score among all names at the ground-truth box (\%); class mean: top-1 averaged over the labels present; localised: some predicted box overlaps the ground-truth box with IoU $\ge0.5$; top-1 $\mid$ loc.: top-1 among localised boxes. Unlocalised boxes and ties count as wrong.}
  \label{tab:ovdacc}
  \adjustbox{max width=\textwidth}{\footnotesize\setlength{\tabcolsep}{4pt}\setlength{\tabcolsep}{4pt}\begin{tabular}{@{}llrrrrrr@{}}
\toprule
& & \multicolumn{2}{c}{COCO boxes} & \multicolumn{4}{c}{LVIS (added) boxes} \\ \cmidrule(lr){3-4}\cmidrule(l){5-8}
Detector & scoring & top-1 & class mean & localised & top-1 & top-1 $\mid$ loc. & class mean \\
\midrule
Grounding DINO T & native, chunks of 40 & 69.7 & 61.3 & 83.0 & 23.8 & 28.6 & 15.5 \\
 & anchored, bank A & 79.7 & 73.8 & 76.4 & 15.5 & 20.3 & 9.1 \\
 & anchored, bank B & 48.5 & 35.6 & 67.6 & 13.3 & 19.7 & 10.6 \\
\midrule
MM-Grounding DINO T & native, chunks of 40 & 65.3 & 53.6 & 88.4 & 34.2 & 38.7 & 24.5 \\
 & anchored, bank A & 85.0 & 79.9 & 70.1 & 14.7 & 21.0 & 8.8 \\
 & anchored, bank B & 43.0 & 24.0 & 72.8 & 17.6 & 24.2 & 12.0 \\
\midrule
Grounding DINO B & native, chunks of 40 & 89.7 & 86.3 & 85.1 & 24.7 & 29.0 & 16.6 \\
 & anchored, bank A & 88.0 & 85.3 & 74.1 & 16.5 & 22.3 & 12.0 \\
 & anchored, bank B & 51.3 & 43.6 & 71.5 & 17.3 & 24.2 & 17.4 \\
\midrule
OWLv2 B (control) & native & 62.6 & 67.6 & 89.3 & 55.4 & 62.0 & 49.5 \\
\bottomrule
\end{tabular}}}
\end{table*}

\section{Reserved images}
\label{sec:reservefull}
This section reports the confirmation of \cref{sec:reserve} in full. $2{,}000$ ADE20K training images were drawn with a recorded seed, listed in a manifest and never used in development; one has no sampled source-label pixel and is excluded (\cref{sec:sampling}). They were passed through CAT-Seg ViT-L with the $683$-name bank once, and the stored scores were analysed twice.

\paragraph{Source labels.}
The first analysis used a design fixed before the reserved images were evaluated: calibration on the first $999$ images and evaluation on the rest (\cref{tab:confirm}). The risks quoted in \cref{sec:reserve} are those of this single split, so the anchored $\ConfPrimaryAnchRisk\%$ varies around the expected risk. Over \NumSplits\ random splits of the reserved images, the anchored risk is $\ConfSecondaryAnchRisk\%$ and the reused native risk $\ConfSecondaryNativeFrozenRisk\%$.
\begin{table*}[tp]
  \centering
  \caption{\textbf{Reserved images} ($\ConfImages$ ADE20K training images not used in development; CAT-Seg ViT-L, $683$-name bank; $150\!\to\!845$). The pre-declared split calibrates on the first $999$ images and evaluates on the rest.}
  \label{tab:confirm}
  \adjustbox{max width=\textwidth}{\footnotesize\setlength{\tabcolsep}{4pt}\begin{tabular}{@{}lrrrrr@{}}
\toprule
& & \multicolumn{2}{c}{risk} & \multicolumn{2}{c}{size} \\
\cmidrule(lr){3-4}\cmidrule(lr){5-6}
Design & $\alpha$ & native reused & anchored & anchored & native recal. \\
\midrule
pre-declared split & .10 & 13.8 & 8.4 & 2.42$\to$4.55 & 3.71 \\
pre-declared split & .05 & 7.0 & 4.5 & 3.80$\to$8.82 & 8.07 \\
500 random splits & .10 & 15.0 & 9.7 & 2.33$\to$4.43 & 3.49 \\
500 random splits & .05 & 8.3 & 4.7 & 3.83$\to$9.11 & 7.56 \\
\bottomrule
\end{tabular}}}
\end{table*}

\paragraph{Added labels.}\label{sec:reserveadded}
The second analysis tested the added-label guarantee under a protocol written after the development results of \cref{sec:labelguarantee} and before any added-label annotation of a reserved image was decoded; the protocol is released with the code. The ADE20K-847 labels at the stored pixels come from the ADE20K-full export and agree with the ADE20K-150 annotation on $\LxRsvAgree\%$ of the pixels labelled in both. Only designs that use the stored scores were pre-specified. The primary test was whether the anchored label-level threshold meets the target under a random source vocabulary. Its added-label risk is $\LxRsvLabRisk\%$ at $\alpha=10\%$ and $\LxRsvBLabRisk\%$ at $5\%$, with $\LxRsvLabSize$ and $\LxRsvBLabSize$ labels per pixel. Native recalibration at $845$ labels is also valid ($\LxRsvRecalRisk$ and $\LxRsvBRecalRisk\%$) with larger sets ($\LxRsvRecalSize$ and $\LxRsvBRecalSize$), and the image-level threshold is not ($\LxRsvImgRisk$ and $\LxRsvBImgRisk\%$). On the benchmark's own split the label-level threshold misses $\LxRsvSplitLabRisk\%$ at $\alpha=10\%$, with an added--source gap of $\Delta(\hat t)=\LxRsvSplitLabDelta$ points (\cref{prop:shift}).

\section{Beyond the marginal risk}
\label{sec:beyondfull}
The guarantee of \cref{prop:transfer} concerns the marginal image-level risk. This section examines how the risk is distributed over classes and scenes (\cref{sec:beyond}), how per-name thresholds bound it for each name (\cref{sec:pername}), whether the anchored threshold remains valid under stricter losses (\cref{sec:losses}) and other set constructions (\cref{sec:constructions}), and how many calibration images it needs (\cref{sec:calsize}).

\subsection{Risk by class and scene}
\label{sec:beyond}
Averaged over the source classes, the anchored threshold has a lower risk than native recalibration in every ADE20K configuration ($\VitLBankMacroAnch$ against $\VitLBankMacroReplay\%$ for ViT-L) and for CAT-Seg and SED on Pascal Context and COCO-Stuff, but not for FC-CLIP on those benchmarks (\cref{tab:strata,tab:stratapascal,tab:stratacoco}). By scene it is worse: in the worst ADE20K scene category (\VitBWorstSceneName) its risk exceeds that of recalibration for every model except SED with the $683$-name bank, most for ViT-B with the COCO-171 bank ($\VitBWorstSceneAnch$ against $\VitBWorstSceneReplay\%$; no other stratum in any configuration exceeds $2\alpha$). The $683$-name bank narrows this gap for each model run with both banks (CAT-Seg ViT-L, SED and FC-CLIP).
\begin{table*}[tp]
  \centering
  \caption{\textbf{Risk by stratum} on ADE20K at $845$ labels ($\alpha=10\%$): class-averaged risk over the source classes, and the worst ADE20K scene category (scenes with at least $20$ evaluation images). The marginal risk of the anchored model is at the target in every row.}
  \label{tab:strata}
  \adjustbox{max width=\textwidth}{\footnotesize\setlength{\tabcolsep}{4pt}\begin{tabular}{@{}lrrrrrr@{}}
\toprule
& \multicolumn{3}{c}{class-averaged risk} & \multicolumn{3}{c}{worst ADE scene} \\
\cmidrule(lr){2-4}\cmidrule(lr){5-7}
Configuration & native reused & native recal. & anchored & native reused & native recal. & anchored \\
\midrule
CAT-Seg ViT-B, COCO-171 bank & 33.0 & 23.3 & 17.2 & 18.7 & 11.1 & 20.7 \\
CAT-Seg ViT-L, COCO-171 bank & 30.2 & 20.6 & 14.3 & 15.4 & 9.7 & 17.9 \\
CAT-Seg ViT-L, 683-name bank & 30.2 & 20.6 & 16.6 & 15.4 & 9.7 & 12.2 \\
CAT-Seg + DeCLIP EVA-L, 683-name bank & 25.2 & 17.6 & 13.8 & 15.3 & 9.9 & 13.5 \\
SED ConvNeXt-L, COCO-171 bank & 14.2 & 20.5 & 14.9 & 7.5 & 11.3 & 16.9 \\
SED ConvNeXt-L, fast mode (top-8 rejection) & 25.8 & 20.7 & 14.9 & 13.0 & 9.9 & 16.9 \\
SED ConvNeXt-L, 683-name bank & 14.2 & 20.5 & 17.3 & 7.5 & 11.3 & 10.7 \\
FC-CLIP ConvNeXt-L, 133-name bank & 24.1 & 18.7 & 17.4 & 15.8 & 10.9 & 15.0 \\
FC-CLIP ConvNeXt-L, 701-name bank & 24.1 & 18.7 & 16.6 & 15.8 & 10.9 & 13.5 \\
\bottomrule
\end{tabular}}}
\end{table*}
\begin{table*}[tp]
  \centering
  \caption{\textbf{Class-averaged risk on Pascal Context} (columns as in \cref{tab:strata}; no scene labels).}
  \label{tab:stratapascal}
  \adjustbox{max width=\textwidth}{\footnotesize\setlength{\tabcolsep}{4pt}\begin{tabular}{@{}lrrr@{}}
\toprule
& \multicolumn{3}{c}{class-averaged risk} \\
\cmidrule(l){2-4}
Configuration & native reused & native recal. & anchored \\
\midrule
CAT-Seg ViT-B, COCO-171 bank, Pascal 59$\to$458 & 19.4 & 12.2 & 9.5 \\
CAT-Seg ViT-B, 683-name bank, Pascal 59$\to$458 & 19.4 & 12.2 & 11.0 \\
CAT-Seg ViT-L, 683-name bank, Pascal 59$\to$458 & 17.4 & 11.1 & 9.8 \\
CAT-Seg + DeCLIP EVA-L, 683-name bank, Pascal 59$\to$458 & 16.0 & 10.5 & 9.2 \\
SED ConvNeXt-L, 683-name bank, Pascal 59$\to$458 & 5.7 & 12.1 & 10.9 \\
FC-CLIP ConvNeXt-L, 701-name bank, Pascal 59$\to$458 & 15.9 & 9.3 & 10.4 \\
\bottomrule
\end{tabular}}}
\end{table*}
\begin{table*}[tp]
  \centering
  \caption{\textbf{Class-averaged risk on COCO-Stuff} (columns as in \cref{tab:strata}; no scene labels).}
  \label{tab:stratacoco}
  \adjustbox{max width=\textwidth}{\footnotesize\setlength{\tabcolsep}{4pt}\begin{tabular}{@{}lrrr@{}}
\toprule
& \multicolumn{3}{c}{class-averaged risk} \\
\cmidrule(l){2-4}
Configuration & native reused & native recal. & anchored \\
\midrule
CAT-Seg ViT-B, 683-name bank, COCO 171$\to$1,283 & 24.9 & 20.2 & 19.0 \\
CAT-Seg ViT-L, 683-name bank, COCO 171$\to$1,283 & 23.2 & 18.9 & 18.0 \\
CAT-Seg + DeCLIP EVA-L, 683-name bank, COCO 171$\to$1,283 & 22.3 & 17.7 & 17.1 \\
SED ConvNeXt-L, 683-name bank, COCO 171$\to$1,283 & 8.6 & 18.8 & 18.6 \\
FC-CLIP ConvNeXt-L, 701-name bank, COCO 171$\to$1,283 & 16.0 & 9.4 & 10.0 \\
\bottomrule
\end{tabular}}}
\end{table*}

\subsection{Per-name thresholds}
\label{sec:pername}
The image-level threshold bounds the risk averaged over images, and single names can lie far above it. A deployer who needs a bound for each name can calibrate each name on its own images instead, as in Mondrian, or class-conditional, conformal prediction \cite{vovk2005alrw,ding2023classconditional}. For a name $c$, let $\ell_{i,c}(t)$ be the fraction of $c$'s pixels in calibration image $i$ scored below $t$, over the $n_c$ calibration images annotated with $c$, and let $\hat t_c$ be the largest grid threshold with $(\sum_i\ell_{i,c}(t)+1)/(n_c+1)\le\alpha$. If the images containing $c$ are exchangeable, a test image containing $c$ misses in expectation at most a fraction $\alpha$ of $c$'s pixels. Conditioning on $n_c$ preserves exchangeability, and a threshold exists once $n_c\ge 1/\alpha-1$ ($9$ images at $\alpha=10\%$, $19$ at $5\%$). The bound holds for each name separately, not for all names at once.

\paragraph{What anchoring adds.} On anchored scores, $\ell_{i,c}$ does not depend on the requested vocabulary. Each $\hat t_c$ therefore stays valid as names are added, and a new name is calibrated from its own images, each annotated only with that name's mask. The images must be drawn because the name is present, not curated, so that they are exchangeable with the test images containing it; a name of frequency $p$ needs about $(1/\alpha-1)/p$ screened images. On native scores, per-name thresholds calibrated at the source vocabulary leave $\PnNatOverAdeVitL$ of $\PnNatEstAdeVitL$ source names above $\alpha$ for CAT-Seg ViT-L on ADE20K and $\PnNatOverAdeFc$ of $\PnNatEstAdeFc$ for FC-CLIP once the vocabulary grows (\cref{tab:pername}). Native per-name thresholds calibrated at the grown vocabulary are valid, but every added name then requires recalibrating every name.

\paragraph{Two options and their hybrid.} The image-level threshold needs one calibration, includes every name and gives the smallest sets. Per-name thresholds bound each calibrated name, at the price of labelled images per name and larger sets, and names with fewer than $1/\alpha-1$ calibration images get no threshold. The two combine: calibrated names take the lower of $\hat t_c$ and the image-level threshold, and the other names take the image-level threshold. The loss is non-increasing as a threshold falls, so each image's loss is at most its loss under the image-level threshold, and each name's loss at most its loss under $\hat t_c$; both guarantees hold at once.

\paragraph{Results.} We repeat $500$ random half/half splits of the images on six score stores (\cref{tab:pername}). Per-name thresholds leave no calibrated name above $\alpha$ in any store at either level, while the image-level threshold leaves $\PnImgOverAdeVitL$ of the $\PnCalAdeVitL$ calibrated ADE20K names above $\alpha=10\%$ for CAT-Seg ViT-L. Their sets are $\PnSizeRatioLo$--$\PnSizeRatioHi\times$ those of the image-level threshold at $\alpha=10\%$. The hybrid keeps the image risk at or below $\PnHybRiskHi\%$, leaves no calibrated name above $\alpha$, and its sets are $\PnHybRatioLo$--$\PnHybRatioHi\times$ those of the image-level threshold. At $\alpha=10\%$, $\PnCalAdeVitL$ of the $\PnPresentAdeVitL$ names present in the ADE20K images have enough calibration images. With at most $20$ calibration images per name, chosen at random, no calibrated name is above $\alpha$ either.
\begin{table*}[tp]
  \centering
  \caption{\textbf{Per-name thresholds} on anchored scores at the grown vocabulary ($500$ half/half splits, set sizes on the first $20$). \emph{Calibrated}: names with a per-name threshold per split, of the names present. \emph{Over $\alpha$}: calibrated names, in at least $10$ splits, whose held-out risk averaged over splits exceeds $\alpha$ by more than three standard errors (for the image-level threshold, among the same names). Per-name sets hold the calibrated names only. \emph{Lower of both}: calibrated names take the lower of the two thresholds, the others the image-level one; image risk is the loss \eqref{eq:loss}. \emph{Native, reused}: native per-name thresholds calibrated at the source vocabulary, of the source names calibrated in at least $10$ splits.}
  \label{tab:pername}
  \adjustbox{max width=\textwidth}{\footnotesize\setlength{\tabcolsep}{4pt}\begin{tabular}{@{}lllrrrrrrrrr@{}}
\toprule
& & & & \multicolumn{2}{c}{image-level} & \multicolumn{2}{c}{per-name} & \multicolumn{3}{c}{lower of both} & native, reused \\
\cmidrule(lr){5-6}\cmidrule(lr){7-8}\cmidrule(lr){9-11}\cmidrule(l){12-12}
$\alpha$ & Benchmark & Model & calibrated & over $\alpha$ & size & over $\alpha$ & size & image risk & over $\alpha$ & size & over $\alpha$ \\
\midrule
\multirow{6}{*}{$10\%$} & ADE20K & CAT-Seg ViT-L & 160 of 836 & 154 & 4.47 & 0 & 10.94 & 4.8 & 0 & 13.30 & 89 of 131 \\
 & ADE20K & SED & 160 of 836 & 145 & 6.29 & 0 & 14.61 & 4.1 & 0 & 18.46 & 0 of 131 \\
 & ADE20K & FC-CLIP & 160 of 836 & 135 & 11.56 & 0 & 25.29 & 4.7 & 0 & 33.12 & 65 of 131 \\
 & Pascal Context & CAT-Seg ViT-L & 87 of 270 & 59 & 2.45 & 0 & 3.58 & 4.9 & 0 & 4.48 & 41 of 59 \\
 & Pascal Context & SED & 87 of 270 & 70 & 2.86 & 0 & 5.52 & 4.3 & 0 & 6.71 & 0 of 59 \\
 & COCO-Stuff & SED & 198 of 660 & 176 & 3.61 & 0 & 11.68 & 3.4 & 0 & 13.04 & 0 of 161 \\
\midrule
\multirow{6}{*}{$5\%$} & ADE20K & CAT-Seg ViT-L & 96 of 836 & 82 & 8.95 & 0 & 9.04 & 2.5 & 0 & 15.26 & 65 of 104 \\
 & ADE20K & SED & 96 of 836 & 76 & 15.96 & 0 & 13.12 & 1.9 & 0 & 25.67 & 0 of 104 \\
 & ADE20K & FC-CLIP & 96 of 836 & 67 & 31.37 & 0 & 21.06 & 1.9 & 0 & 46.70 & 37 of 104 \\
 & Pascal Context & CAT-Seg ViT-L & 61 of 270 & 41 & 3.86 & 0 & 4.12 & 2.1 & 0 & 6.04 & 35 of 54 \\
 & Pascal Context & SED & 61 of 270 & 44 & 4.90 & 0 & 6.88 & 1.9 & 0 & 9.55 & 0 of 54 \\
 & COCO-Stuff & SED & 129 of 660 & 125 & 6.25 & 0 & 12.03 & 2.2 & 0 & 15.20 & 0 of 147 \\
\bottomrule
\end{tabular}}}
\end{table*}

\subsection{Calibration-set size}
\label{sec:calsize}
We calibrate on the first $n$ images of each split, for $n$ from $10$ to $\CalSizeMaxN$. At every $n$, the anchored and the recalibrated native thresholds have similar mean risk, spread across splits and fraction of splits above $\alpha$ (\cref{tab:calsize}). At small $n$ both are conservative because of CRC's finite-sample correction. On ADE20K the overhead settles within a fifth of a label of its final value from $n=\VitLBankCalSettleN$ for ViT-L and $n=\SedBankCalSettleN$ for SED with the $683$-name bank, and from $n=\VitBCalSettleN$ for ViT-B with the COCO-171 bank, while for FC-CLIP both set sizes keep shrinking up to $n=\CalSizeMaxN$. At very small $n$ the anchored sets can be smaller than the recalibrated native ones, because the native score at $845$ labels has the heavier tail.
\begin{table*}[tp]
  \centering
  \caption{\textbf{Risk (\%) / set size at the grown vocabulary against the number of calibration images $n$} ($\alpha=10\%$). Evaluation on the second half of each split; calibration on the first $n$ images of the first half.}
  \label{tab:calsize}
  \adjustbox{max width=\textwidth}{\footnotesize\setlength{\tabcolsep}{4pt}\InputIfFileExists{tables/calsize}{}{\errmessage{missing file tables/calsize}}}
\end{table*}

\subsection{Stricter losses}
\label{sec:losses}
We replace the loss \eqref{eq:loss} by four stricter image losses, $\ind[\text{miss}>25\%]$, $\ind[\text{miss}>10\%]$, a class-balanced miss rate and the worst class in the image, and recalibrate by CRC on each. The anchored threshold is at the target in every configuration ($\LossAnchRiskLo$--$\LossAnchRiskHi\%$), and the reused native threshold fails or is over-conservative under each loss, as under \eqref{eq:loss}. Under the worst-class loss the anchored sets are smaller than the recalibrated native ones for every CAT-Seg configuration, but not for FC-CLIP on ADE20K (\cref{tab:losses}). Like \eqref{eq:loss}, each of these losses counts a source-labelled pixel as covered only if its annotated name is in the set; a set that holds only a finer added name for the same object misses it. Pixels of added classes enter only the added-label risk (\cref{sec:labelguarantee}).
\begin{table*}[tp]
  \centering
  \caption{\textbf{Risk (\%) and set size under five image losses} ($\alpha=10\%$); thresholds are recalibrated by CRC on each loss. Risk $0.0$ with $845.0$ labels: CRC returned the all-label endpoint, as no threshold on the grid meets the target.}
  \label{tab:losses}
  \adjustbox{max width=\textwidth}{\footnotesize\setlength{\tabcolsep}{4pt}\InputIfFileExists{tables/losses_a}{}{\errmessage{missing file tables/losses_a}}}
\end{table*}
\begin{table*}[tp]
  \centering
  \caption{\textbf{\Cref{tab:losses}, continued} (part 2 of 2).}
  \adjustbox{max width=\textwidth}{\footnotesize\setlength{\tabcolsep}{4pt}\InputIfFileExists{tables/losses_b}{}{\errmessage{missing file tables/losses_b}}}
\end{table*}

\subsection{Set constructions}
\label{sec:constructions}
\Cref{prop:transfer} requires the set construction, and not only the score, to be candidate-independent. With the per-label threshold, the anchored thresholds calibrated at the source and at the grown vocabulary coincide in every configuration, as the proposition predicts. Constructions that normalise over the requested vocabulary remain candidate-dependent and fail even on anchored scores, which is why the main text uses the per-label threshold \cite{sadinle2019least}: the softmax cutoff has $\SoftmaxAnchRiskLo$--$\SoftmaxAnchRiskHi\%$ risk, RAPS \cite{angelopoulos2021raps} up to $\RapsAnchRiskHi\%$ and APS \cite{romano2020aps} $\ApsAnchRiskLo$--$\ApsAnchRiskHi\%$ (\cref{tab:constructions}).
\begin{table*}[tp]
  \centering
  \caption{\textbf{Four set constructions on the same scores} ($\alpha=10\%$; RAPS with $\lambda=0.01$, $k=5$). The anchored per-label threshold is the only construction whose reused and recalibrated sizes coincide.}
  \label{tab:constructions}
  \adjustbox{max width=\textwidth}{\footnotesize\setlength{\tabcolsep}{4pt}\InputIfFileExists{tables/constructions_a}{}{\errmessage{missing file tables/constructions_a}}}
\end{table*}
\begin{table*}[tp]
  \centering
  \caption{\textbf{\Cref{tab:constructions}, continued} (part 2 of 2).}
  \adjustbox{max width=\textwidth}{\footnotesize\setlength{\tabcolsep}{4pt}\InputIfFileExists{tables/constructions_b}{}{\errmessage{missing file tables/constructions_b}}}
\end{table*}

\section{Deployments with growing vocabularies}
\label{sec:deploy}
This section describes deployments in which the vocabulary changes after calibration and the guarantee that applies to each. \Cref{tab:regimes} compares calibrating once with recalibrating in each case.

\paragraph{Robots asked for new objects.} OK-Robot builds its memory of a home with an open-vocabulary detector run on a fixed list of ScanNet200 names and later answers requests for arbitrary objects \cite{liu2024okrobot}. HomeRobot sets the detector's class list for each skill and episode, and its goal objects include unseen categories \cite{yenamandra2023homerobot}. VLMaps receives its list of obstacle categories at run time \cite{huang2023vlmaps}. The anchored threshold keeps the source-label guarantee (\cref{prop:transfer}) for the classes calibrated before deployment, whatever names are requested later.

\paragraph{Names chosen per image.} Image taggers, captioners and language models choose the names from the test image \cite{zhang2024ram,ren2024groundedsam,gu2024conceptgraphs,vocabfree2025,pi2023detgpt}. Such a rule cannot be replayed on the calibration images unless it is fixed and known; the anchored threshold keeps the source-label guarantee under any rule (\cref{prop:transfer,sec:adaptive}).

\paragraph{Catalogues that grow.} The iNaturalist species benchmark grew from about $5{,}000$ species \cite{vanhorn2018inat} to $10{,}000$ \cite{vanhorn2021inat}, and taxonomic names ``are added, removed, and changed as biologists discover and classify new and existing species'' \cite{stevens2024bioclip}. Anomaly-segmentation benchmarks for driving state that networks trained on a fixed set of classes are ``ill-equipped to handle previously-unseen objects'' \cite{chan2021smiyc}. When the annotated classes are a random sample of the catalogue, label-level calibration bounds the miss rate of the others (\cref{prop:labels}); when the added classes are harder, the gap bound applies (\cref{prop:shift}).

\paragraph{Scope.} Open-vocabulary models are less accurate than supervised ones in these domains: on the MESS benchmark, zero-shot CAT-Seg reaches about half of the supervised mIoU in earth monitoring and medicine, and less in engineering and agriculture \cite{blumenstiel2023mess}. The guarantee bounds the miss rate at any accuracy, so lower accuracy appears as larger sets rather than more misses. The risk is pooled over the pixels and images of the calibration distribution. Classes that need a bound of their own, such as hazards, require class-conditional calibration (\cref{sec:beyond}), and the images seen in deployment must be exchangeable with the calibration images.

\end{document}